\documentclass[letterpaper]{article} 
\usepackage{aaai23}  
\usepackage{times}  
\usepackage{helvet}  
\usepackage{courier}  
\usepackage[hyphens]{url}  
\usepackage{graphicx} 
\usepackage{natbib}  
\usepackage{caption} 
\usepackage{algorithm}
\usepackage{algorithmic}

\usepackage{hhline}
\usepackage{amsmath,amssymb}
\usepackage{colortbl}
\usepackage{diagbox}
\usepackage{subeqnarray}
\usepackage{cases}
\usepackage{setspace}
\usepackage{tabularx}
\usepackage{booktabs}
\usepackage{dcolumn}
\usepackage{graphicx}
\usepackage{subfigure}
\usepackage{float}
\usepackage{verbatim}
\usepackage{epsfig}
\usepackage{amsmath}
\usepackage{amssymb}
\usepackage{array}
\usepackage{epstopdf}
\usepackage{color}
\usepackage{xcolor}
\usepackage{cite}
\usepackage{psfrag}
\usepackage{cite}
\usepackage{multirow}
\usepackage{color}
\usepackage{bm}
\usepackage{adjustbox}
\usepackage{overpic}
\usepackage{rotating}
\usepackage{multicol}

\usepackage{newfloat}
\usepackage{listings}
\DeclareCaptionStyle{ruled}{labelfont=normalfont,labelsep=colon,strut=off} 
\floatstyle{ruled}
\newfloat{listing}{tb}{lst}{}
\floatname{listing}{Listing}
\title{Learning Single Image Defocus Deblurring with Misaligned Training Pairs}
\author {
    Yu Li\textsuperscript{\rm 1}, 
    Dongwei Ren\textsuperscript{\rm 1}\thanks{Corresponding Author},
    Xinya Shu\textsuperscript{\rm 1},
    Wangmeng Zuo\textsuperscript{\rm 1}
}
\affiliations{
    \textsuperscript{\rm 1} School of Computer Science and Technology, Harbin Institute of Technology\\
    liyuhit@outlook.com, rendongweihit@gmail.com, shuxinyahit@outlook.com, wmzuo@hit.edu.cn
}

\usepackage{bibentry}

\begin{document}
	
	\maketitle
	
	\begin{abstract}
		By adopting popular pixel-wise loss, existing methods for defocus deblurring heavily rely on well aligned training image pairs. 
		Although training pairs of ground-truth and blurry images are carefully collected, \emph{e.g.}, DPDD dataset, misalignment is inevitable between training pairs, making existing methods possibly suffer from deformation artifacts.  
		%
		In this paper, we propose a joint deblurring and reblurring learning (JDRL) framework for single image defocus deblurring with misaligned training pairs. 
		%
		Generally, JDRL consists of a deblurring module and a spatially invariant reblurring module, by which deblurred result can be adaptively supervised by ground-truth image to recover sharp textures while maintaining spatial consistency with the blurry image.
		First, in the deblurring module, a bi-directional optical flow-based deformation is introduced to tolerate spatial misalignment between deblurred and ground-truth images.  
		%
		Second, in the reblurring module, deblurred result is reblurred to be spatially aligned with blurry image, by predicting a set of isotropic blur kernels and weighting maps.    
		Moreover, we establish a new single image defocus deblurring (SDD) dataset, further validating our JDRL and also benefiting future research.  
		Our JDRL can be applied to boost defocus deblurring networks in terms of both quantitative metrics and visual quality on DPDD, RealDOF and our SDD datasets. 

	\end{abstract}
	
	\section{Introduction}
	
	Defocus blur occurs when the Depth of Field (DoF) of a camera is not enough to cover the whole captured scene. 
	Although having potential benefits in photography aesthetics, defocus blur degrades visual quality of images, thereby bringing difficulty for downstream tasks like object detection \cite{kong2020foveabox,Zheng_2022_CVPR} and tracking \cite{Sun_2022_CVPR}.
	Defocus deblurring is a challenging task due to spatially variant blur kernels. 
	According to the input data type, existing defocus deblurring methods can be categorized into single image defocus deblurring and dual-pixel defocus deblurring. 
	Compared with dual-pixel defocus deblurring \cite{abuolaim2020defocus,pan2021dual,abuolaim2020learning,xin2021defocus} that requires specialized hardware \cite{sliwinski2013simple,Herrmann_2020_CVPR}, single image defocus deblurring is a more general yet challenging task, which is the focus of this work. 
	A typical solution to single image defocus deblurring is to estimate a defocus blur map at first, and then non-blind deconvolution \cite{krishnan2009fast,fish1995blind} can be adopted to predict the latent sharp image~\cite{Shi_2015_CVPR,yi2016lbp,d2016non,karaali2017edge,Lee_2019_CVPR,Park_2017_CVPR}. 
	However, methods in this line yield unsatisfactory performance in real-world scenes due to inaccurate estimation of defocus maps.
	
	Recently, deep learning has boosted the performance of defocus deblurring methods \cite{abuolaim2020defocus,son2021single,lee2021iterative}, making the acquirement of abundant training data an urgent need. 
	\begin{figure}[!t]\small
		\hspace{1cm}
		\centering\begin{overpic}[width=0.47\textwidth]{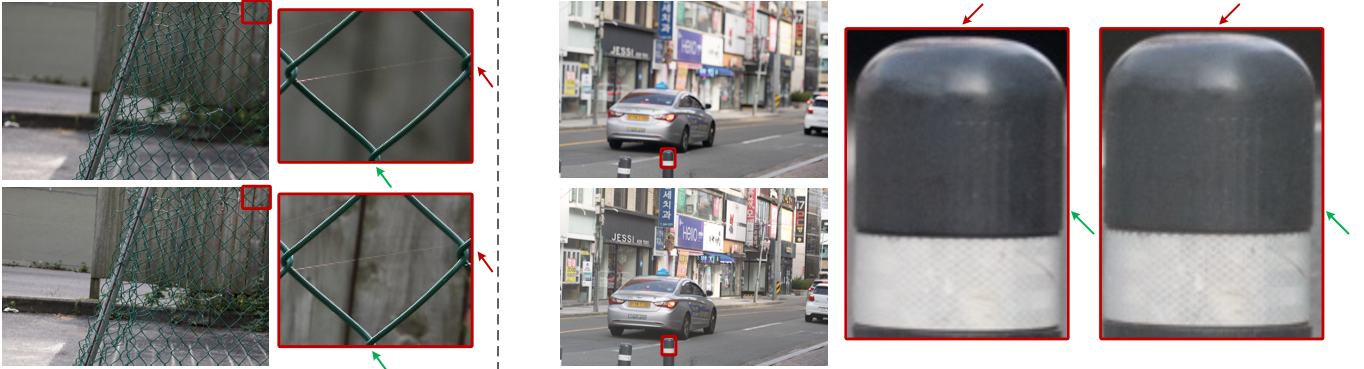}
			\put(-4.5,20){\footnotesize $\bm{I}_B$}
			\put(-4.5,7){\footnotesize $\bm{I}_S$}
			\put(36.7,7){\footnotesize $\bm{I}_S$}
			\put(36.5,20){\footnotesize $\bm{I}_B$}
			
			\put(13,-3){\small (a) DPDD}
			\put(59,-3){\small (b) RealDOF}
		\end{overpic}
		\caption{\small Misaligned image pairs of ground-truth $\bm{I}_S$ and blurry $\bm{I}_B$ in DPDD and RealDOF datasets. 
			Although ground-truth and blurry image pairs in these datasets are designed to be aligned, slight misalignment still exists.}
		\label{fig:intro}
	\end{figure}
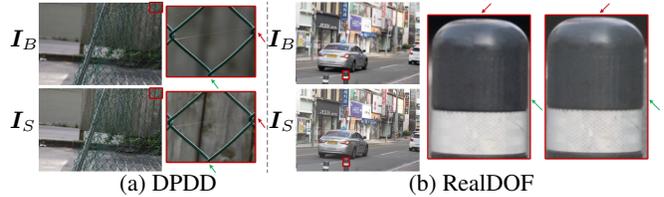
	A common choice for expanding training sources is data synthesis, yet the domain gap between real-world scenes and synthesized ones makes it inappropriate for training defocus deblurring networks. 
	%
	%
	{{Currently, DPDD \cite{abuolaim2020defocus} is the most popular real-world defocus deblurring dataset.}}
	Although originally targeted at dual-pixel defocus deblurring, DPDD also provides pairwise blurry and sharp images, thus benchmarking single image defocus deblurring.
	In DPDD dataset, training pairs are intended to be well aligned by adjusting camera aperture via remote control signal. 
	However, misalignment between blurry and sharp images are still inevitable as shown in Fig. \ref{fig:intro}.
	In another testing dataset RealDOF \cite{lee2021iterative}, a dual-camera system is established to capture image pairs, but spatial misalignment can still be observed.  
	%
	By adopting pixel-wise loss, \emph{e.g.}, MSE or $\ell_1$-norm, spatial deformation is also learned by deblurrinlg network, possibly yielding deformation artifacts in deblurred image, as shown in Fig.~\ref{fig:MDD_distort}.

	In this work, we propose a joint deblurring and reblurring learning (JDRL) framework for better exploiting misaligned data as well as a new dataset SDD for single image defocus deblurring. 
	As shown in Fig. \ref{fig:overall}, JDRL consists of a deblurring module with adaptive deblurring loss and a spatially invariant reblurring module. 
	First, in deblurring module, we introduce optical flow-based deformation to tolerate misalignment between deblurred result and ground-truth sharp image, and then pixel-wise loss can be imposed to recover sharp textures in deblurred image. 
	To tackle the possible inaccurate deformation, we introduce a calibration mask and cycle deformation.  
	Second, in reblurring module, we suggest that the deblurred result should be spatially consistent with the blurry image. 
	To this end, we propose a reblurring network to generate a blurred image, where isotropic blur kernels are predicted to guarantee the spatial invariance between deblurred, reblurred and blurry images. 
	By adopting JDRL, deblurring networks can learn to adaptively exploit sharp textures from ground-truth image, while keeping spatial consistency with input. 
	JDRL is not coupled with the network architecture, and thus it can be applied to enhance existing deblurring methods.
	
	Moveover, a new dataset SDD with high resolution and diverse contents is established for single image defocus deblurring, where a HUAWEI X2381-VG camera is employed to collect blurry and sharp image pairs by adjusting camera motor {{or aperture size}}. 
	When collecting SDD dataset, we also try to keep alignment between training pairs, but misalignment still exists due to the consumer camera and collecting settings.
	There are two types of misalignment: zoom misalignment and shift misalignment (Fig.~\ref{fig:SDD misalign}). The misalignment in SDD is generally much intenser than DPDD, making it both a good testing bed for JDRL and a benchmark to benefit future research. 
	
	Extensive experiments on DPDD and SDD datasets indicate that JDRL can boost the performance of existing methods. 
	Especially on SDD dataset, the adverse effects of severe misalignment can be significantly relieved by JDRL. 
	%
	When tested on RealDoF dataset, our method also reveals better generalization ability. 
	In summary, the contributions of this paper are three-fold: 
	\begin{itemize}
		\vspace{-0.05in}
		\item 
		A joint deblurring and reblurring learning framework is proposed for better exploiting misaligned training pairs in training single image defocus deblurring networks.
		\item 
		A novel reblurring module is proposed to reconstruct reblurred defocus image, where isotropic blur kernels are predicted to keep the spatial consistency between deblurred result and input blurry image.  
		\item 
		A new dataset SDD with high resolution image pairs and diverse contents is established for single image defocus deblurring, benefiting future research in this field. 
	\end{itemize}
	
	\section{Related Work}\label{sec:related}
	
	\subsection{Defocus Deblurring Methods}
	A popular paradigm for single image defocus deblurring is first estimating a defocus map and then adopting non-blind deconvolution \cite{krishnan2009fast,fish1995blind} to predict the sharp image ~\cite{Shi_2015_CVPR,yi2016lbp,d2016non,karaali2017edge,Lee_2019_CVPR,Park_2017_CVPR}. However, methods of this kind not only lead to excessive time consumption, but also yield unsatistactory performance due to inaccurately estimated defocus maps and over idealized blur kernel models.

	Recent methods mostly employ deep learning networks to ameliorate the defects of traditional methods \cite{abuolaim2020defocus,son2021single,lee2021iterative}. Abuolaim et al. \cite{abuolaim2020defocus} proposed the first deep learning based model to perform defocus deblurring and released the DPDD dataset. IFAN \cite{lee2021iterative} proposed an iterative filter adaptive network, which is trained with an extra reblur loss. 
	Son et al. \cite{son2021single} introduced a framework with a kernel-sharing parallel atrous convolution block for defocus deblurring. 
	{{DRBNet \cite{ruan2022learning} adopts an extra light field dataset in the training stage, which considerably improves the network performance.}}
	In spite of decent performance, existing methods are limited to the vanilla training framework using pixel-wise loss, which can not handle the misalignment between training image pairs. 
	{{In comparison, this paper firstly takes into account the issue of misalignment tolerance, and establishes the JDRL framework to improve the network performance without using extra datasets.}}
	
	\subsection{Defocus Deblurring Datasets}
	The performance of CNN-based methods is largely dependent on the training data. Nevertheless, it can be a labor-intensive project to establish a well qualified defocus deblurring dataset. Although there exist several blur-related datasets, \emph{i.e.}, CUHK \cite{shi2014discriminative}, DUT \cite{zhao2018defocus}, SYNDOF \cite{Lee_2019_CVPR}, {{MC-blur \cite{zhang2021benchmarking}, LFDOF \cite{ruan2022learning} and RealDOF \cite{lee2021iterative} dataset, 
			most of them are mainly built for network testing \cite{lee2021iterative} or blur detection \cite{cun2020defocus,tang2019defusionnet}. DPDD is the only widely adopted real-world training dataset.}}
	%
	A premise of DPDD is the alignment between blurry-sharp image pairs. In this regard, the adjustment of camera aperture size is conducted by a remote signal control to avoid human interference. However, since the image pairs are not captured simultaneously, slight misalignment can inevitably arise. Lee et al. \cite{lee2021iterative} built the RealDOF dataset using a dual-camera system, which can concurrently capture the blurry and sharp images. Following \cite{rim2020real}, a post-processing stage is also adopted to deal with the misalignment. However, as shown in Fig. 1, there still exists misalignment in RealDOF. In this paper, instead of endeavoring to build a perfectly aligned dataset, we propose to (i) make possible the easy acquirement of training data by relaxing the constraint of pairwise alignment and (ii) improve the results of defocus deblurring on existing datasets by taking misalignment into account. Additionally, we propose a new dataset named SDD to validate the effectiveness of our JDRL framework.

	\section{Proposed Method}
	\label{sec:method}
	In this section, we first introduce the pipeline of joint deblurring and reblurring learning (JDRL),  
	then the deblurring module and spatially invariant reblurring module are elaborated, 
	and finally our SDD dataset is established.  
	
	\subsection{Joint Deblurring and Reblurring Learning }
	
	\begin{figure}[!t]
		\centering
		\setlength{\abovecaptionskip}{0pt} 
		\setlength{\belowcaptionskip}{0pt}
		\begin{overpic}[width=0.46\textwidth]{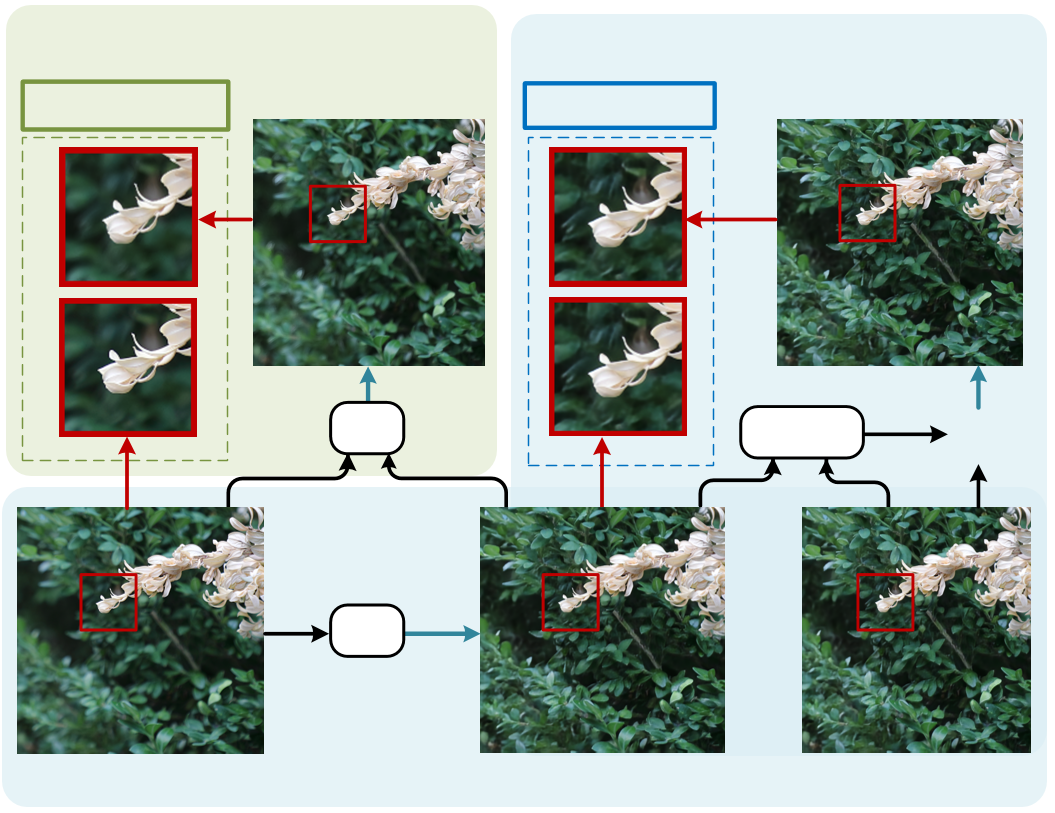}
			\put(11,2){ $\bm{I}_B$}
			\put(31,67){ $\hat{\bm{I}}_B$}
			\put(55,1.5){ $\hat{\bm{I}}$}
			\put(84,2){ $\bm{I}_S$}
			\put(81,67){ $\bm{I}_S^{w}$}
			\put(32,16){ $\mathcal{F}$}
			\put(32,35.2){ $\mathcal{R}$}
			\put(70,35){ $\mathcal{F}_{flow}$}
			\put(9,66){ ${\rm \mathcal{L}_{r}}$}
			\put(55,66){ ${\rm \mathcal{L}_{d}}$}
			\put(89,34.5){ $\mathcal{W}$}
			\put(20,71){ {\textcolor[RGB]{121,149,64}{$\mathcal{M}_{r}$}}}
			\put(68,71){ {\textcolor[RGB]{0,112,192}{$\mathcal{M}_{d}$}}}
		\end{overpic}
		\caption{\small The overall pipeline of JDRL, which consists of two modules, \emph{i.e.}, a deblurring module $\mathcal{M}_{d}$ with adaptive deblurring loss ${\rm \mathcal{L}_{d}}$ to exploit spatially adaptive sharp textures from ground-truth image $\bm{I}_S$, and a spatially invariant reblurring module $\mathcal{M}_{r}$ with reblurring loss ${\rm \mathcal{L}_{r}}$ to keep spatial consistency with input blurry image $\bm{I}_B$.
			During training, deblurred image $\hat{\bm{I}}$ exploits sharp textures from $\bm{I}_S$, while it keeps spatial consistency with $\bm{I}_B$ by generating reblurred image $\hat{\bm{I}}_B$.  }
		\label{fig:overall}
	\end{figure}
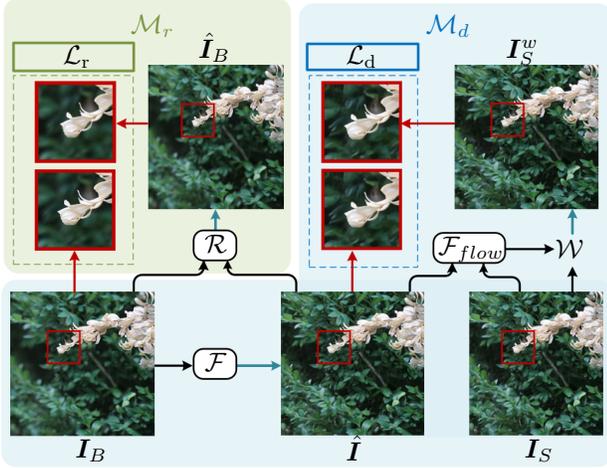
	Given $N$ training pairs $\{\bm{I}_B^n, \bm{I}_S^n\}_{n=1}^N$ where $\bm{I}_B$ is a defocus blurry image and $\bm{I}_S$ is the corresponding sharp image, a defocus deblurring network $\mathcal{F}$ can be learned to predict the deblurred image $\hat{\bm{I}} = \mathcal{F}(\bm{I}_B)$. 
	Since the widely adopted DPDD dataset is designed to keep blurry image and sharp image aligned, existing methods are usually trained with pixel-wise loss, \emph{e.g.}, mean square error (MSE) and $\ell_1$-norm loss \cite{charbonnier1994two}. 
	However, as shown in Fig. \ref{fig:intro}, spatial misalignment between $\bm{I}_B$ and $\bm{I}_S$ is inevitable, forcing deblurring networks to learn an undesirable deformation mapping. 
	Consequently, deformation artifacts may be produced by the deblurring networks, \emph{e.g.}, in Fig. \ref{fig:MDD_distort}, \emph{Brick Lines} are distorted with deformation artifacts. 
	
	In this work, we propose an adaptive learning framework JDRL, where deblurred result can be adaptively supervised by ground-truth image to recover sharp textures while maintaining spatial consistency with blurry image.  
	As shown in Fig. \ref{fig:overall}, JDRL consists of a deblurring module $\mathcal{M}_d$ with adaptive deblurring loss $\mathcal{L}_{d}$ and a spatially invariant reblurring module $\mathcal{M}_{r}$ with reblurring loss $\mathcal{L}_r$. The total loss function $\mathcal{L}$ can be formulated as:
	\begin{equation}\label{eq:jdrl}
	\mathcal{L} = \sum_{n=1}^{N}\mathcal{L}_{d}(\hat{\bm{I}}^n, \bm{I}^n_S) + \alpha \sum_{n=1}^{N}\mathcal{L}_{r}(\mathcal{R}(\hat{\bm{I}}^n, \bm{I}_B^n), \bm{I}^n_B),
	\end{equation}
	where $\alpha$ is a trade-off parameter, and is empirically set as 0.5 in our experiments. 
	First, in $\mathcal{M}_d$, we propose to introduce optical flow-based deformation to tolerate misalignment between $\bm{I}_S$ and $\hat{\bm{I}}$ instead of directly imposing pixel-wise loss. 
	In this way, deblurred result $\hat{\bm{I}}$ is allowed to adaptively learn sharp textures from ground-truth image $\bm{I}_S$, but is not forced to be pixel-wisely consistent with it.  
	To further relieve deformation artifacts possibly caused by optical flow, we introduce calibration mask and cycle deformation.  
	Second, in $\mathcal{M}_r$, we suggest that $\hat{\bm{I}}$ should be spatially consistent with $\bm{I}_B$. 
	To this end, we propose a reblurring network $\mathcal{R}$ for generating reblurred image $\mathcal{R}(\hat{\bm{I}},\bm{I}_B)$ and making it close to $\bm{I}_B$ by minimizing $\mathcal{L}_r$. 
	In reblurring network $\mathcal{R}$, isotropic blur kernels are predicted in polar coordinates, by which spatial consistency between $\hat{\bm{I}}$ and $\bm{I}_B$ can be guaranteed. 

	\begin{figure*}[!t]
		\centering
		\setlength{\abovecaptionskip}{0pt} 
		\setlength{\belowcaptionskip}{0pt}
		\begin{overpic}[width=0.95\textwidth]{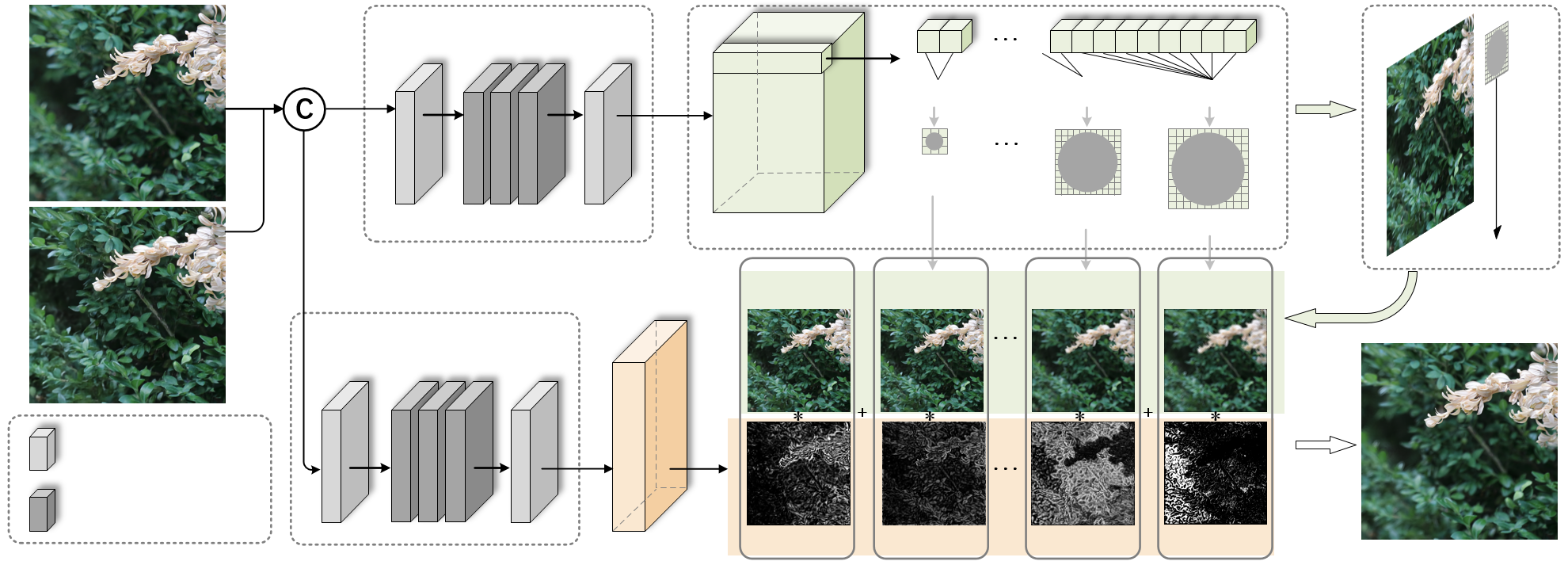}
			\put(-1.2,28){ $\bm{I}_B$}
			\put(-1.2,17){ $\hat{\bm{I}}$}
			\put(5,7){\footnotesize Convolution}
			\put(5,3){\footnotesize Residual block}
			\put(92,15){ $\hat{\bm{I}}_B$}
			\put(50,17){\scriptsize$\hat{\bm{I}}_B^1$}
			\put(58,17){\scriptsize$\hat{\bm{I}}_B^2$}
			\put(67.2,17){\scriptsize$\hat{\bm{I}}_B^{m\!-\!1}$}
			\put(76,17){\scriptsize$\hat{\bm{I}}_B^m$}
			\put(49.3,1){\scriptsize$\bm{W}_1$}
			\put(57.8,1){\scriptsize$\bm{W}_2$}
			\put(66,1){\scriptsize$\bm{W}_{m\!-\!1}$}
			\put(75.5,1){\scriptsize$\bm{W}_m$}
			\put(58.8,29.8){\footnotesize$\mathit{s}_{_\mathrm{2}}$}
			\put(67,29.8){\footnotesize$\mathit{s}_{m\!-\!1}$}
			\put(75.5,29.8){\footnotesize$\mathit{s}_m$}
			\put(58.5,24){\footnotesize$k_2$}
			\put(67.2,22){\footnotesize$k_{m\!-\!1}$}
			\put(76,21.3){\footnotesize$k_m$}
			\put(96.6,32){\rotatebox{270}{\footnotesize Sliding over $\hat{\bm{I}}$}}
			\put(88.8,33.3){ $\hat{\bm{I}}$}
			\put(29.5,33.5){ $\mathcal{R}_{kpn}$}
			\put(25,13.5){ $\mathcal{R}_{wpn}$}
			\put(39.2,0.3){\scriptsize $\bm{W}$}
			\put(48,20.8){\scriptsize $\bm{S}$}
		\end{overpic}
		
		\caption{The structure of reblurring network $\mathcal{R}$. $\mathcal{R}_{kpn}$ predicts isotropic defocus blur kernels, which are then used to generate reblurred images with different blur levels. $\mathcal{R}_{wpn}$ predicts weight maps for integrating reblurred images.}
		\label{fig:reblurnet}
	\end{figure*}
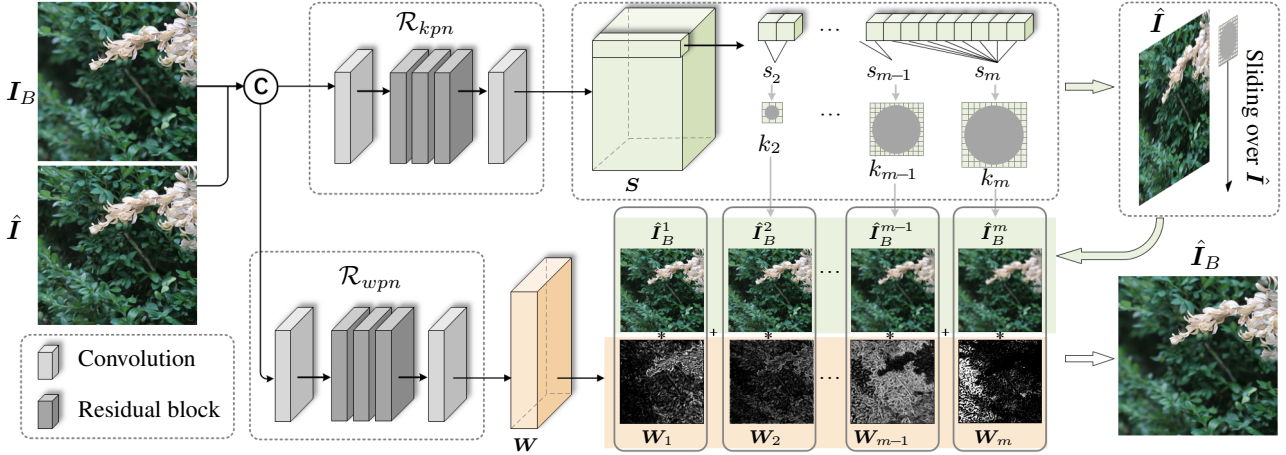
	
	In summary, JDRL allows deblurring network $\mathcal{F}$ to exploit spatially adaptive sharp textures from ground-truth image and keep spatial consistency with blurry image.  
	During inference, deblurring network $\mathcal{F}$ can directly take blurry image as input and predict latent sharp image. 
	Our JDRL is not coupled with deblurring network $\mathcal{F}$, and thus it can be applied to boost existing deblurring network, \emph{e.g.}, IFAN \cite{lee2021iterative} and MPRNet \cite{zhang2019deep}. 
	In experiments, we also provide a baseline deblurring network, where a plain UNet \cite{ronneberger2015u} is adopted to act as $\mathcal{F}$. 

	\subsection{Deblurring Module $\mathcal{M}_d$}
	\label{sec:deblur module}
	
	To exploit spatially adaptive sharp textures from $\bm{I}_S$, we introduce an optical flow-based deformation in $\mathcal{M}_d$, by which possible misalignment between $\bm{I}_S$ and $\hat{\bm{I}}$ can be tolerated. 
	In particular, an optical flow estimating network $\mathcal{F}_{flow}$ \cite{sun2018pwc} is employed to estimate the optical flow $\bm{\Phi}_{\bm{I}_{S}\rightarrow\hat{\bm{I}}}$ from $\bm{I}_S$ to $\hat{\bm{I}}$, 
	\begin{equation}\label{eq:flow}
	\bm{\Phi}_{\bm{I}_{S}\rightarrow\hat{\bm{I}}} = \mathcal{F}_{flow} (\bm{I}_S, \hat{\bm{I}}), 
	\end{equation}
	and $\bm{I}_S$ is deformed towards $\bm{\hat{I}}$ using estimated optical flow:
	\begin{equation}
	\bm{I}_S^{w} = \mathcal{W} (\bm{I}_S, \bm{\Phi}_{\bm{I}_{S}\rightarrow\hat{\bm{I}}}),
	\end{equation}
	where $\mathcal{W}$ denotes linear deformation operation \cite{sun2018pwc}. 
	Here we adopt a calibration mask $\bm{M}$ to filter out the regions where the optical flow is inaccurately estimated. 
	Specifically, the average value $\overline{\bm{\Phi}}$ of optical flow is first computed, and then the calibration mask can be defined as:
	\begin{equation}\label{eq:mask}
	\bm{M}_{\bm{I}_{S}\rightarrow\hat{\bm{I}}} = [(1-\lambda) \times \overline{\bm{\Phi}}_{\bm{I}_{S}\rightarrow\hat{\bm{I}}}<\bm{\Phi}_{\bm{I}_{S}\rightarrow\hat{\bm{I}}}<(1+\lambda)\times \overline{\bm{\Phi}}_{\bm{I}_{S}\rightarrow\hat{\bm{I}}}],
	\end{equation}
	where the value of $\bm{M}$ is 1 if the condition in $[\cdot]$ is satisfied, and otherwise 0. 
	The reason why Eq.~\eqref{eq:mask} works is that the amount of misalignment between $\bm{I}_S$ and $\hat{\bm{I}}$ only varies slightly across different spatial positions, so that the inaccurate estimation of optical flow can be identified by the abnormality in its magnitude. 
	Furthermore, we introduce a cycle deformation, \emph{i.e.}, the optical flow $\bm{\Phi}_{\hat{\bm{I}}\rightarrow \bm{I}_{S}}$ can be computed in the reverse order and the reverse deformation can be deployed, by which the robustness of deformation process can be enhanced. 
	
	Finally, the adaptive deblurring loss can be calculated based on Charbonnier loss~\cite{charbonnier1994two}:
	\begin{equation}\label{eq:deblurloss}
	\begin{split}
	{ \mathcal{L}_{d}}=\sum_{n=1}^{N}\bigl(\sqrt{\parallel \bm{M}^n_{\bm{I}_{S}\rightarrow\hat{\bm{I}}}*(\bm{I}_S^{w,n} - \hat{\bm{I}}^{n}) \parallel^{2} + \varepsilon^2} + \\ \sqrt{\parallel \bm{M}^n_{\hat{\bm{I}}\rightarrow \bm{I}_{S}}*(\hat{\bm{I}}^{w,n} - \bm{I}_S^n) \parallel^{2} + \varepsilon^2}\bigr),
	\end{split}
	\end{equation}
	where where $*$ is element-wise product, and $\varepsilon$ is empirically set as $1\times 10^{-3}$ in all the experiments.
	
	\subsection{Reblurring Module $\mathcal{M}_r$}
	
	By only using $\mathcal{L}_d$, deblurring network $\mathcal{F}$ would still learn spatial deformation. 
	Thus, we propose a reblurring module $\mathcal{M}_r$ to keep $\hat{\bm{I}}$ spatially consistent with $\bm{I}_B$.  
	A spatially invariant reblurring network $\mathcal{R}$ is deployed to reblur $\hat{\bm{I}}$ to $\hat{\bm{I}}_B$ using predicted isotropic blurring kernels, and then $\hat{\bm{I}}_B$ is supervised by $\bm{I}_B$ using pixel-wise loss. 
	As shown in Fig.~\ref{fig:reblurnet}, reblurring network $\mathcal{R}$ consists of a Kernel Prediction Network ($\mathcal{R}_{kpn}$) and a Weight Prediction Network ($\mathcal{R}_{wpn}$). The concatenation of $\hat{\bm{I}}$ and $\bm{I}_B$ is taken as the input of both $\mathcal{R}_{kpn}$ and $\mathcal{R}_{wpn}$.
	One may note that a reblurring loss is adopted in IFAN \cite{lee2021iterative}, in which  reblurring filters are implicitly predicted. 
	In contrast, we explicitlly predict isotropic blur kernels so as to guarantee the spatial invariance between the deblurred image and its reblurred version. 
	
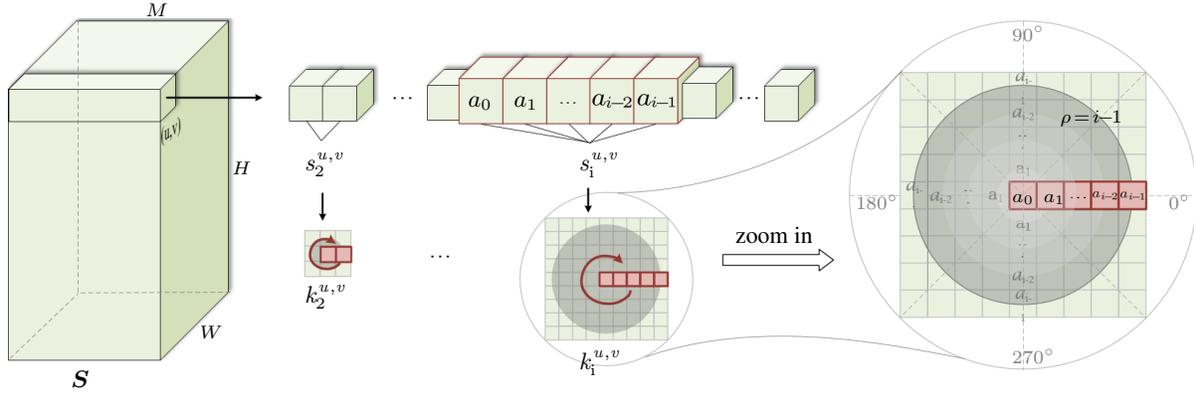
\begin{figure*}[t]
		\centering
		\begin{overpic}[width=0.9\textwidth]{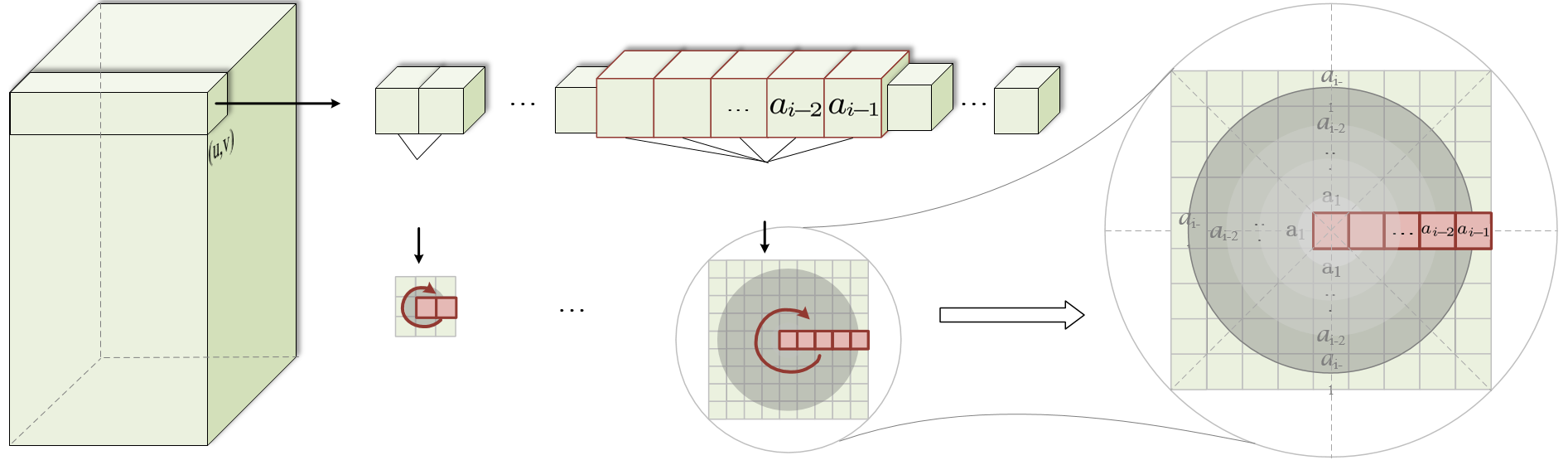}
			\put(5.8,-1){\small $\bm{S}$}
			\put(19.2,16.7){\scriptsize $H$}	
			\put(16.5,3){\scriptsize $W$}
			\put(12,29.8){\scriptsize $M$}	
			\put(38.8,22.3){\small $a_0$}
			\put(42.6,22.3){\small $a_1$}
			\put(84,14.4){\scriptsize $a_0$}
			\put(86.6,14.4){\scriptsize $a_1$}
			\put(88,21){\scriptsize $\rho\!=\!{i\!\!-\!\!1}$}
			\put(97,13.8){\scriptsize {\textcolor[RGB]{127,127,127}{$0^\circ$}}}
			\put(84,27.8){\scriptsize {\textcolor[RGB]{127,127,127}{$90^\circ$}}}
			\put(71,13.8){\scriptsize {\textcolor[RGB]{127,127,127}{$180^\circ$}}}
			\put(84,1){\scriptsize {\textcolor[RGB]{127,127,127}{$270^\circ$}}}
			\put(61,11){\small zoom in}	
			\put(25.2,17){\footnotesize $\mathit{s}_{\mathrm{2}}^\mathit{u,v}$}	
			\put(48,17){\footnotesize $\mathit{s}_{\mathrm{i}}^\mathit{u,v}$}	
			\put(25.2,6){\footnotesize $\mathit{k}_2^\mathit{u,v}$}	
			\put(48,0.5){\footnotesize $\mathit{k}_{\mathrm{i}}^\mathit{u,v}$}
			
		\end{overpic}
		\caption{Illustration of how isotropic blur kernels are generated in polar coordinates. 
			For a feature vector located at $(u,v)$, it is first split into a set of kernel seeds $\{s_i^{u,v}\}$ and then converted to blur kernels $\{k_i^{u,v}\}$. For $k_i^{u,v}$, the value of each element is interpolated using polar coordinates in terms of the distance between this element and the center of $k_i^{u,v}$.}
		\label{fig:isotropic}
	\end{figure*}

	\textbf{Kernel Prediction Network} $\bm{\mathcal{R}_{kpn}}$: Our aim is to predict blur kernel for each pixel. 
	Considering the isotropic property of defocus blur kernel, $\bm{\mathcal{R}_{kpn}}$ first predicts kernel seeds. 
	The functionality of $\mathcal{R}_{kpn}$ is described as
	\begin{equation}
	\bm{S} = \mathcal{R}_{kpn}(\hat{\bm{I}}, \bm{I}_B),
	\end{equation}
	where the concatenation of $\hat{\bm{I}}$ and $\bm{I}_B$ is taken as input. 
	$\hat{\bm{I}}$ and $\bm{I}_B$ are of size $3\times H\times W$, and the output $\bm{S}$ is a $ M \times H \times W $ feature volume. For each position $(u,v)$, the corresponding $ M \times 1 \times 1 $ feature vector is split into a set of kernel seeds $\{s_i^{u,v} \ |\  i=2,3,...,m,\ \sum_{i=2}^{m}i = M\}$, which are then used to generate a set of isotropic kernels. Each kernel $k_i^{u,v}$ is a single-channel map of size $(2i-1)\times(2i-1)$. 
	The kernel generation process is illustrated in Fig.~\ref{fig:isotropic}.
	We take the kernel seed $s_i^{u,v}=[a_0, a_1, ..., a_{i-1}]^{\rm T}$ as example and explain in detail how the corresponding kernel $k_i^{u,v}$ is generated. For a single element of $k_i^{u,v}$, its value is determined by its distance from the center of $k_i^{u,v}$ using interpolation in polar coordinates. Specifically, we first represent the elements of $k_i^{u,v}$ with the form $(\rho,\theta)$ in polar coordinates. Then the kernel values are calculated as follows:
	\begin{spacing}{1}
		\begin{equation}
		\begin{split}
		k_i^{u,v}(\rho,\theta) = 
		\left\{
		\begin{aligned}
		& a_{\rho}\ ,\ {\rm if} \ \rho\le i-1 \ {\rm and}\ \rho\  {\rm is} \ {\rm integer}\\
		& 0\ ,\ {\rm if}\ \rho>i-1\\
		& \dfrac{\rho - \lceil\rho\rceil}{\lfloor\rho\rfloor- \lceil\rho\rceil}a_{\lfloor\rho\rfloor} + \dfrac{\rho - \lfloor\rho\rfloor}{\lceil\rho\rceil- \lfloor\rho\rfloor}a_{\lceil\rho\rceil},\ {\rm else}
		\end{aligned}
		\right.
		\end{split}
		\end{equation}
	\end{spacing}
	\noindent where $\lfloor\rfloor$ and $\lceil\rceil$ denote the floor and ceiling operations. The calculated kernel values are then normalized using a \texttt{Softmax} function, such that $\sum_{\rho,\theta} k_i^{u,v}(\rho,\theta) = 1, k_i^{u,v}(\rho,\theta)>0$. By these operations, we can get $m-1$ isotropic defocus blur kernels of sizes $\{3\times 3, 5\times 5, ..., (2m-1)\times (2m-1)\}$, accounting for different positions of $\hat{\bm{I}}$. Next, $m-1$ blurry images $\{\hat{\bm{I}}_B^2, ..., \hat{\bm{I}}_B^{m}\}$ in different blur levels can be obtained by convolving $\hat{\bm{I}}$ with corresponding blur kernels. Note that we also take no-blur image $\hat{\bm{I}}_B^1$ into account, which is actually the deblurred image $\hat{\bm{I}}$.
	
	\textbf{Weight Prediction Network }$\bm{\mathcal{R}_{wpn}}$:
	Since the amount of blur is spatially variant across a blurry image, it is necessary to generate weight maps to integrate the reblurred images. 
	The functionality of $\mathcal{R}_{wpn}$ can be formulated as:
	\vspace{-0.1cm}
	\begin{equation}
	\bm{W} = \mathcal{R}_{wpn}(\hat{\bm{I}}, \bm{I}_B),
	\end{equation}
	where $\bm{W}$ is a $m\times H\times W$ feature volume which is then split into $m$ single-channel weight maps \{$\bm{W}_1, \bm{W}_2, ..., \bm{W}_m$\}.
	The weight maps are normalized by the \texttt{Softmax} function such that the values at the same position $(u,v)$ add up to 1, \emph{i.e.}, $\sum_i \bm{W}_i^{u,v} = 1$, $\bm{W}_i^{u,v}\ge0$. 
	The reblurred image can be reconstructed as:
	\begin{equation}
	\hat{\bm{I}}_B = \sum_{i=1}^{m}\bm{W}_i*\hat{\bm{I}}_B^i.
	\end{equation}
	The reblurring loss in $\mathcal{M}_{R}$ is defined as:
	\begin{equation}
	{\mathcal{L}_{r}}=\sum_{n=1}^{N}(\sqrt{\parallel \bm{I}_B^n - \hat{\bm{I}}_B^n) \parallel^{2} + \varepsilon^2},
	\end{equation}
	where $\varepsilon=1\times 10^{-3}$ in experiments. 
	We note that the spatial consistency between $\hat{\bm{I}}$, $\hat{\bm{I}}_B$ and ${\bm{I}}_B$ can be guaranteed due to the isotropic property of predicted blur kernels (\emph{referring to Fig. 4 in the suppl.}).
	As for the network architectures of $\mathcal{R}_{kpn}$ and $\mathcal{R}_{wpn}$, we adopt simple networks with basic convolutional layers and residual blocks as shown in Fig. \ref{fig:reblurnet}. 
	
	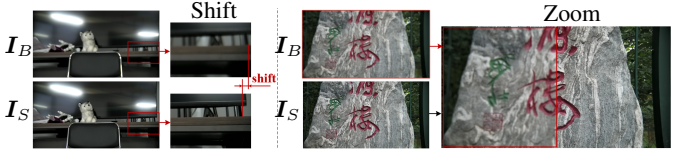
\begin{figure}[t]
		\centering
		\setlength{\abovecaptionskip}{0pt} 
		\setlength{\belowcaptionskip}{0pt}
		\begin{overpic}[width=0.48\textwidth]{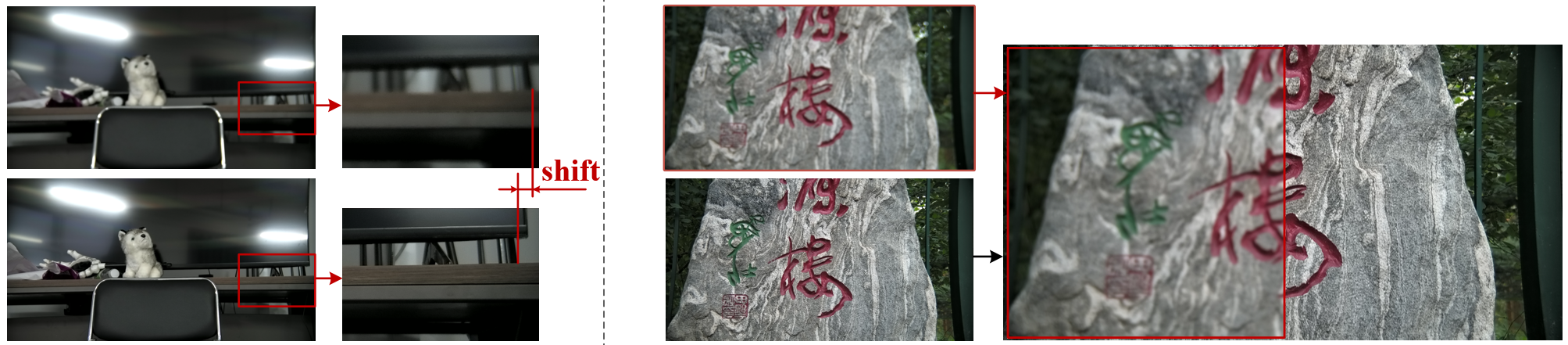}
			\put(-4,15){\small $\bm{I}_B$}
			\put(-4,5){\small $\bm{I}_S$}
			\put(38,5){\small $\bm{I}_S$}
			\put(38,15){\small $\bm{I}_B$}
			
			\put(25,20.5){\small Shift}
			\put(80,20){\small Zoom}
		\end{overpic}
		\caption{\small Illustration of misalignments in SDD dataset.}
		\label{fig:SDD misalign}
	\end{figure}
	\begin{figure}[!t]
		\centering
		\setlength{\abovecaptionskip}{0pt} 
		\setlength{\belowcaptionskip}{0pt}
		\begin{overpic}[width=0.48\textwidth]{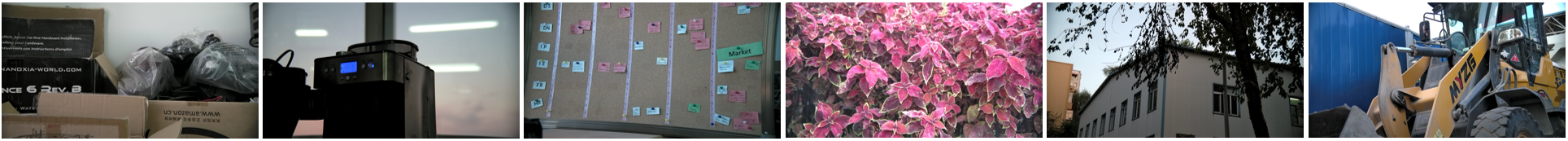}
			
		\end{overpic}
		\caption{\small Sample images from our SDD dataset. 
		}
		\label{fig:SDD}
	\end{figure}
	\begin{figure*}[t]
		\small
		\centering
		\setlength{\abovecaptionskip}{0pt} 
		\setlength{\belowcaptionskip}{0pt}
		\begin{tabular}{cc}
			\footnotesize
			\begin{adjustbox}{valign=t}
				\begin{tabular}{cccccc}			
					\includegraphics[width=0.135\textwidth]{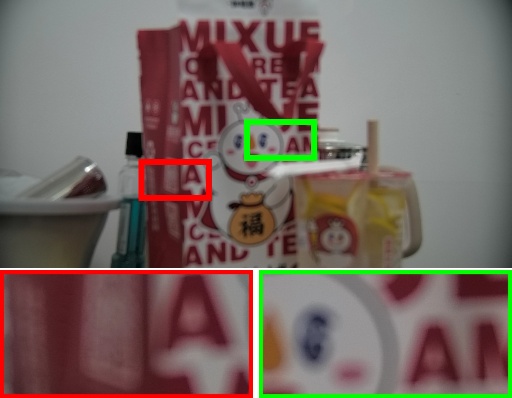}&\hspace{-4mm}
					\includegraphics[width=0.135\textwidth]{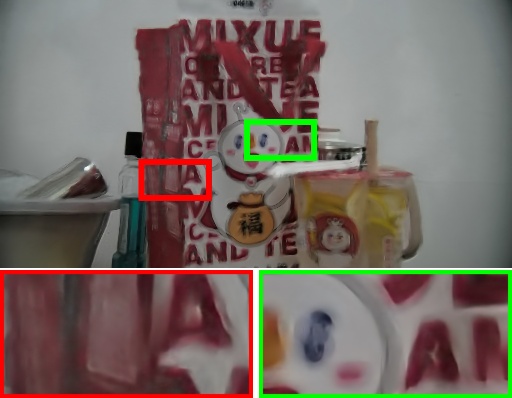}&\hspace{-4mm}
					\includegraphics[width=0.135\textwidth]{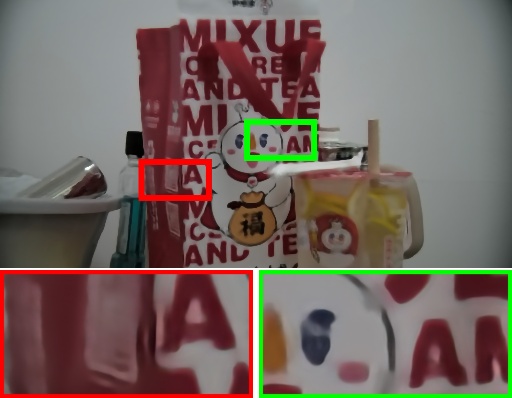}&\hspace{-4mm}
					\includegraphics[width=0.135\textwidth]{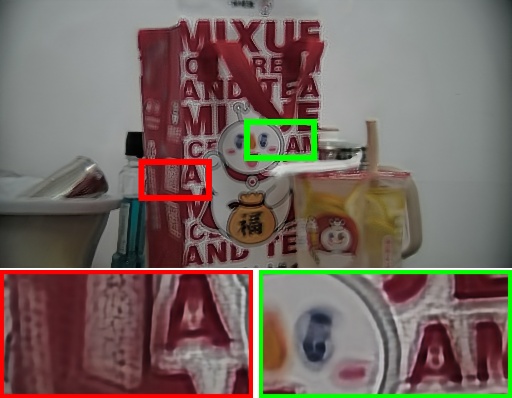}&\hspace{-4mm}
					\includegraphics[width=0.135\textwidth]{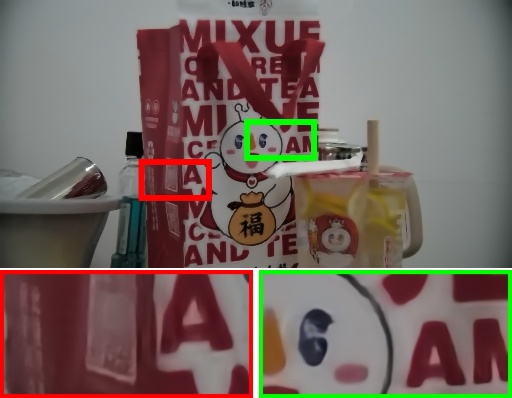}&\hspace{-4mm}
					\includegraphics[width=0.135\textwidth]{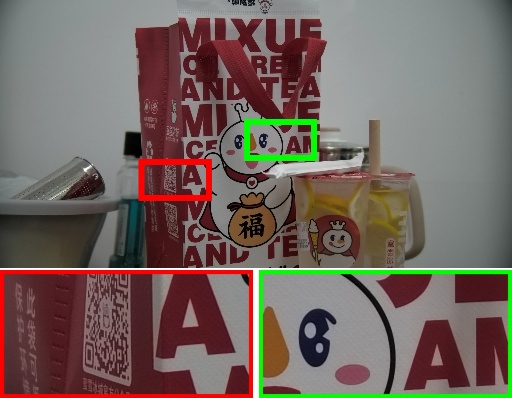}
					\\
					\includegraphics[width=0.135\textwidth]{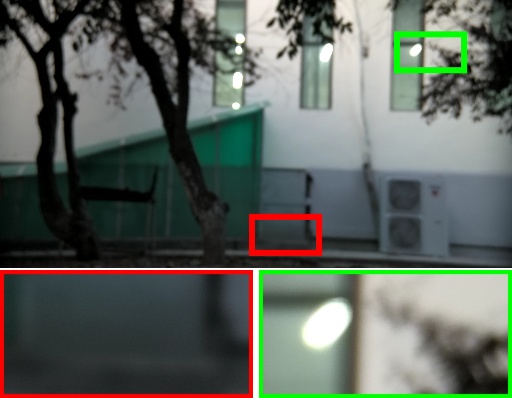}&\hspace{-4mm}
					\includegraphics[width=0.135\textwidth]{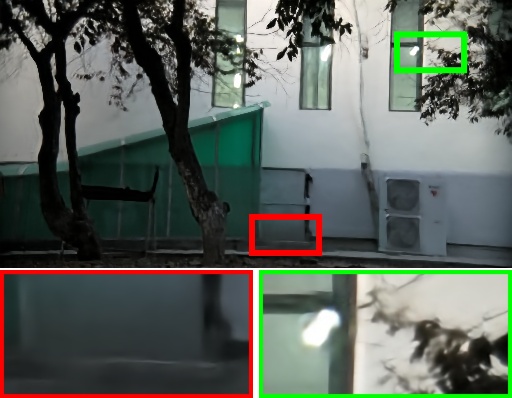}&\hspace{-4.2mm}
					\includegraphics[width=0.135\textwidth]{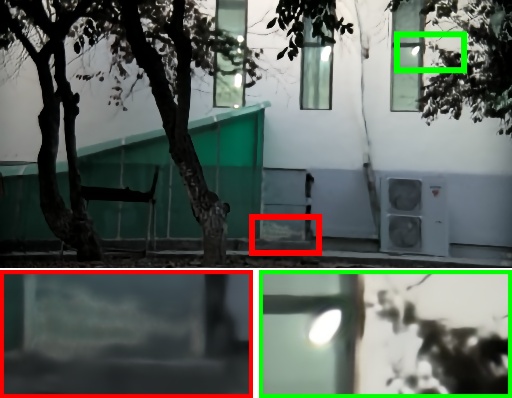}&\hspace{-4mm}
					\includegraphics[width=0.135\textwidth]{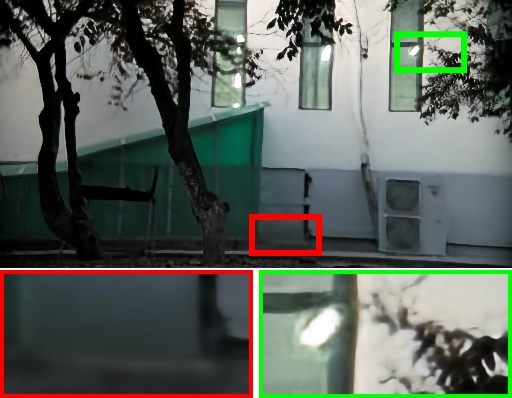}&\hspace{-4mm}
					\includegraphics[width=0.135\textwidth]{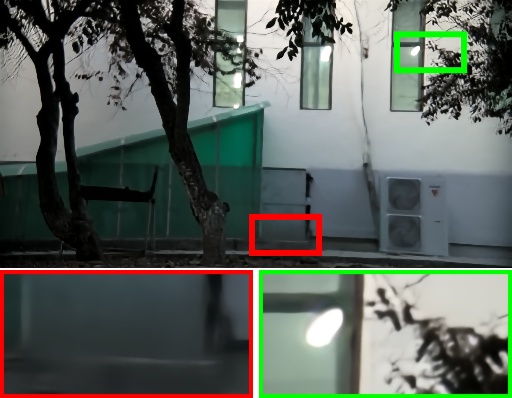}&\hspace{-4mm}
					\includegraphics[width=0.135\textwidth]{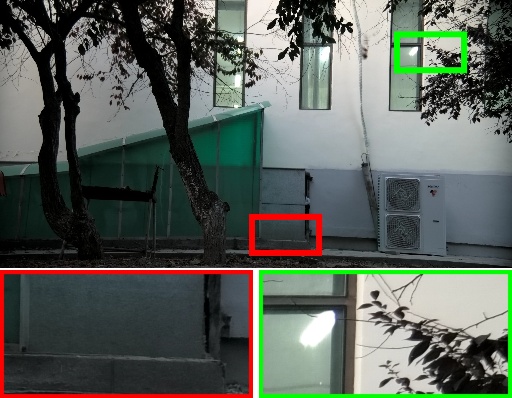}
					\\
					\includegraphics[width=0.135\textwidth]{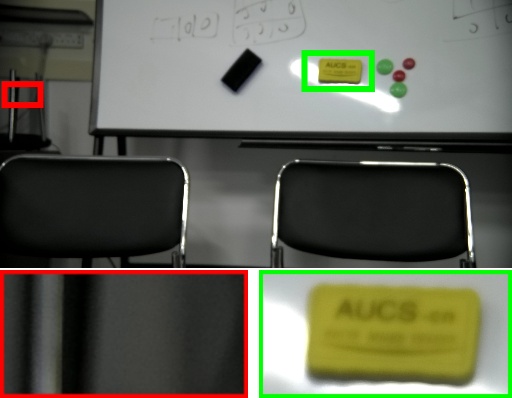}&\hspace{-4mm}
					\includegraphics[width=0.135\textwidth]{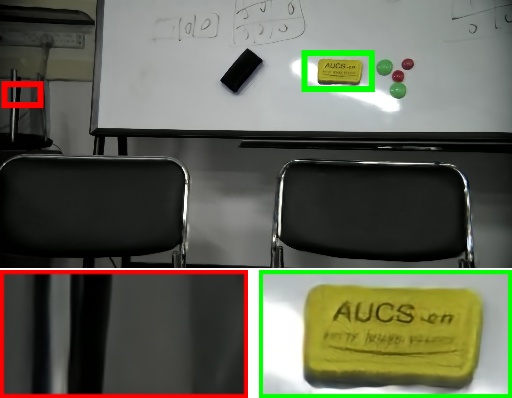}&\hspace{-4mm}
					\includegraphics[width=0.135\textwidth]{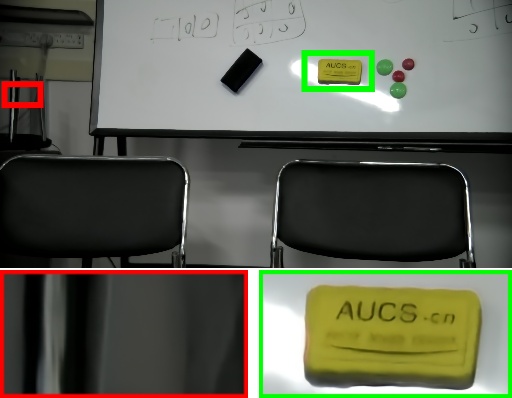}&\hspace{-4mm}
					\includegraphics[width=0.135\textwidth]{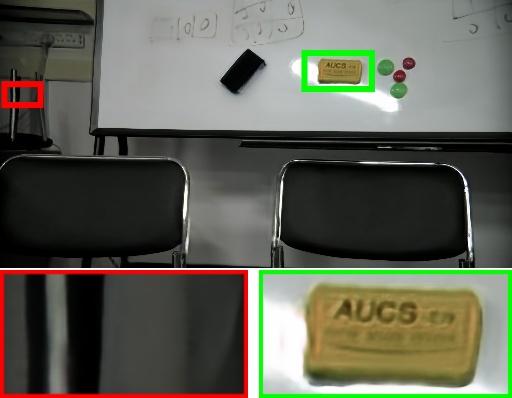}&\hspace{-4mm}
					\includegraphics[width=0.135\textwidth]{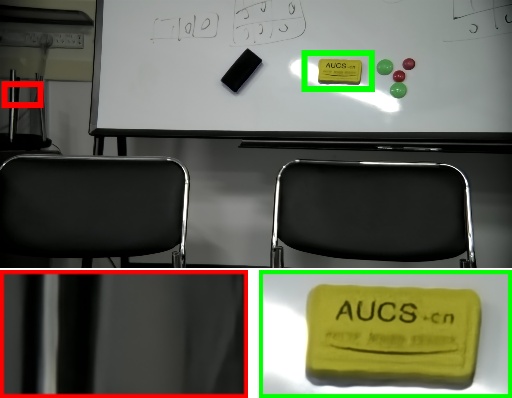}&\hspace{-4mm}
					\includegraphics[width=0.135\textwidth]{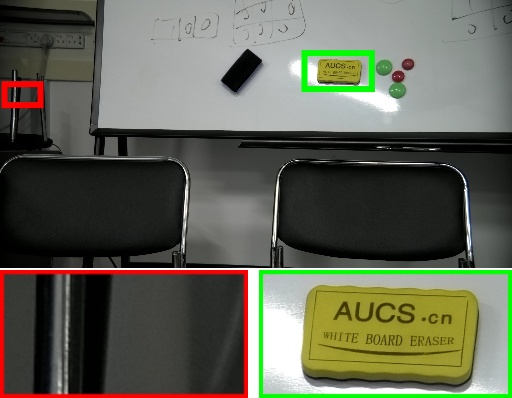}
					\\
					Input&\hspace{-4mm}
					DMPHN&\hspace{-4mm}
					MPRNet& \hspace{-4mm}
					DPDNet$_{S}$&\hspace{-4mm}
					MPRNet*&\hspace{-4mm}
					GT
					\\
				\end{tabular}
			\end{adjustbox}
			
		\end{tabular}
		\caption{\small Visual comparison between different methods on SDD dataset. With the proposed JDRL framework applied, MPRNet* performs visually better than the original MPRNet in terms of textures and structures.}
		\label{MDD_fig}
	\end{figure*}
	
	\subsection{New Dataset for Single Image Defocus Deblurring}
	Considering that DPDD is originally developed for dual-pixel defocus deblurring, we establish a new dataset named SDD for single image defocuse deblurring. %
	In particular, a HUAWEI X2381-VG camera is employed. The DoF of this camera can be adjusted by vertically moving the lens or by changing its aperture size. 
	We totally capture 150 blurry/sharp image pairs, each of which is of resolution $4096\times 2160$.
	These image pairs are split into a training set with 115 image pairs and a testing set with 35 image pairs. Similar to \cite{abuolaim2020defocus}, the training image pairs are resized and cropped into 4,830 patches of size $512\times 512$.
	The SDD dataset covers a variety of indoor and outdoor scenes. There are 50 indoor scenes and 65 outdoor scenes in the training set, and 11 indoor scenes and 24 outdoor scenes in the testing set.
	Sample images from SDD dataset are exhibited in Fig.~\ref{fig:SDD}.
	
	When collecting our SDD dataset, we also try to keep alignment between training pairs, but misalignments are inevitable for the consumer camera and collecting settings.
	{{For example, due to the fact that adjusting the aperture size will cause illumination change between blur and sharp images, we adjust DoF mainly by moving the lens back and forth. Only when the moving range is not enough for the desired DoF, we then change the aperture size, hence limiting the illumination change to small amounts.
			As a result, misalignment is inevitable during the capturing process. 
	}}
	In practice, two types of misalignment would come up in tandem with the establishment of SDD dataset: zoom misalignment and shift misalignment, which are caused by the vertical and horizontal movement of camera lens, respectively (see Fig.~\ref{fig:SDD misalign}).
	Misalignments in SDD dataset are usually more severe than DPDD dataset, and SDD is a testing bed for validating the effectiveness of our JDRL. 
	
	\section{Experiments}
	\label{sec:experiment}
	\subsection{Implementation Details}
	The source code, SDD dataset and the supplementary file
	{{are available at {\url{https://github.com/liyucs/JDRL}}.}
		The source code is implemented in Pytorch~\cite{paszke2019pytorch}, and also we provide an implementation in HUAWEI Mindspore, which is available at \url{https://github.com/Hunter-Will/JDRL-mindspore}}.  
	%
	
	\textbf{Configuration:} During training, $\lambda$ is set as 0.35 for generating the calibration masks, and the maximal radius of blur kernels $m$ is set as 8. The baseline deblurring network $\mathcal{F}$ of JDRL is implemented as a vanilla U-Net structure, the details of which are described in suppl. 
	
	\noindent\textbf{Datasets:} The proposed JDRL framework is evaluated on three datasets: SDD, DPDD and RealDOF, where RealDOF is only used for testing the models trained on the DPDD dataset. The taining patches are all of size $512\times512$. In the DPDD dataset, there are 350/76/74 image triplets for training/testing/validation, respectively. Each blurry image comes along with a sharp image. We use 350 blurry-sharp image pairs for training and 76 blurry images for testing. 
	
	\noindent\textbf{Network training:} 
	During the initial stage of training, $\hat{\bm{I}}$ is of low quality, hence the predicted optical flow $\bm{\Phi}_{\bm{I}_{S}\rightarrow\hat{\bm{I}}}$ and $\bm{\Phi}_{\hat{\bm{I}}\rightarrow \bm{I}_{S}}$ are hardly informative. In this regard, $\bm{\Phi}_{\bm{I}_{S}\rightarrow\hat{\bm{I}}}$ is calculated by $\mathcal{F}_{flow} (\bm{I}_S, \bm{I}_B)$ at the beginning of training. After an initialization stage of $N$ training epochs, the optical flow is instead obtained by Eq.~\eqref{eq:flow}. Empirically, $N$ is set as 15 in our experiments to ensure the quality of $\hat{\bm{I}}$.
	The parameters of JDRL are initialized using the strategy proposed by He et al. \cite{he2015delving}, and are optimized using the Adam optimizer \cite{kingma2014adam}. The learning rate is initialized as 2$\times$10$^{-5}$ and is halved every 60 epochs. The entire training stage ends with 100 epochs. 
	
	\begin{figure}[!t]
		\small
		\centering
		\setlength{\abovecaptionskip}{0pt} 
		\setlength{\belowcaptionskip}{0pt}
		\footnotesize
		\begin{tabular}{c}	
			
			\includegraphics[width=0.47\textwidth]{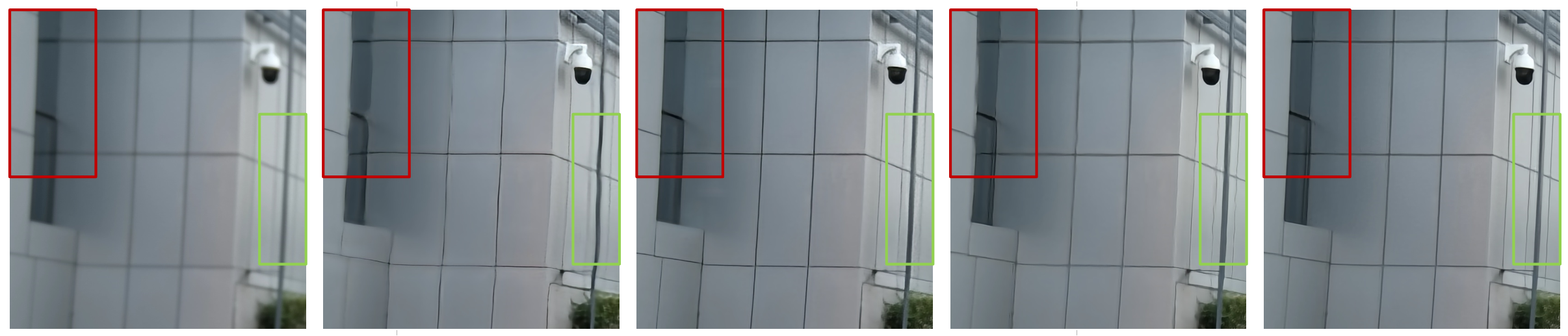}
			\\
		\end{tabular}
		\begin{tabular}{ccccc}	
			\hspace{0.4cm}	
			\small{Input}&\hspace{0.6cm}
			\small{UNet}&\hspace{0.4cm}
			\small{UNet*}&\hspace{0.1cm}
			\small{MPRNet}&\hspace{0cm}
			\small{MPRNet*}\hspace{0.2cm}
			\\
		\end{tabular}
		\caption{\small Deformation artifacts by UNet and MPRNet, which can be alleviated by training with JDRL (UNet* and MPRNet*). }
		\label{fig:MDD_distort}
	\end{figure}
	
	\subsection{Experiments on SDD dataset}
	The proposed method is compared with SOTA single image defocus deblurring methods including JNB~\cite{Shi_2015_CVPR}, EBDB~\cite{karaali2017edge}, DMENet~\cite{Lee_2019_CVPR} and DPDNet$_{S}$~\cite{abuolaim2020defocus}. Since the former three methods are originally designed for defocus map estimation, we adopt non-blind deconvolution \cite{krishnan2009fast} to get the final estimated sharp images. Additionally, we retrain two SOTA motion deblurring methods DMPHN \cite{zamir2021multi} and MPRNet \cite{zhang2019deep} on the SDD dataset for extra comparison. Son et al. \cite{son2021single} and IFAN \cite{lee2021iterative} are excluded from comparison because Son et al. has not release the training code and IFAN requires additional dual-pixel data during training.
	As introduced before, the proposed JDRL framework can be applied to different existing methods. Therefore, we also replace the baseline deblurring network UNet with MPRNet, which achieves the best performance among all methods without using JDRL, for further comparison. 
	
	\begin{table}[t]
		\centering
		\caption{\small{Quantitive comparison between different methods on SDD dataset. `UNet' and `UNet*' refer to the baseline deblurring network without using and using the  JDRL framework, respectively. `MPRNet*' refers to MPRNet trained by JDRL framework.}}
		\label{tab:MDD}
		\footnotesize
		\begin{tabular}{c|cccc}
			\toprule
			Method  &  \ PSNR$\uparrow$ \  & \ SSIM$\uparrow$\  &\  MAE$\downarrow$\  &\ LPIPS$\downarrow$\ \\
			\midrule
			JNB &  24.06 & 0.720 & 0.0400 & 0.443 \\
			EBDB & 24.06 & 0.720 & 0.0398 & 0.442 \\
			DMENet &  24.02 & 0.732 & 0.0395 & 0.426 \\
			\midrule
			\ \ \ \ DMPHN \ \ \ \ \ & \ 25.00 \ & \ 0.769 \ & \ \ \ 0.0351 \ \ & \ \ 0.326  \ \ \\
			DPDNet$_{S}$ & 24.81 & 0.760 & 0.0364 & 0.343 \\	
			MPRNet & 26.28 & 0.796 & 0.0293 & 0.302 \\
			UNet & 24.62 & 0.758 & 0.0360 & 0.344 \\	
			\midrule
			UNet*(Ours) & 25.82 & 0.783 & 0.0320 & 0.305 \\	
			\ MPRNet*(Ours) \  & \textbf{26.88} &\textbf{0.810} & \textbf{0.0283} & \textbf{0.265} \\		
			\bottomrule
			\hline
		\end{tabular}
	\end{table}

	For evaluation, we use the testing set of SDD, which contains 35 blurry-sharp image pairs. Since the blurry images and ground-truth sharp images of SDD are misaligned, the evaluation metrics such as PSNR can not be directly calculated. In this regard, we use a pretrained PWCNet to compute the optical flow between $\hat{\bm{I}}$ and $\bm{I}_S$, and then deform $\bm{I}_S$ towards $\hat{\bm{I}}$. The evaluation metrics are then calculated between $\hat{\bm{I}}$ and the deformed ground-truth $\bm{I}_S^w$. Four metrics are adopted to quantitatively evaluate the compared methods: Peak Signal-to-Noise Ratio (PSNR), Structural Similiarity (SSIM)~\cite{wang2004image}, Mean Absolute Error (MAE), and Learned Perceptual Image Patch Similarity (LPIPS)~\cite{zhang2018unreasonable}.

	Table~1 reports the evaluation results on SDD test set. By utilizing JDRL framework, both UNet* and MPRNet* outperform UNet and MPRNet respectively by a large margin. 
	Moreover, it can be seen from Fig. 7 that MPRNet* performs favorably against the others in terms of both the restoration of image texture and the preservation of objects' shapes. This is because existing methods are only trained with vanilla pixel-wise loss, by which spatial deformation is also learned during training. Taking the last row of Fig.~\ref{MDD_fig} as example, the silver cylinder within the red bounding box is heavily distorted in all methods except for MPRNet*. Another example is given in Fig.~\ref{fig:MDD_distort}, which shows that the issue of line distortion can be obviously alleviated by using JDRL.

	\begin{table}[t]
		\centering
		\setlength{\abovecaptionskip}{0pt} 
		\setlength{\belowcaptionskip}{0pt}
		\tabcolsep=0.13cm
		\caption{\small{Quantitative comparison on DPDD. IFAN* and MPRNet* refer to IFAN and MPRNet trained by JDRL, respectively.}}
		\hspace{-1.em}
		\label{DPDD_tab}
		\footnotesize
		\begin{tabular}{c|c|c}
			\toprule
			&GT & Deformed GT \\
			\midrule
			Method &  PSNR/SSIM/MAE/LPIPS& PSNR/SSIM/MAE/LPIPS\\
			\midrule
			JNB& 22.13/0.676/0.0560/0.480 & 21.30/0.692/0.0620/0.458\\
			EBDB& 23.19/0.713/0.0510/0.419 &  22.66/0.737/0.0540/0.397\\
			DMENet&  23.33/0.715/0.0506/0.411 &  22.89/0.740/0.0536/0.388\\
			\midrule
			DPDNet$_{S}$& 24.39/0.750/0.0435/0.277 &  24.48/0.778/0.0435/0.262\\	
			Son et al. & 25.22/0.774/0.0403/0.227 & 25.49/0.807/0.0398/0.212\\
			IFAN &25.37/0.789/0.0394/0.217 &  25.86/0.825/0.0374/0.192\\
			UNet & 24.60/0.757/0.0426/0.281 & 24.63/0.785/0.0424/0.261 \\
			MPRNet& 25.72/0.791/0.0385/0.235 & 26.03/0.820/0.0372/0.214\\
			\midrule
			IFAN* & 25.54/0.788/0.0389/0.207 &  26.09/0.826/0.0372/0.177\\	
			UNet* & 24.73/0.762/0.0422/0.273 & 24.97/0.794/0.0413/0.248\\
			MPRNet* & \textbf{25.73}/\textbf{0.792}/\textbf{0.0384}/\textbf{0.232} &  \textbf{26.21}/\textbf{0.825}/\textbf{0.0365}/\textbf{0.207}\\		
			\bottomrule
			\hline
		\end{tabular}
	\end{table}
	
	\begin{table}[t]
		\centering
		\setlength{\abovecaptionskip}{0pt} 
		\setlength{\belowcaptionskip}{0pt}
		\tabcolsep=0.13cm
		\caption{\small{Quantitative comparison on RealDOF. All the methods are trained on DPDD dataset, since RealDOF has no training set.}}
		\hspace{-1.em}
		\label{realdof_tab}
		\footnotesize
		\begin{tabular}{c|c|c}
			\toprule
			&{GT} & {Deformed GT} \\
			\midrule
			Method &  PSNR/SSIM/MAE/LPIPS&  PSNR/SSIM/MAE/LPIPS\\
			\midrule
			JNB&  22.33/0.635/0.0514/0.598 &  21.02/0.651/0.0607/0.590\\
			EBDB& 22.38/0.640/0.0508/0.590 &  21.37/0.659/0.0581/0.579 \\
			DMENet&  22.40/0.639/0.0508/0.593 &  21.38/0.658/0.0579/0.581\\
			\midrule
			DPDNet$_S$& 22.85/0.668/0.0497/0.421 & 22.66/0.702/0.0504/0.397\\		
			Son et al.& 23.98/0.716/0.0433/0.339 &  24.35/0.756/0.0414/0.308\\
			IFAN& 24.71/\textbf{0.749}/0.0407/0.306 &  25.39/\textbf{0.794}/0.0380/0.264\\
			UNet & 23.43/0.694/0.0463/0.396 & 23.51/0.729/0.0456/0.368 \\
			MPRNet & 24.37/0.734/0.0413/0.346 &  24.66/0.771/0.0399/0.313\\
			\midrule
			IFAN* & \textbf{24.87}/0.748/\textbf{0.0395}/\textbf{0.288} &  \textbf{25.61}/0.793/\textbf{0.0367}/\textbf{0.243}\\
			UNet* & 23.55/0.696/0.0456/0.382 & 23.86/0.736/0.0441/0.349\\
			MPRNet* & 24.54/0.736/0.0403/0.339 &  24.90/0.775/0.0387/0.305\\	
			\bottomrule
			\hline
		\end{tabular}
	\end{table}
	
	\subsection{Experiments on DPDD and RealDoF datasets}
	For the DPDD dataset, apart from four methods mentioned above, the SOTA single image defocus deblurring methods Son et al. \cite{son2021single} and IFAN \cite{lee2021iterative} are also compared. It is worth noting that although the DPDD dataset is meant to be aligned, there actually exists slight misalignment as shown in Fig.~\ref{fig:intro}. Therefore, we calculate the evaluation metrics in view of both ($\hat{\bm{I}}$ \emph{v.s.} $\bm{I}_S$) and ($\hat{\bm{I}}$ \emph{v.s.} $\bm{I}_S^w$). 
	We also apply our JDRL framework to IFAN and MPRNet (denoted by IFAN* and MPRNet*). 
	All the evaluated methods are tested on both DPDD test set and RealDOF dataset. The experimental results are reported in Table~\ref{DPDD_tab} and Table~\ref{realdof_tab}. 
	
	From Table~\ref{DPDD_tab} and Table~\ref{realdof_tab}, we have a similar observation as in Table~\ref{tab:MDD}, \emph{i.e.}, the learning-based methods outperform the traditional ones, and the performance of UNet, IFAN and MPRNet can be notably improved on DPDD and RealDOF test sets, when they are embedded into our proposed JDRL. By tolerating the slight misalignment existing in DPDD dataset, JDRL contributes to better performance on both testing sets of DPDD and RealDoF. Additionally, since the evaluated methods are only trained on DPDD dataset, the improved performance on RealDOF also indicates the generalizing ability of JDRL. 
	The visual comparison on DPDD is given in suppl.  
	
	\subsection{Ablation Study}
	Six network variants are designed to analyze the performance of JDRL on SDD dataset in terms of its different components:
	\#1: the baseline network; \#2: without the bi-directional optical flow deforming process; \#3: without the calibration mask; \#4: without the reblurring module $\mathcal{M}_r$; \#5: without the isotropic design; \#6: without the branch of $\mathcal{R}_{wpn}$. Variants \#4, \#5 and \#6 are designed to prove the effectiveness of $\mathcal{M}_r$.
	The baseline deblurring network (UNet) is employed for all six ablative experiments.
	
	The quantitative results are reported in Table~\ref{tab:ablation}. From \#1 and \#2, we can see that the absence of the deforming process considerably downgrades the network performance. Besides, the network performance gets slightly improved by incorporating the calibration mask, which is indicated by \#3.
	From \#4, we can see that the absence of $\mathcal{M}_r$ heavily impairs the overall network performance. Finally, the reuslts of \#5 and \#6 prove that the isotropic design and $\mathcal{R}_{wpn}$ in $\mathcal{M}_r$ both benefit the reblurring process.
	The visual results and more details are included in supplementary materials.

	\begin{table}[t]
		\centering
		\setlength{\abovecaptionskip}{0pt} 
		\setlength{\belowcaptionskip}{0pt}
		\caption{\small{Ablation study on SDD dataset.}}
		\label{tab:ablation}
		\footnotesize
		\begin{tabular}{l|c}
			\toprule
			Category &  \ PSNR/SSIM/MAE/LPIPS\\
			\midrule
			\#1 Baseline   & 24.62/0.758/0.0360/0.344 \\
			\#2 w/o Cycle Deformation      &  24.80/0.758/0.0353/0.347 \\
			\#3 w/o Calibration Mask      & 25.78/0.780/0.0326/0.309 \\
			\midrule
			\#4 w/o $\mathcal{M}_{r}$      & 25.58/0.777/0.0335/0.311  \\
			\#5 w/o Isotropic Design     & 25.68/0.778/0.0329/0.311 \\
			\#6 w/o $\mathcal{R}_{wpn}$      & 25.74/0.778/0.0330/0.332 \\
			\midrule
			\ \ \ \ \ Final      &\textbf{25.82}/\textbf{0.783}/\textbf{0.0320}/\textbf{0.305}\\		
			\bottomrule
			\hline
		\end{tabular}
	\end{table}

	\section{Conclusion}
	\label{sec:conclusion}
	In this paper, we proposed a joint deblurring and reblurring learning (JDRL) framework for single image defocus deblurring. 
	On the one hand, a deblurring module $\mathcal{M}_{d}$ with spatially adaptive deblurring loss is developed to tolerate misalignment between image pairs with the help of optical flow. Strategies of calibration mask and cycle deformation are employed to enhance the robustness of deformation process. On the other hand, a spatially invariant reblurring module ($\mathcal{M}_{r}$) is proposed, which consists of a kernel prediction network ($\mathcal{R}_{kpn}$) and a weight prediction network ($\mathcal{R}_{wpn}$). The spatial consistency between deblurred image and blurry image is guaranteed by the isotropic blur kernels predicted by $\mathcal{R}_{kpn}$. The proposed JDRL framework can be widely adopted, since it is not coupled with deblurring network architecture.  
	Experimental results on SDD, DPDD and RealDOF datasets validate the effectiveness of JDRL.
	Our newly established SDD dataset for single image defocus deblurring would also benefit future research in this field.

\clearpage

\twocolumn[
\begin{center}
	{\LARGE \textbf{Learning Single Image Defocus Deblurring with Misaligned Training Pairs\\
	~~~\\
	\emph{Supplemental Materials}}}
	
	\vspace{2cm}
\end{center}]

\setcounter{section}{0}

\begin{figure*}[h]
	\centering
	\setlength{\abovecaptionskip}{0pt} 
\setlength{\belowcaptionskip}{0pt}
\begin{overpic}[width=1\textwidth]{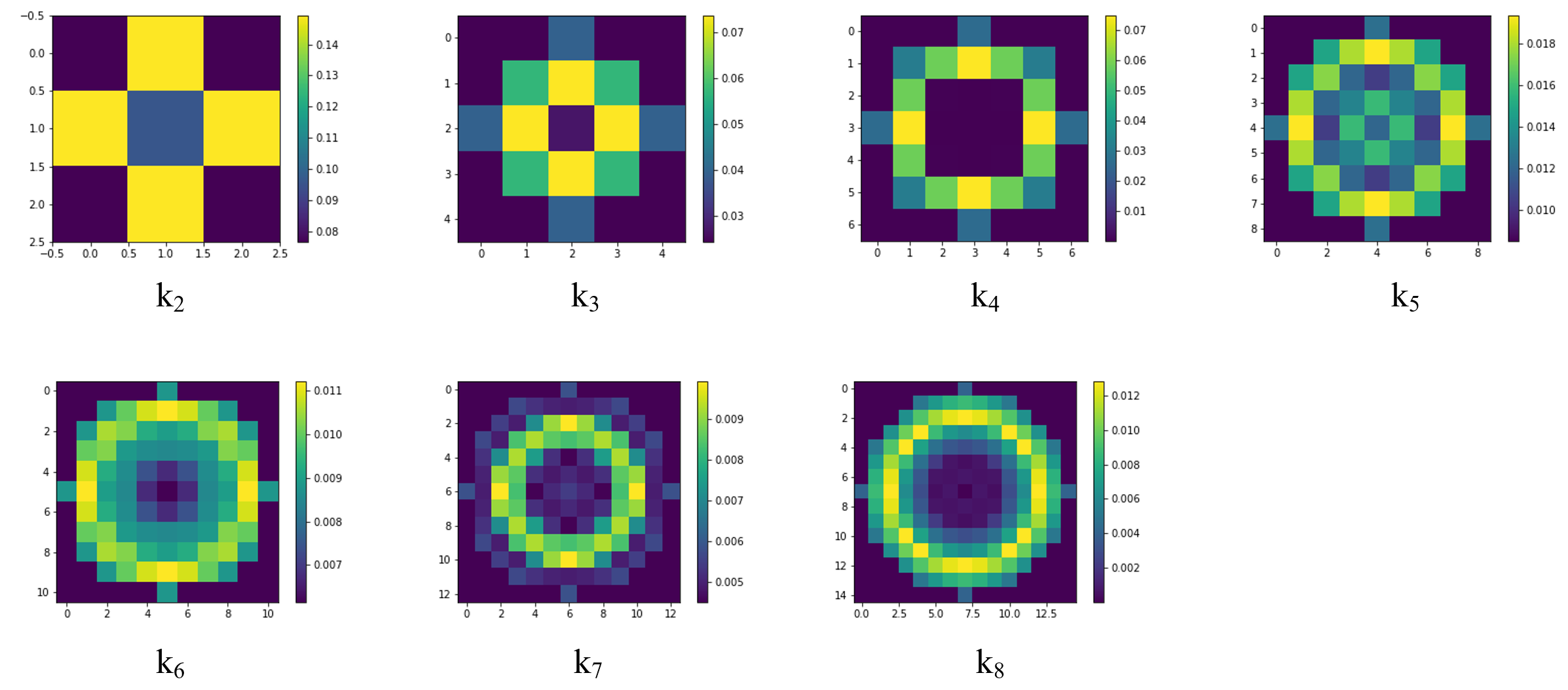}
	\end{overpic}

	\caption{\small Isotropic blur kernels of different kernel sizes generated by $\mathcal{R}_{kpn}$. We randomly visualize one example for each kernel size.}
	\label{fig:kernel}
\end{figure*}

\section{Spatially Invariant Reblurring Module}
	
	Our reblurring network $\mathcal{R}$ is proposed for generating the reblurred image $\hat{\bm{I}}_B$, which is then supervised by input image $\bm{I}_B$. In this process, isotropic blur kernels predicted by $\mathcal{R}_{kpn}$ are used to guarantee the spatial consistency between the deblurred result $\hat{\bm{I}}$ and $\bm{I}_B$. Several examples of isotropic blur kernels generated by $\mathcal{R}_{kpn}$ are given in Fig.~\ref{fig:kernel}.
	
	\section{Discussion on the Misalignment Problem}
	{{
	The phenomenon of misalignment widely exists in real-world datasets, leading to undesirable influence on relevant visual tasks. Thus, researchers have proposed approaches in various aspects to tackle the misalignment problem. 
	
	In the field of image translation, a common strategy to deal with misalignment is cycle-consistency \cite{zhu2017unpaired}. However, the method based on cycle-consistency may produce multiple solutions and is sensitive to perturbation. Therefore, methods like RegGAN \cite{kong2021breaking} have been proposed to address this problem. The spatial deformation predicted by RegGAN is a single-direction process. By contrast, our cycle deformation mechanism works in a bi-directional manner. Moreover, we adopt an dedicated isotropic kernel generation module to maintain spatial consistency during the reblurring process.
	
	Zhang et al. \cite{zhang2019zoom} propose a contextual bilateral loss to address the misalignment problem in super resolution tasks. We have also tried adapting contextual loss to deblurring problems. However, it turned out that the UNet trained with contextual bilateral loss achieved inferior performance compared with the UNet trained with JDRL (24.58/0.756/0.0355/0.326 vs. 25.82/0.783/0.0320/0.305 in terms of PSNR/SSIM/MAE/LPIPS), which is partially attributed to its incapability in dealing with severe misalignment.
	}}
	
	\section{Ablation Study}
\begin{figure*}[t]
		\small
		
		\centering
		\begin{tabular}{cc}
			\footnotesize
			\begin{adjustbox}{valign=t}
				\begin{tabular}{cccccc}
					\includegraphics[width=0.16\textwidth]{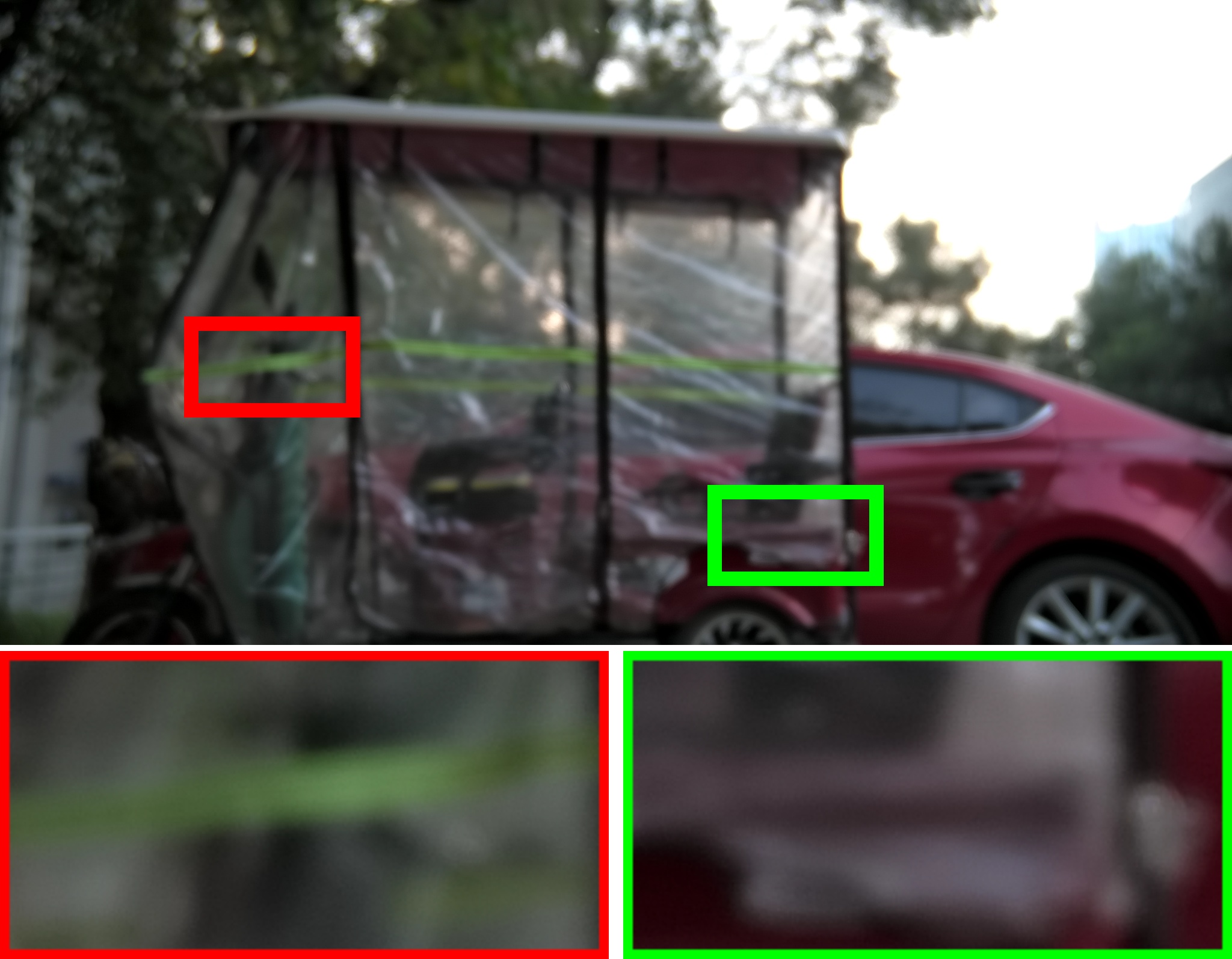}&\hspace{-4.2mm}
					\includegraphics[width=0.16\textwidth]{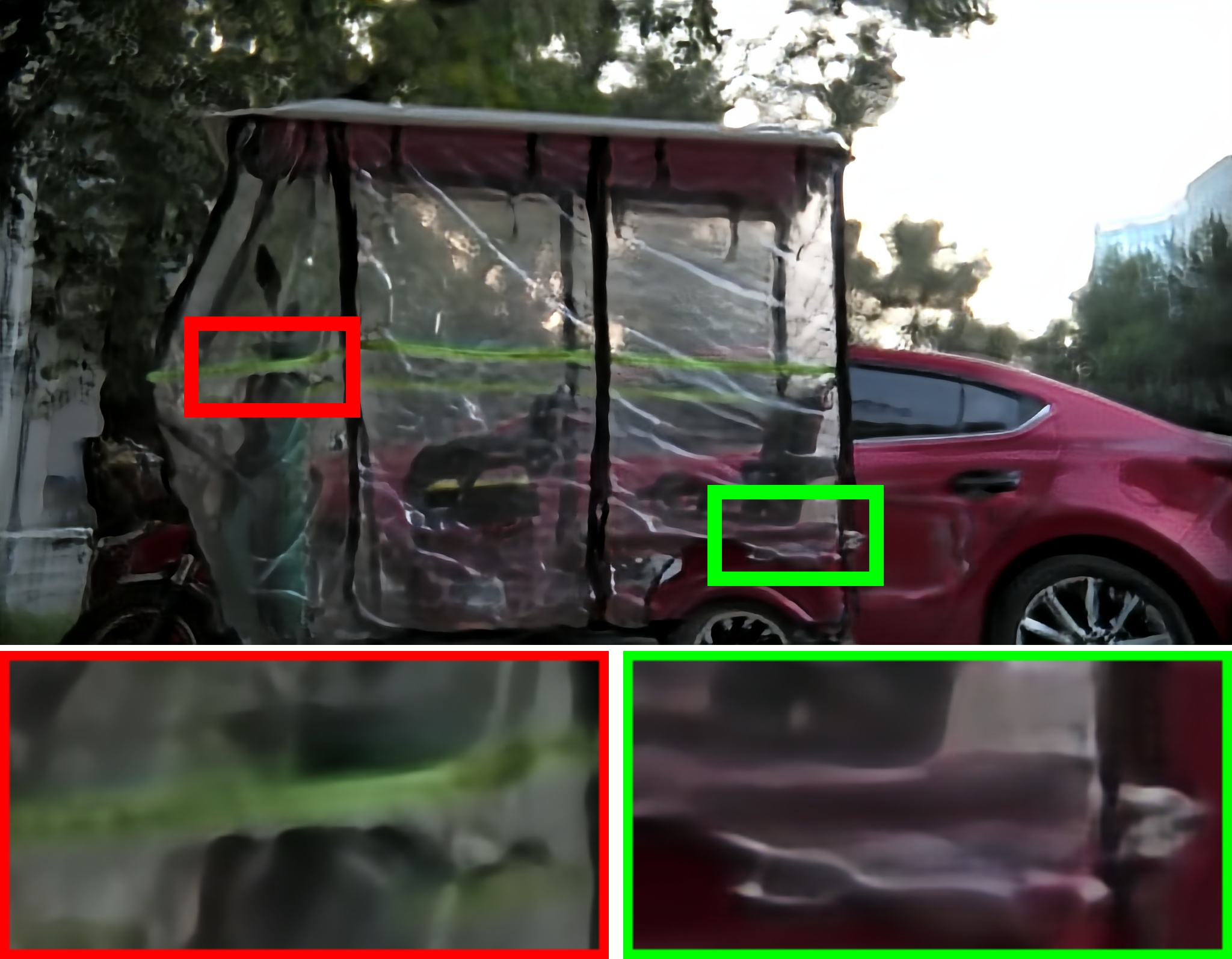}&\hspace{-4.2mm}
					\includegraphics[width=0.16\textwidth]{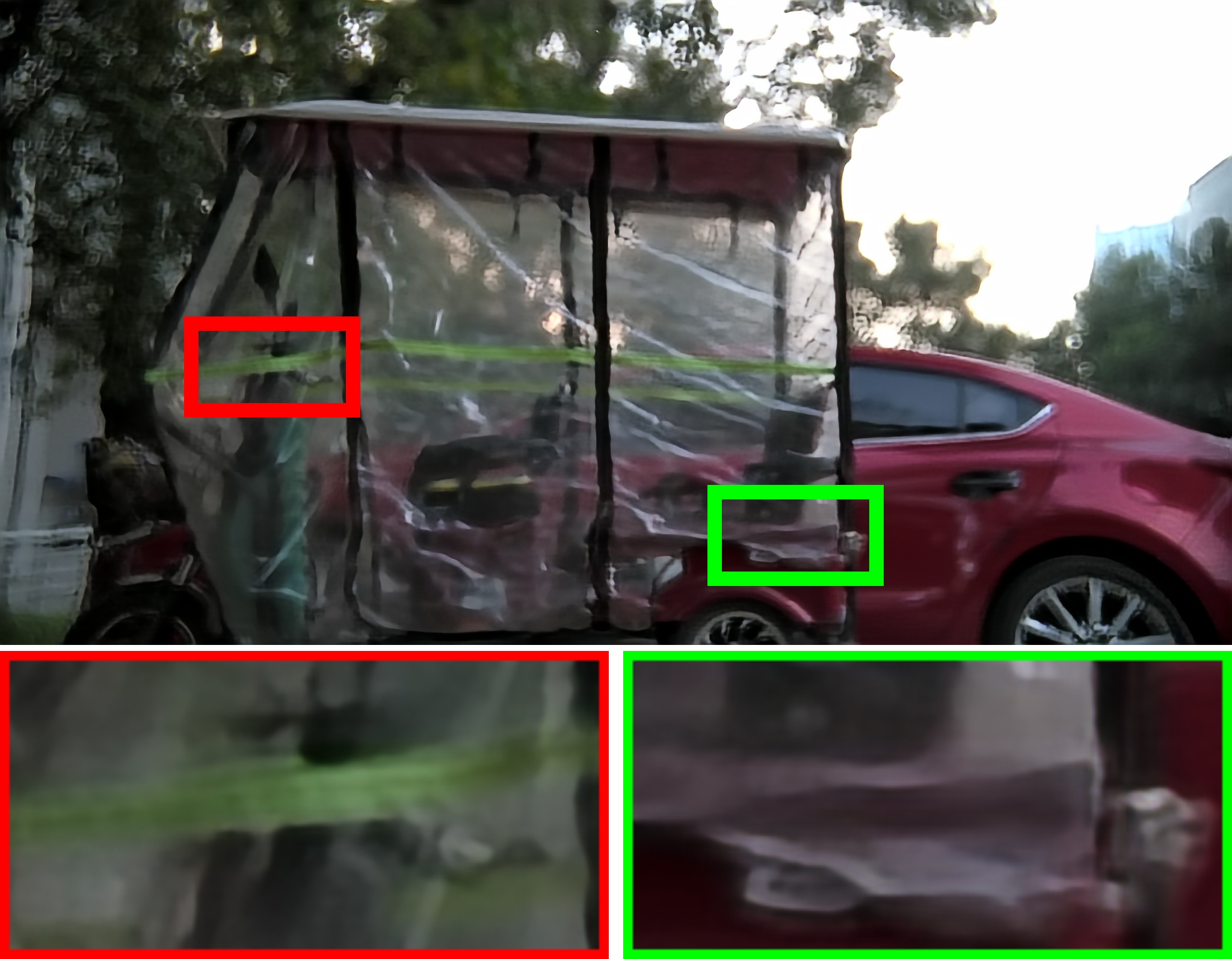}&\hspace{-4.2mm}
					\includegraphics[width=0.16\textwidth]{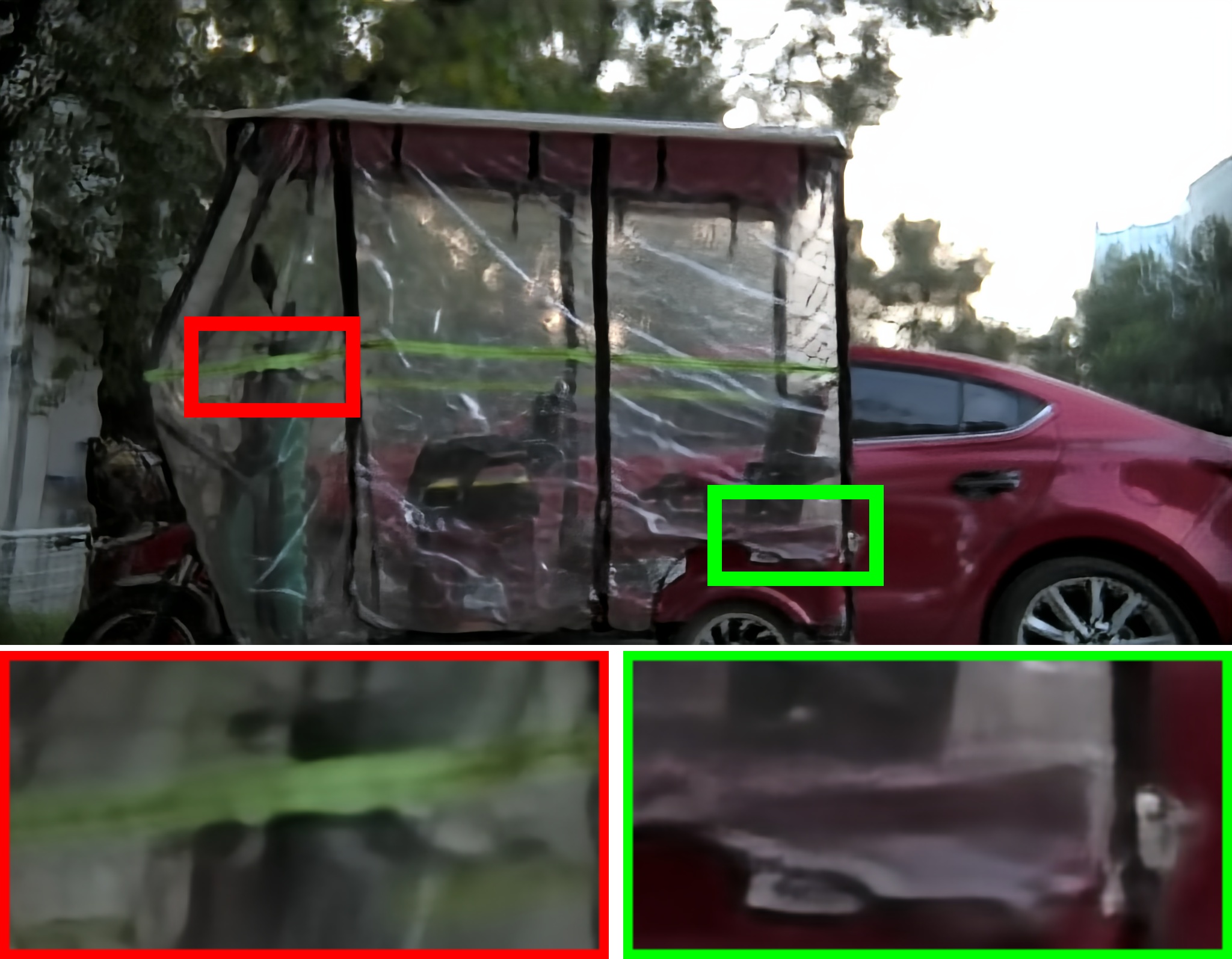}&\hspace{-4.2mm}
					\includegraphics[width=0.16\textwidth]{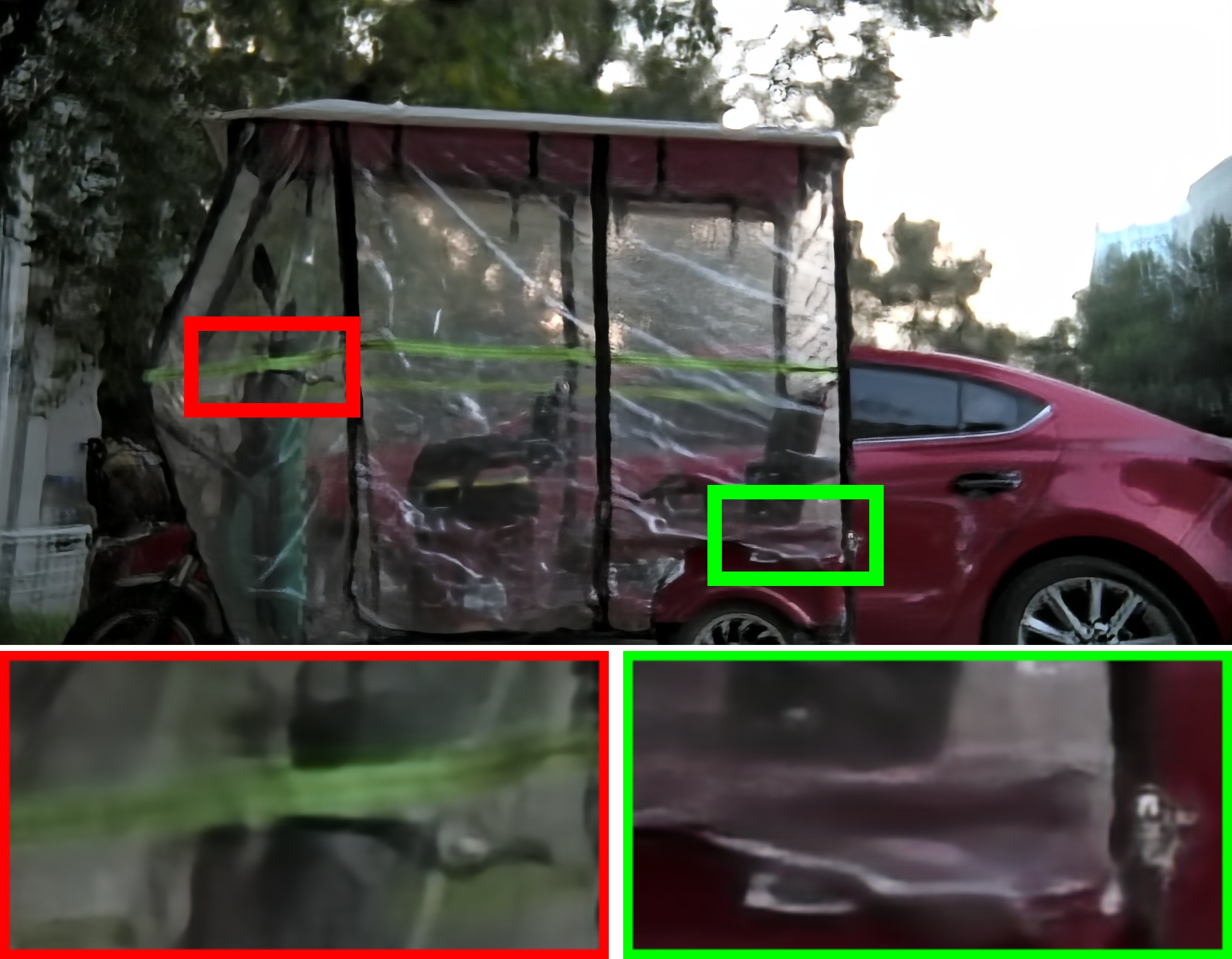}&\hspace{-4.2mm}\includegraphics[width=0.16\textwidth]{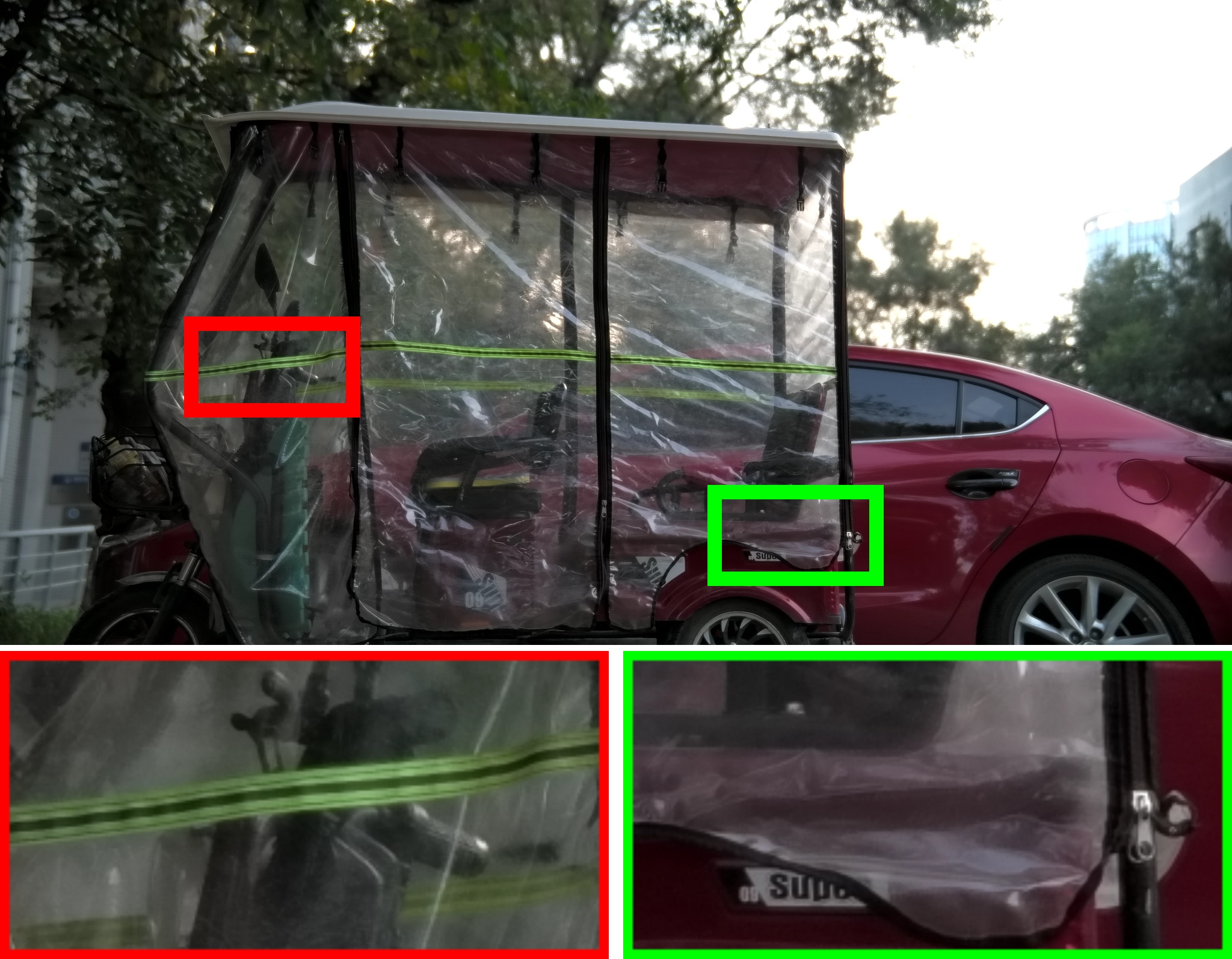}
					\\
					\includegraphics[width=0.16\textwidth]{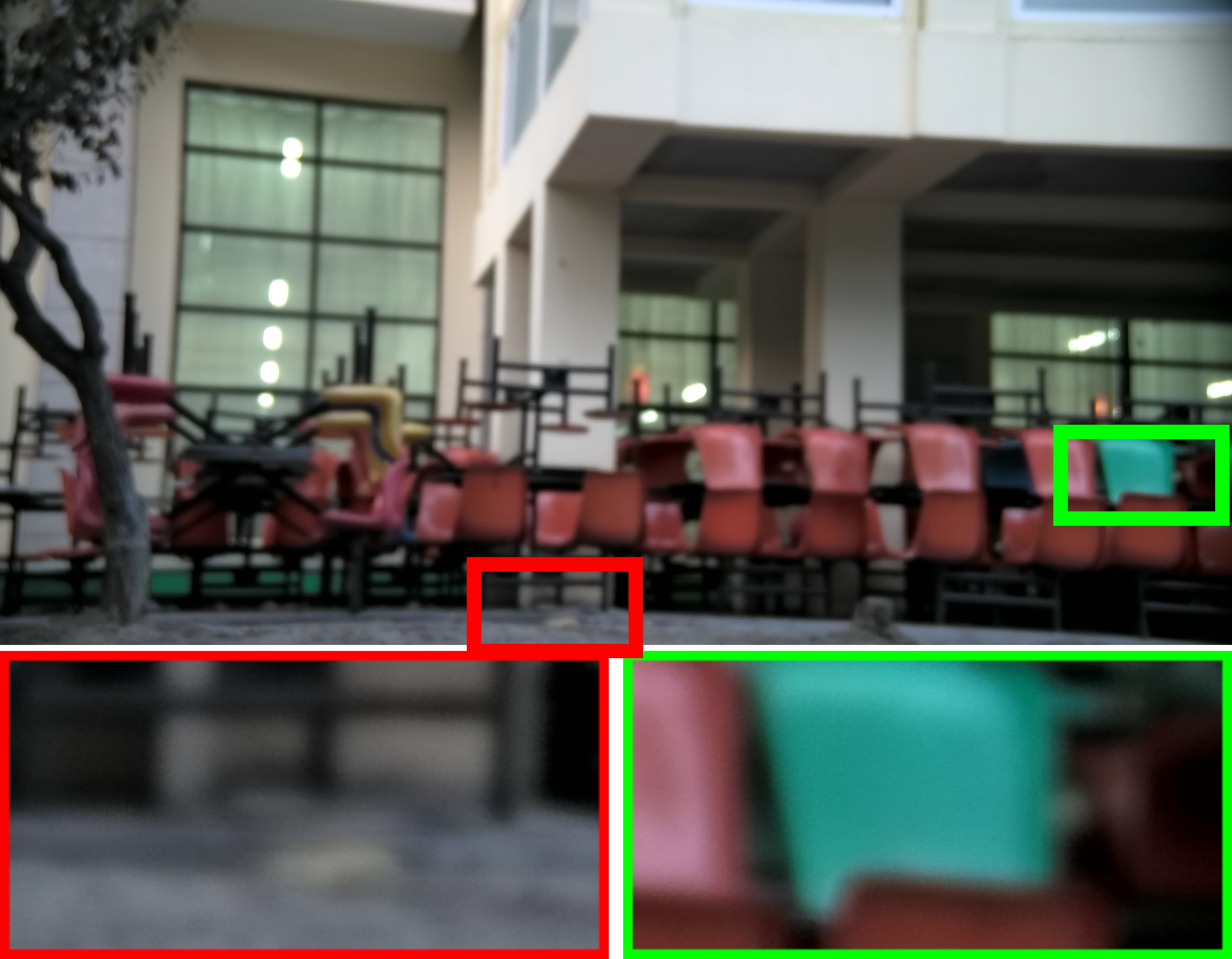}&\hspace{-4.2mm}
					\includegraphics[width=0.16\textwidth]{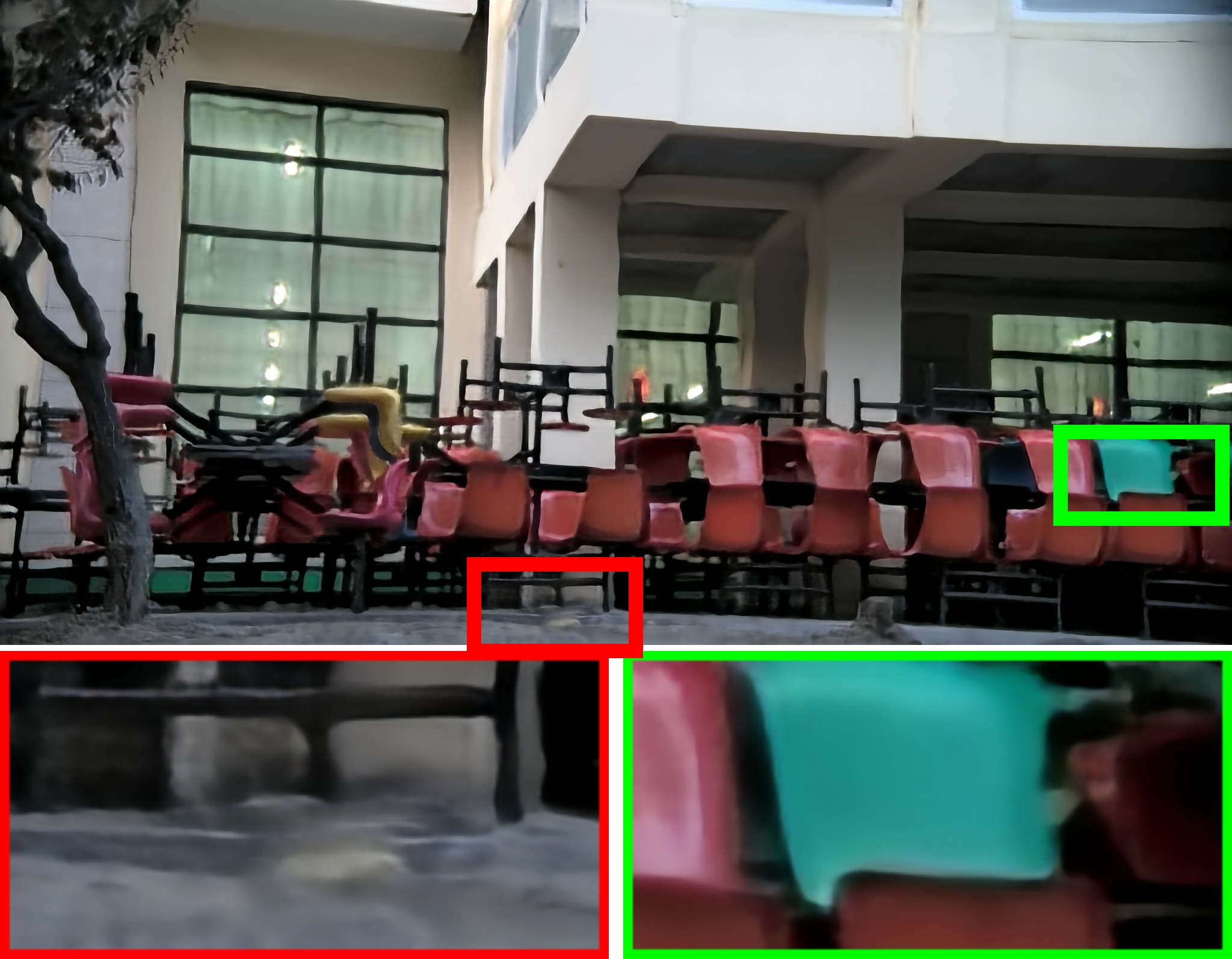}&\hspace{-4.2mm}
					\includegraphics[width=0.16\textwidth]{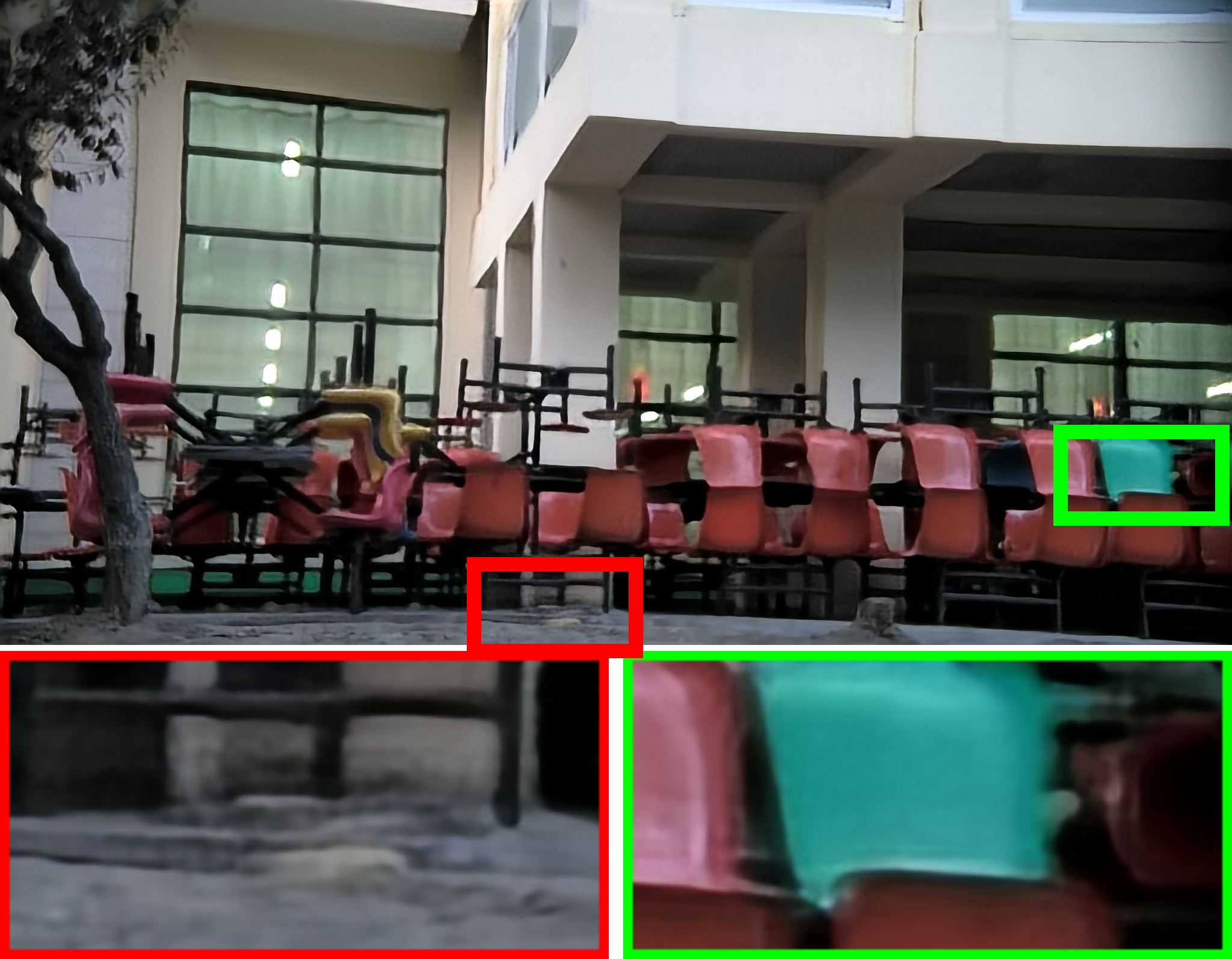}&\hspace{-4.2mm}
					\includegraphics[width=0.16\textwidth]{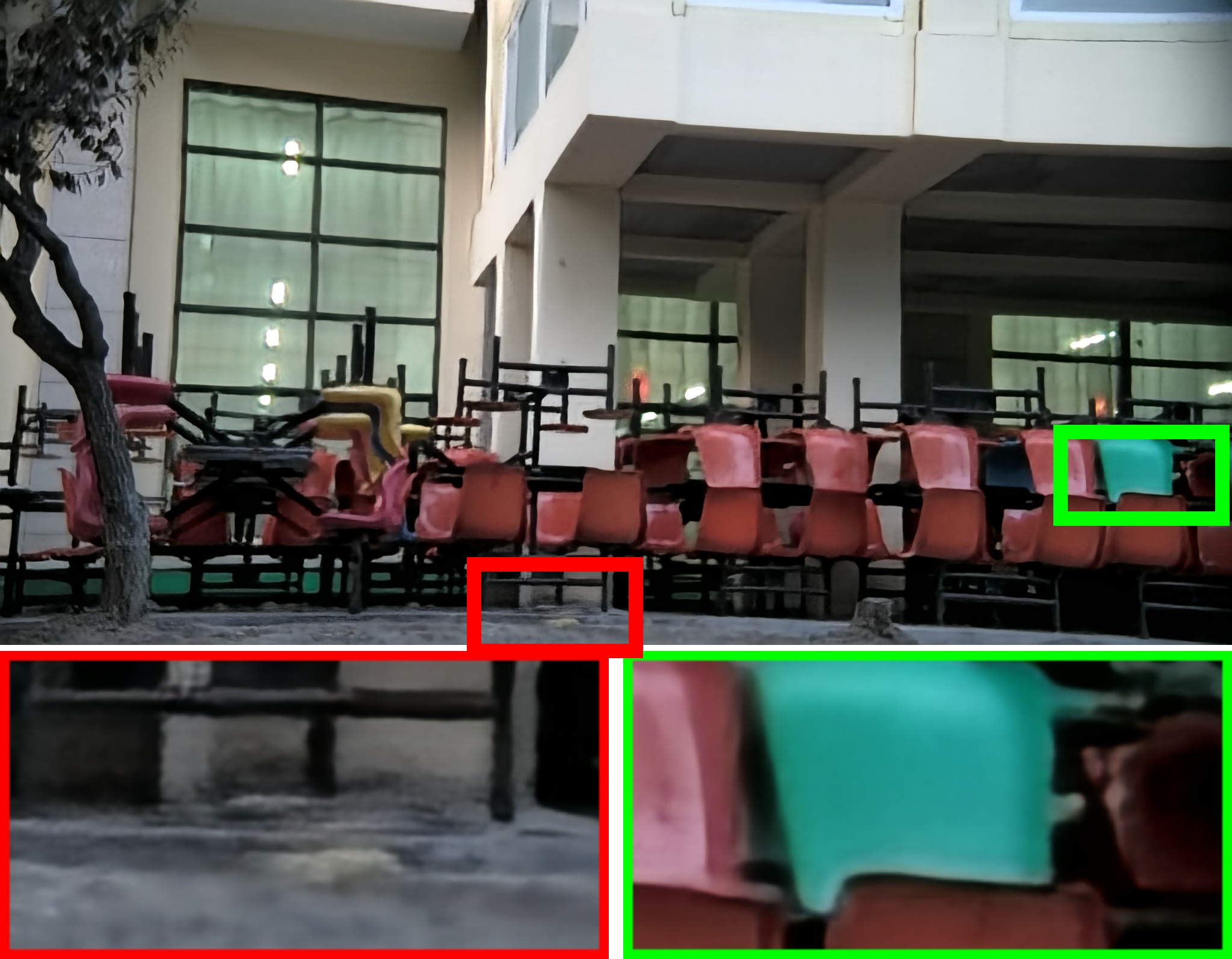}&\hspace{-4.2mm}
					\includegraphics[width=0.16\textwidth]{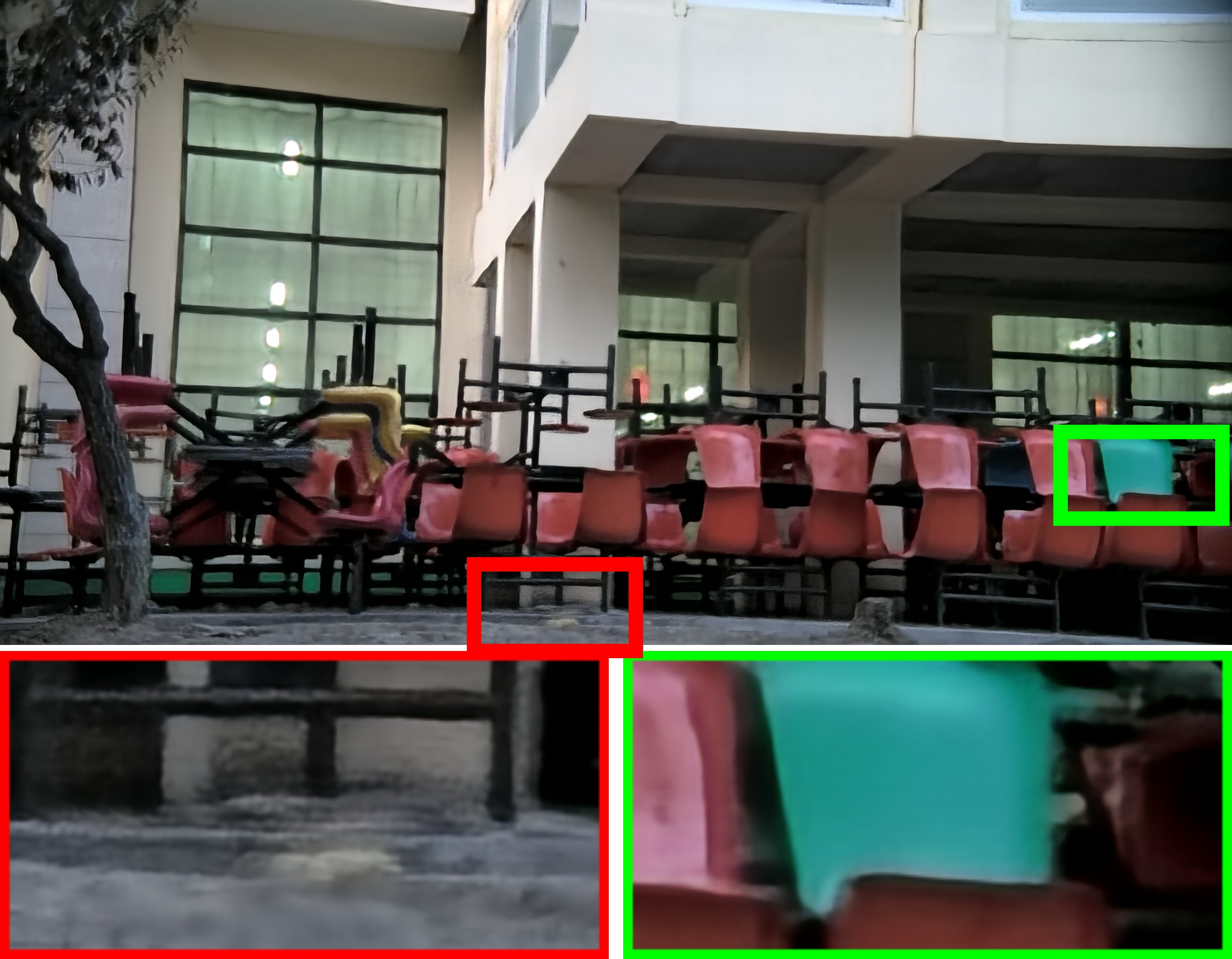}&\hspace{-4.2mm}
					\includegraphics[width=0.16\textwidth]{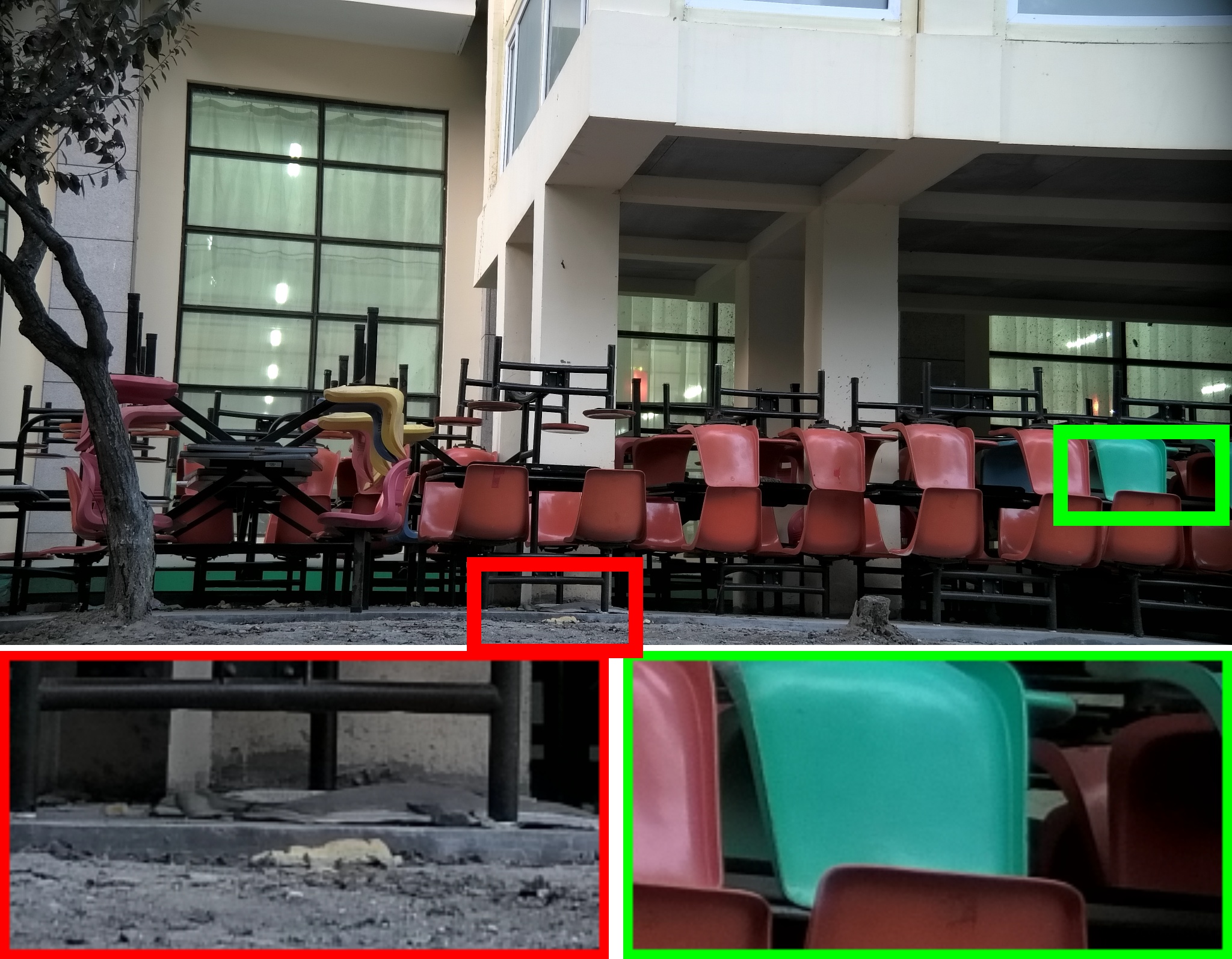}
					\\
					Input&\hspace{-4.2mm}
					\# 1&\hspace{-4.2mm}
					\# 2&\hspace{-4.2mm}
					\# 3&\hspace{-4.2mm}
					Final&\hspace{-4.2mm}
					GT
					\\
				\end{tabular}
			\end{adjustbox}
			
		\end{tabular}
		\caption{Visual results of ablation study on SDD dataset.}
		\label{ab1}
	\end{figure*}

\begin{figure*}[t]
		\small
		
		\centering
		\begin{tabular}{cc}
			\footnotesize
			\begin{adjustbox}{valign=t}
				\begin{tabular}{cccccc}
					\includegraphics[width=0.16\textwidth]{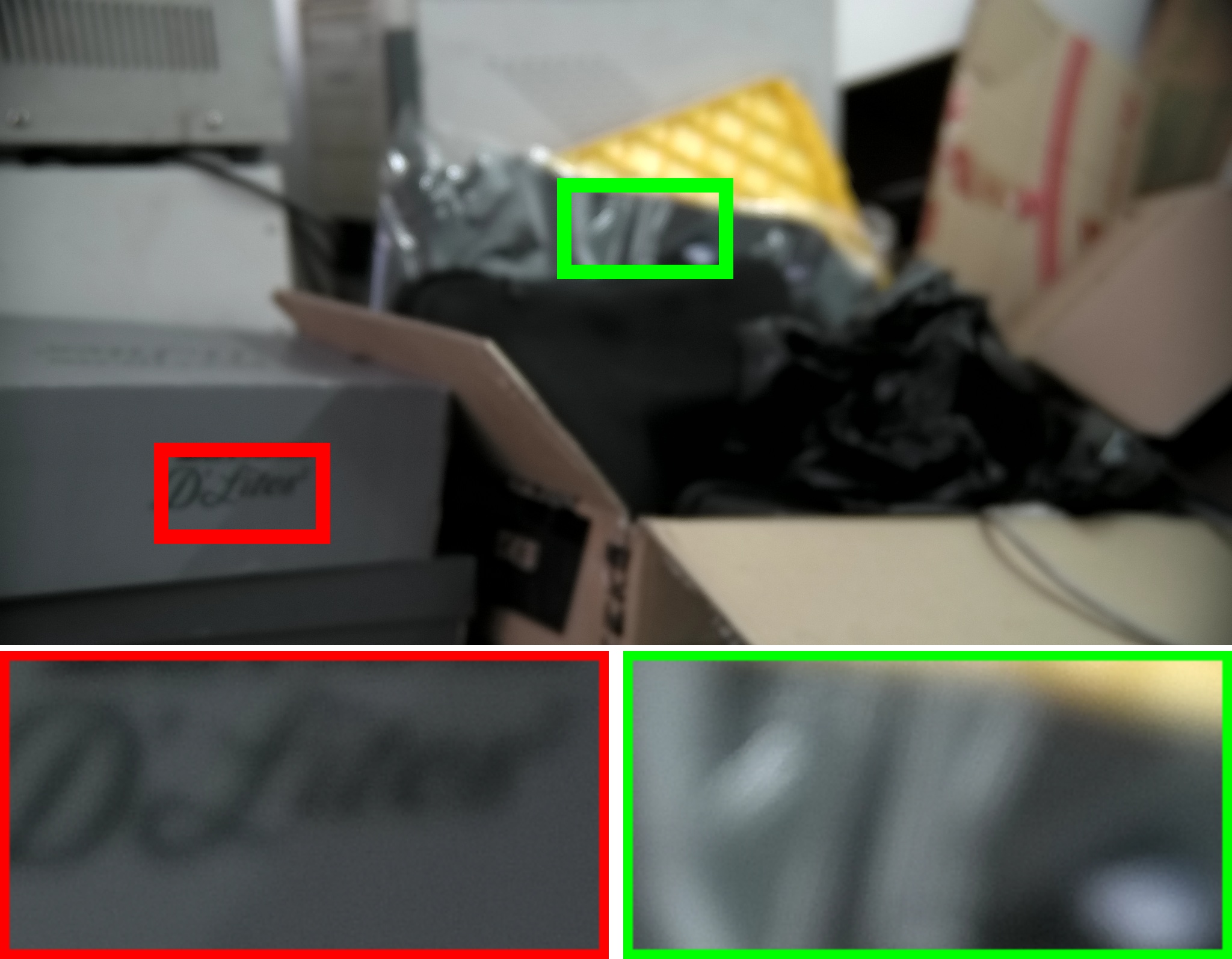}&\hspace{-4.2mm}
					\includegraphics[width=0.16\textwidth]{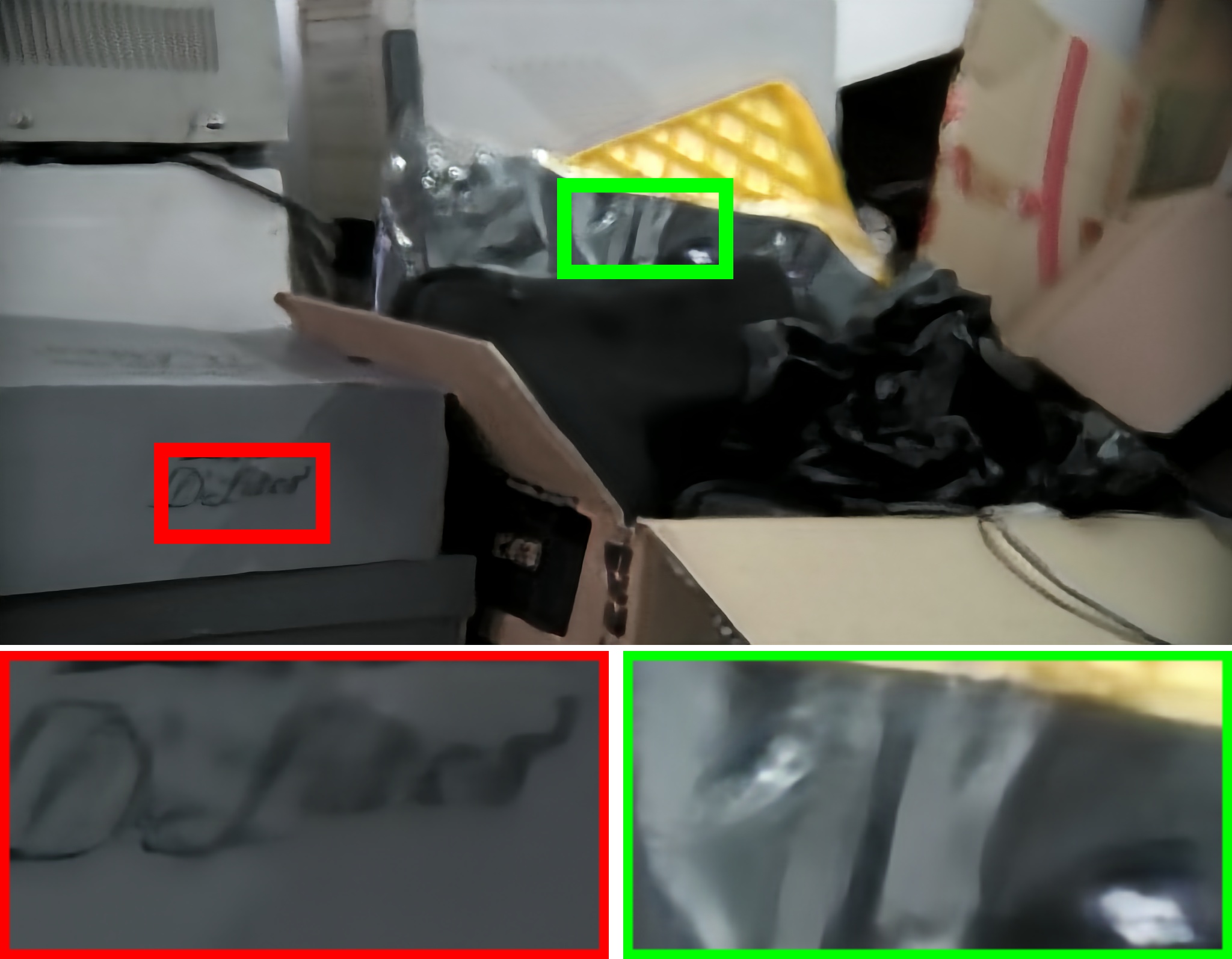}&\hspace{-4.2mm}
					\includegraphics[width=0.16\textwidth]{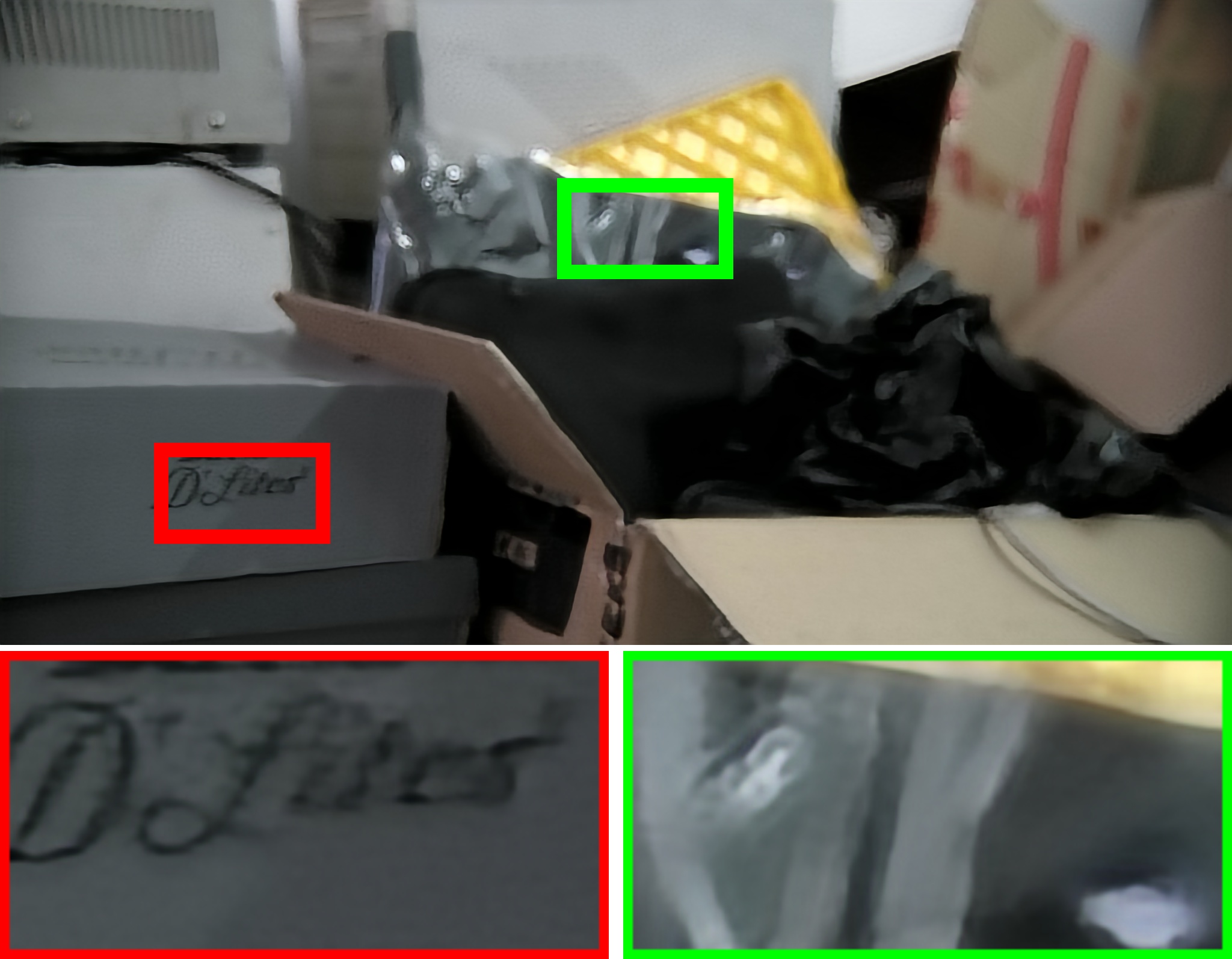}&\hspace{-4.2mm}
					\includegraphics[width=0.16\textwidth]{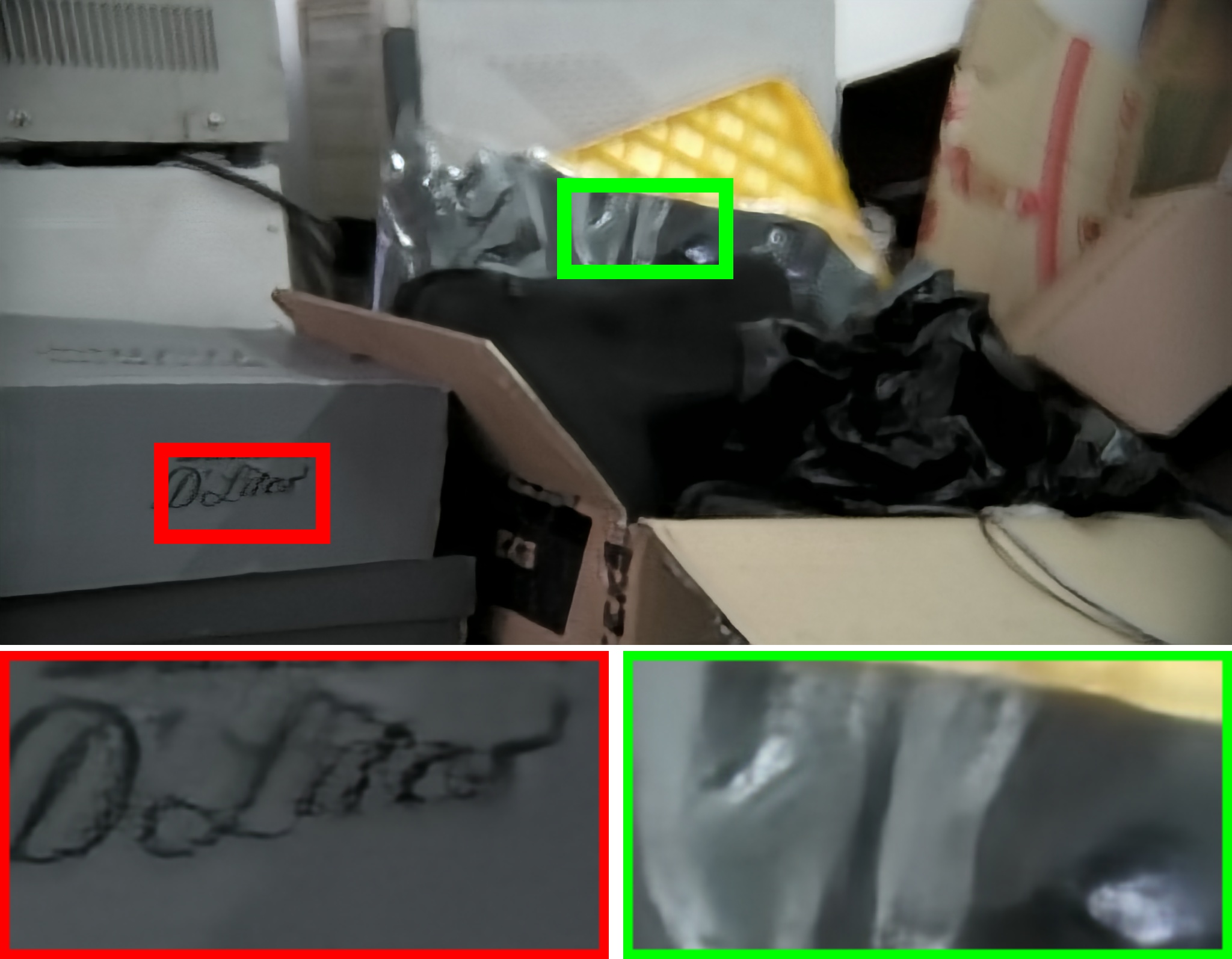}&\hspace{-4.2mm}
					\includegraphics[width=0.16\textwidth]{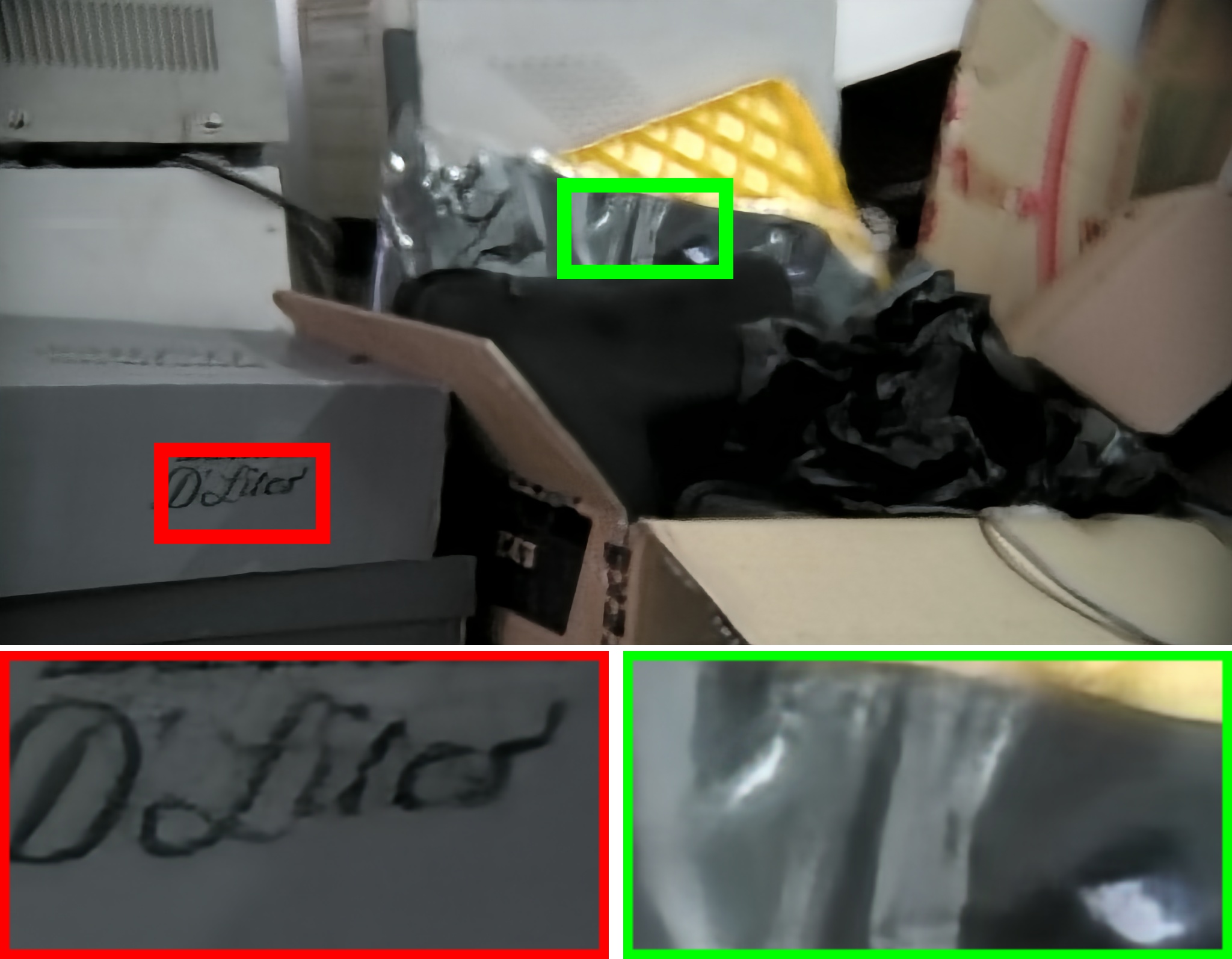}&\hspace{-4.2mm}\includegraphics[width=0.16\textwidth]{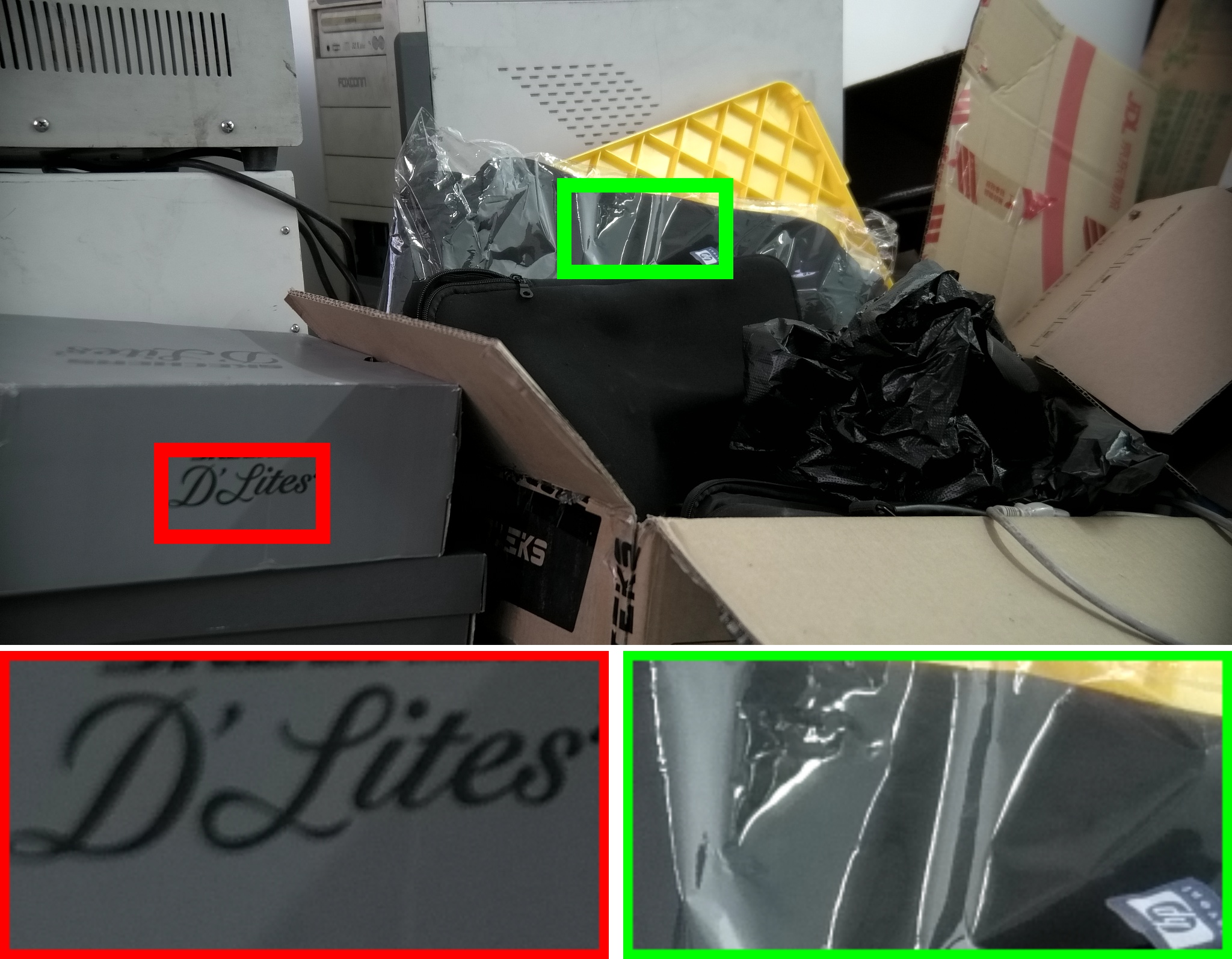}
					\\
					\includegraphics[width=0.16\textwidth]{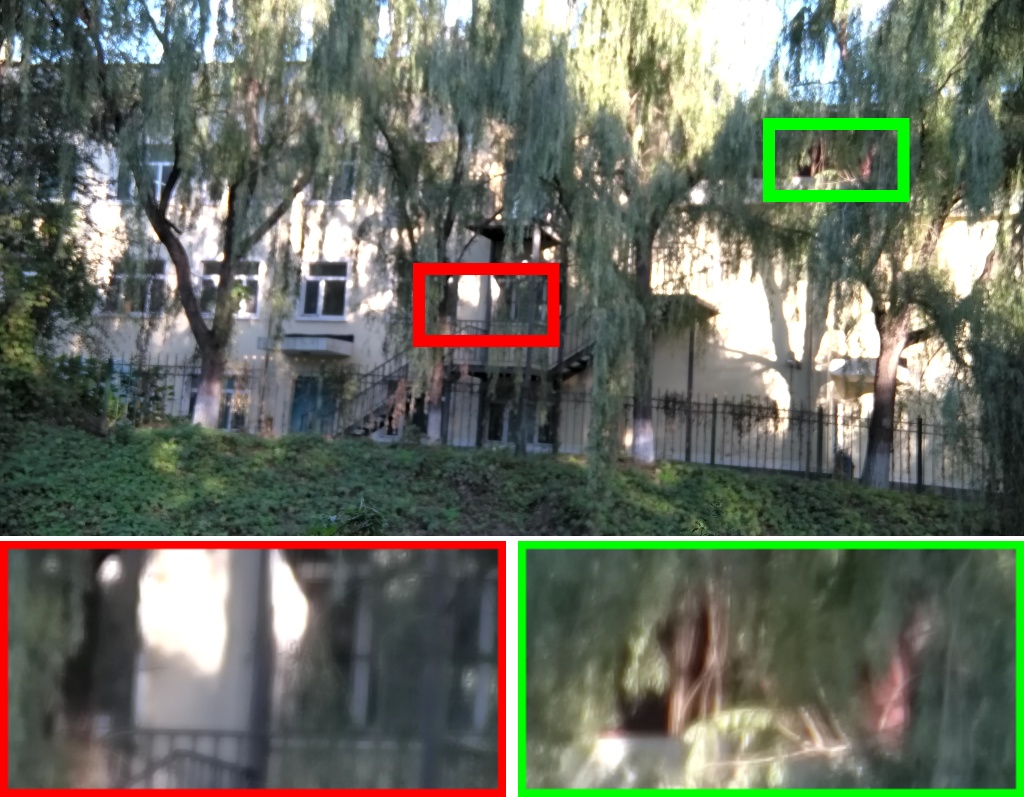}&\hspace{-4.2mm}
					\includegraphics[width=0.16\textwidth]{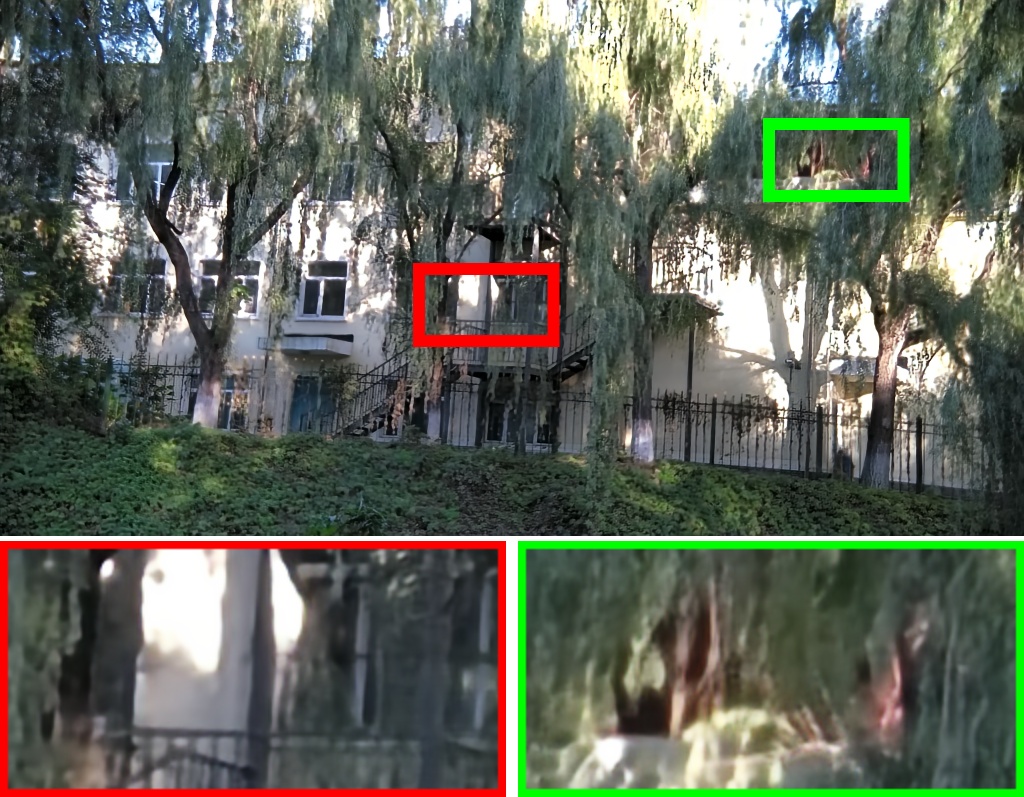}&\hspace{-4.2mm}
					\includegraphics[width=0.16\textwidth]{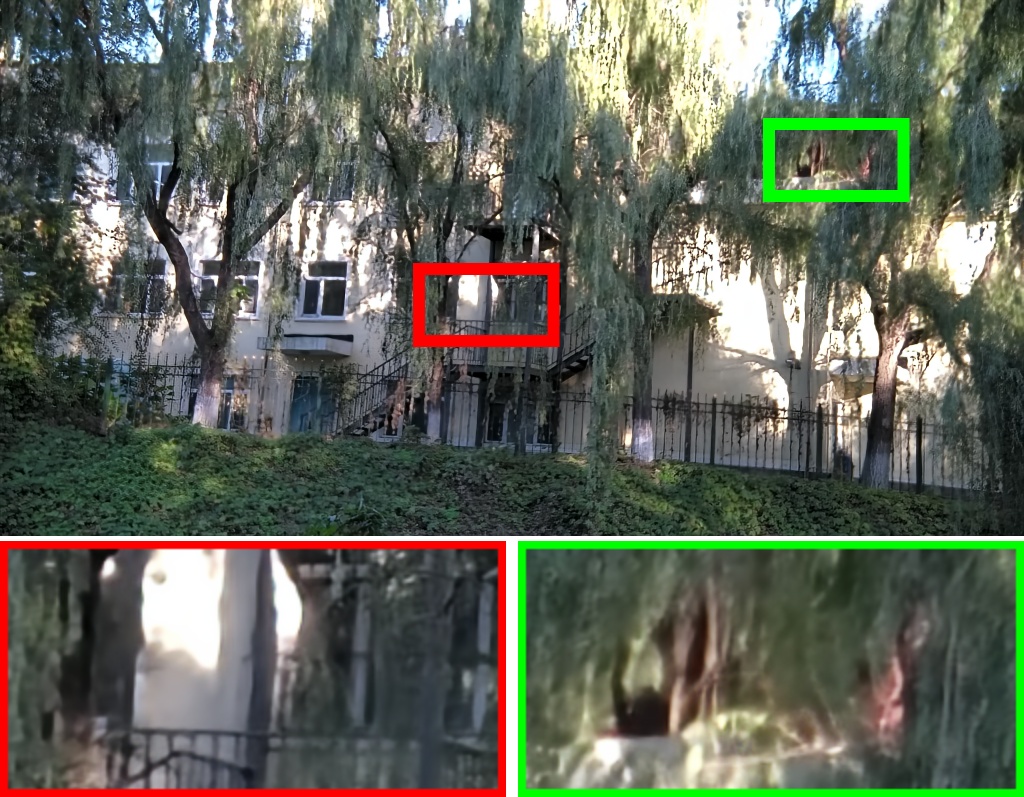}&\hspace{-4.2mm}
					\includegraphics[width=0.16\textwidth]{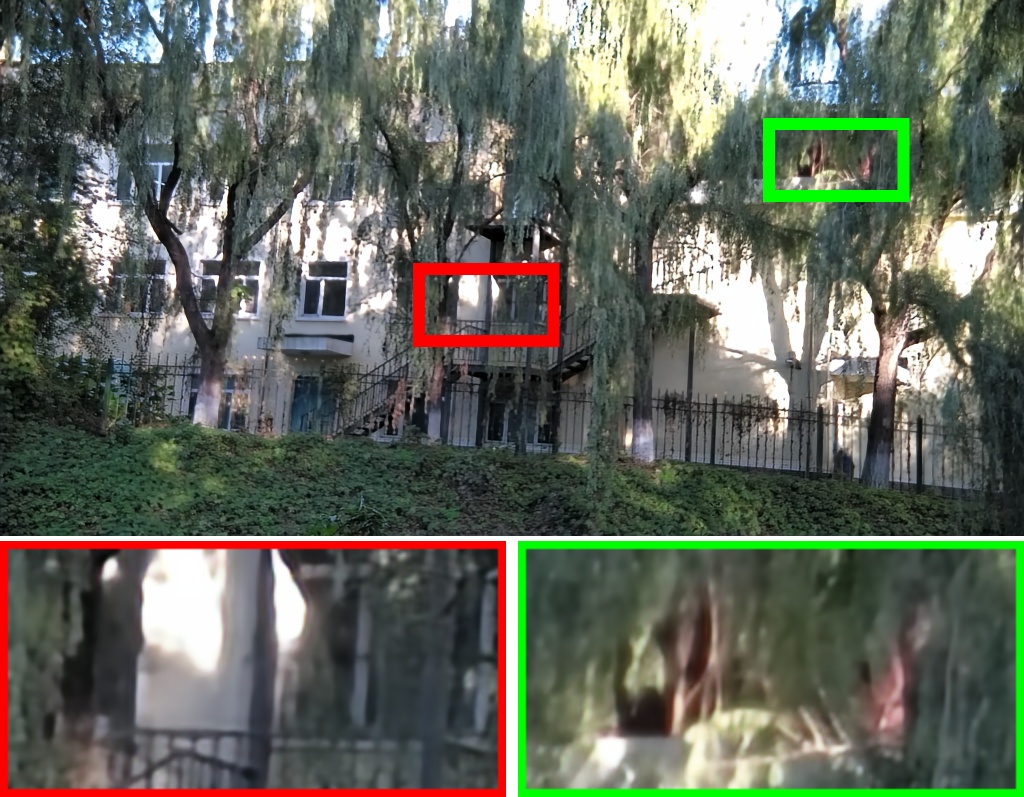}&\hspace{-4.2mm}
					\includegraphics[width=0.16\textwidth]{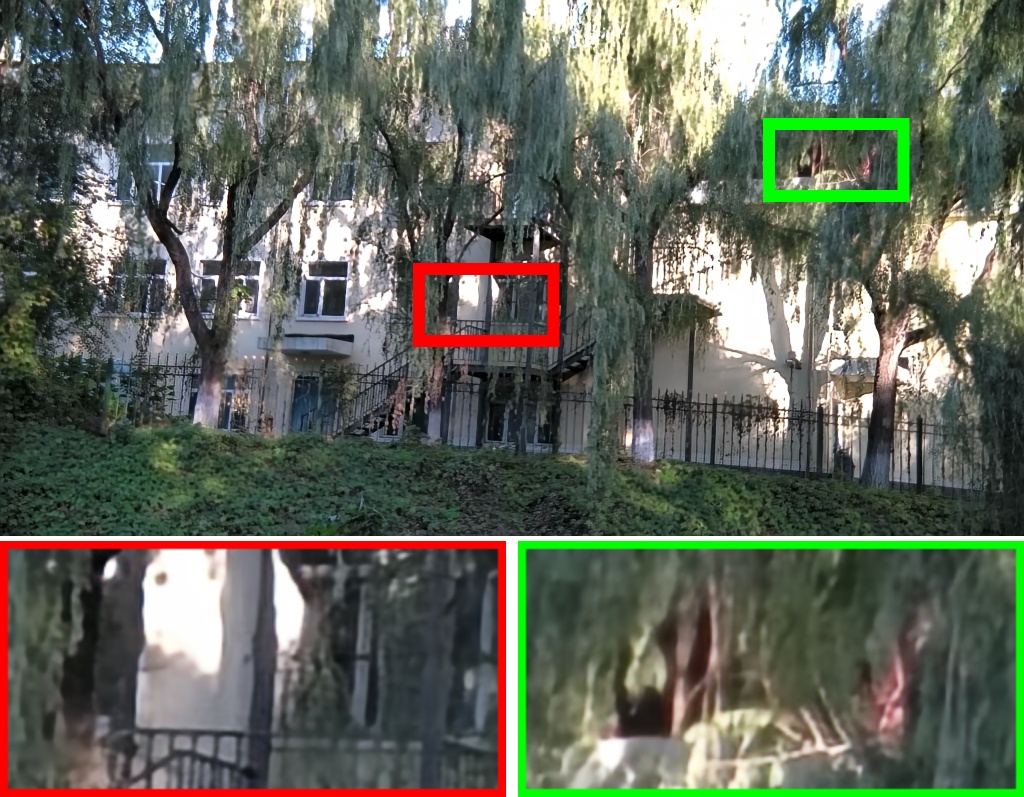}&\hspace{-4.2mm}
					\includegraphics[width=0.16\textwidth]{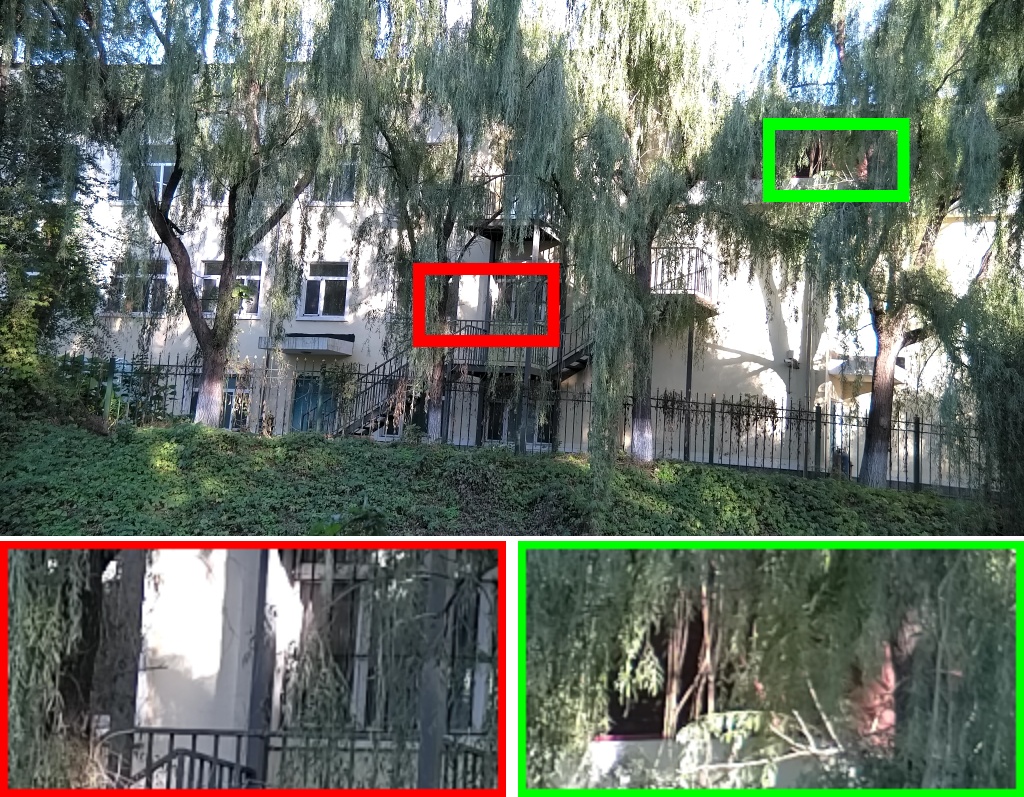}
					\\
					Input&\hspace{-4.2mm}
					\# 4&\hspace{-4.2mm}
					\# 5&\hspace{-4.2mm}
					\# 6&\hspace{-4.2mm}
					Final&\hspace{-4.2mm}
					GT
					\\
				\end{tabular}
			\end{adjustbox}
			
		\end{tabular}
		\caption{Visual results of ablation study on SDD dataset.}
		\label{ab2}
	\end{figure*}

	The visual results of ablation study from the main text are presented in Fig.~\ref{ab1} and Fig.~\ref{ab2}. In the ablation study, six variants of the original network are designed to investigate the significance of different network components:
	\#1: baseline network, \emph{i.e.}, a vanilla UNet; \#2: without the optical flow deforming process; \#3: without the calibration mask; \#4: without the reblurring module $\mathcal{M}_r$; \#5: without the isotropic design; \#6: without the weight prediction network $\mathcal{R}_{wpn}$, \emph{i.e.}, using only fixed-size kernels (15$\times$15) predicted by $\mathcal{R}_{kpn}$.
	
	As shown in Fig.~\ref{ab1}, the results from \#1 and \#2 have fewer textural details and could suffer from shape deformation. 
	This is because the pixel-wise loss encourages positionally consistent mapping between the input and output, while such mapping is hard to learn between unaligned image pairs without the deforming process.
	\#3 performs better than \#1 and \#2 yet still have less textures than the final model due to the missing calibration mask. 
	
	By comparing the visual results in Fig.~\ref{ab2}, one can see that the network performance gets considerably downgraded without the reblurring module (\#4). 
	The absence of isotropic kernels (\#5) would also lead to inferior visual results, since the output of reblurred module can be misaligned with the input image in this case. What's more, the network performance gets impaired without the kernel prediction network $\mathcal{R}_{wpn}$ (\#6). The function of $\mathcal{R}_{wpn}$ (\#6) is generating kernels of different blurry levels, such that resulting in reblurred ouputs that are more coherent with the natural state of blurry images.
	
	\begin{figure*}[htbp]
		\small
		\centering
		\begin{tabular}{cc}
			\footnotesize
			\hspace{-0.4cm}
			\begin{adjustbox}{valign=t}
				\begin{tabular}{c}	
					\includegraphics[width=0.85\textwidth]{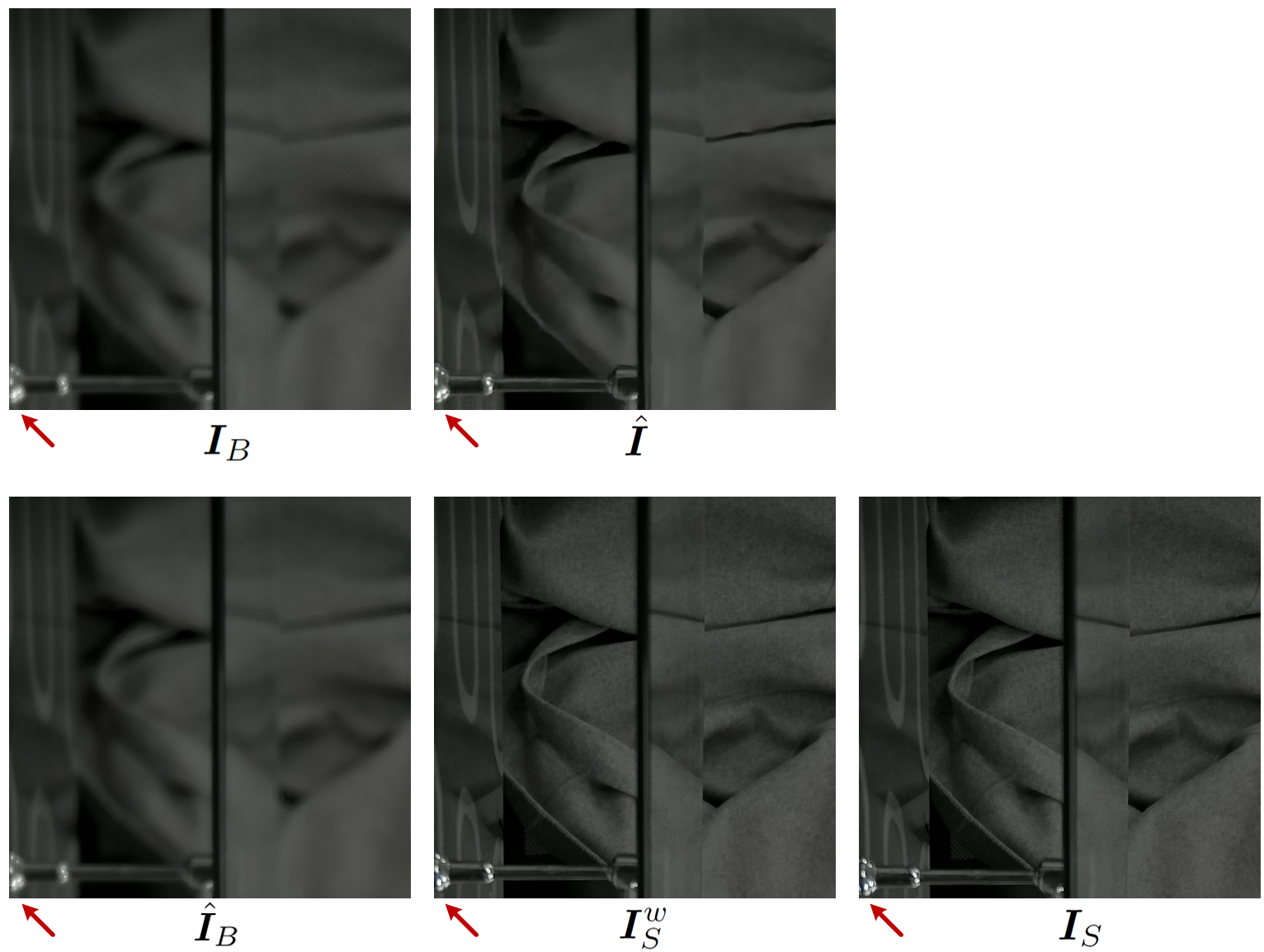}
					\\
					\includegraphics[width=0.85\textwidth]{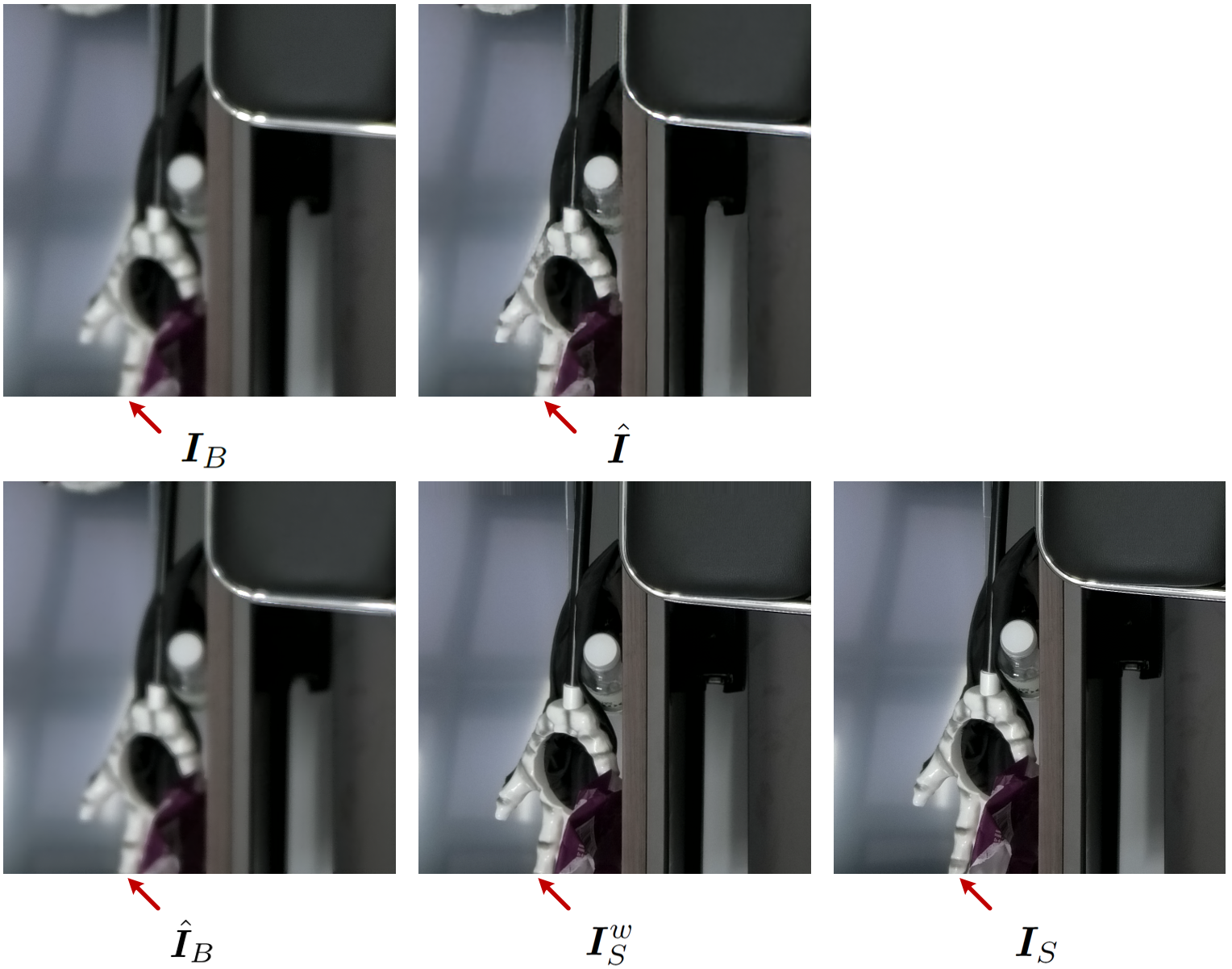}
					
				\end{tabular}
			\end{adjustbox}
			
		\end{tabular}
		\caption{Visualization of reblurred results and deformed ground-truth sharp images.}
		\label{fig:reblur}
	\end{figure*}
	
	\begin{figure*}[h]
		\subfigure[When the optical flow is correctly estimated, the values of calibration mask are all 1 (i.e, all white in RGB color space).]{
			\includegraphics[width=\textwidth]{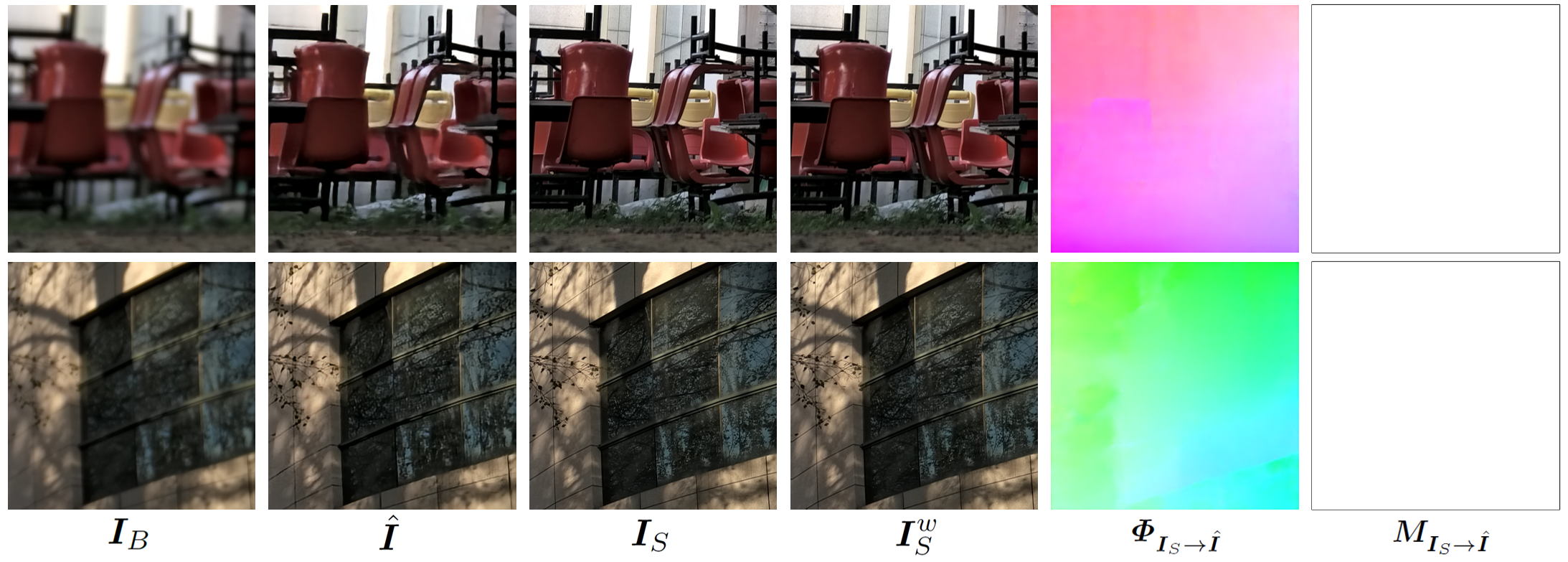}
			\label{fig:mask0}
		}
		\subfigure[When the optical flow is facing estimation error, the ground-truth sharp image will also be wrongly deformed (note the region marked by green rectangles). 
		In this circumstance, the calibration mask will filter out this adverse region, in order to prevent it from influencing the subsequent loss calculation (Eq. (5)).]{
			\includegraphics[width=\textwidth]{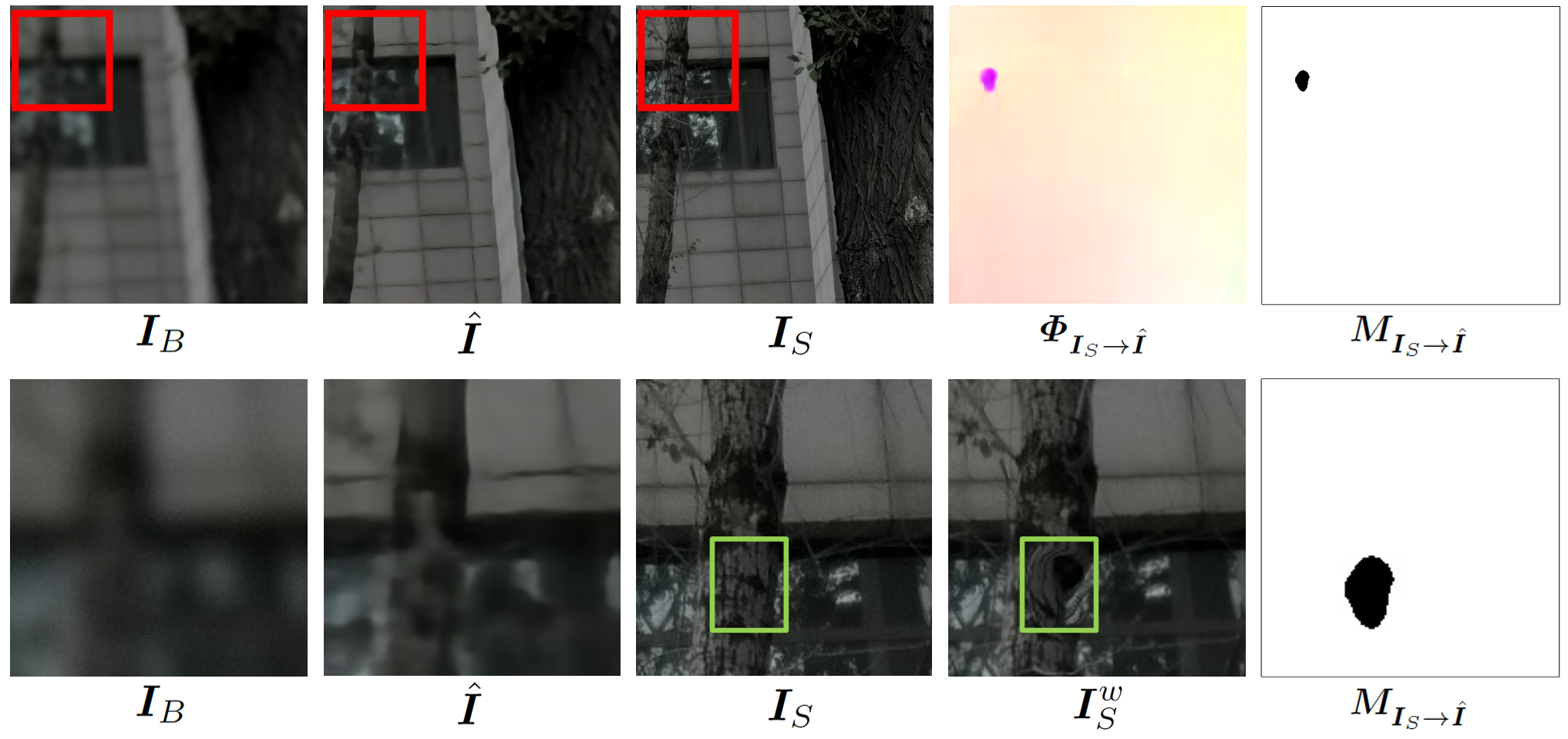}
			\label{fig:mask1}
		}
		\caption{Illustration of how calibration masks work.}
		\label{fig:mask}
	\end{figure*}

	\section{Calibration Mask}
	
	The functionality of calibration masks is illustrated in Fig~\ref{fig:mask}. Because of the characteristic of optical flow, there should not be any sudden change in the magnitude of optical flow values across the image. 
	Based on such motivation, we employ calibration masks to identify the abnormal change of optical flow values according to Eq. (4), such that the calculation of deblurring loss (Eq. (5)) will not be influenced by the inaccurately estimated optical flows.
	
	We also give examples of reblurred result $\hat{\bm{I}}_B$ and deformed ground-truth sharp image ${\bm{I}}_S^w$ in Fig.~\ref{fig:reblur}. It can be seen that that $\bm{I}_B$, $\hat{\bm{I}}$, $\hat{\bm{I}}_B$ and ${\bm{I}}_S^w$ are spatially consistent, while $\bm{I}_S$ has a misalignment with all of them.
	
	\section{Network Details}
	Our baseline deblurring network is designed as a U-Net structure, whose detailed structures are shown in Table~1. 
	The architecture details of $\mathcal{R}_{kpn}$ and $\mathcal{R}_{wpn}$ are presented in Table 2. 
	
	\section{Comparison with DRBNet}
	
	DRBNet adopts an extra light field dataset LFDOF in the training stage, which considerably improves the network performance. A comparison with DRBNet is provided in this section. PSNR/SSIM/LPIPS metrics are shown in Table~\ref{tab:drbnet}. 
	By adopting the same DPDD training set, Ours* is better than DRBNet on both RealDOF and DPDD. On RealDOF, it is reasonable that DRBNet with more training sets DPDD+LFDOF is better than Ours* only with DPDD training set.
	
	\begin{table*}[t]
	\centering
	\caption{\small{PSNR/SSIM/LPIPS comparison between DRBNet and ours.}}
	\label{tab:drbnet}
	\vspace{-0.2cm}
	\footnotesize
	\begin{tabular}{c|c|c|c}
		\toprule
		Method  &  Training set & RealDOF test set  & DPDD test set \\
		\midrule
		DRBNet &  DPDD + LFDOF & 25.75/0.771/0.257 & 25.73/0.791/0.183\\
		DRBNet & DPDD & 24.70/0.744/0.337 & 25.47/0.787/0.246 \\
		Ours &  DPDD & 24.87/0.748/0.288 & 25.73/0.792/0.232 \\
		\bottomrule
		\hline
	\end{tabular}
\vspace{-0.5em}
\end{table*}
	
	\section{More Results}
	In this section, more visual results on SDD, DPDD and RealDOF dataset are presented in Fig. $\sim$ Fig.~\ref{fig:compare}. It can be seen that the methods trained by our JDRL framework (UNet*, MPRNet* and IFAN*) outperform their vanilla counterparts (UNet, MPRNet and IFAN) in view of sharp textures and undistorted shapes. By using JDRL, the lines and contours of objects from the input images are better preserved in the deblurring output. The JDRL versions of DPDNet$_{S}$ and Son et al. are not provided because DPDNet$_{S}$ is basically a vanilla UNet and Son et al. have not released their training codes.
	
	\begin{table*}[]
		\centering
		\caption{The structure of baseline deblurring network. The structure of encoder is shown in the left column and the structure of decoder is shown in the right column. For convolution layers, kernel size=3. For maxpooling layers, kernel size=2, stride=2. For dropout layers, we set the random dropping probability as 0.4. When doing upsampling, the scale factor is set as 2.}
		\begin{tabular}{ccc||ccc}
			\hline
			\hhline{---||---}
			\multicolumn{3}{c||}{\textbf{Encoder}} & \multicolumn{3}{c}{\textbf{Decoder}}           \\ \hhline{---||---}
			Layer     & Output size & Filter      & Layer         & Output sizs & Filter           \\ \hhline{---||---}
			Conv,ReLU & $512\times512$     & $3\rightarrow64$      & Upsample,Conv & $32\times32$       & $1024\rightarrow512$       \\
			Conv,ReLU & $512\times512$     & $64\rightarrow64$     & Concat        & $32\times32$       & $(512+512)\rightarrow1024$ \\
			MaxPool   & $256\times256$     & -          & Conv,ReLU     & $32\times32$       & $1024\rightarrow512$       \\
			&             &             & Conv,ReLU     & $32\times32$       & $512\rightarrow512$        \\ \hhline{---||---}
			Conv,ReLU & $256\times256$     & $64\rightarrow128$    & Upsample,Conv & $128\times128$     & $512\rightarrow256$        \\
			Conv,ReLU & $256\times256$     & $128\rightarrow128$   & Concat        & $128\times128$     & $(256+256)\rightarrow512$  \\
			MaxPool   & $128\times128$     & -          & Conv,ReLU     & $128\times128$     & $512\rightarrow256$        \\
			&             &             & Conv,ReLU     & $128\times128$     & $256\rightarrow256$        \\ \hhline{---||---}
			Conv,ReLU & $128\times128$     & $128\rightarrow256$   & Upsample,Conv & $256\times256$     & $256\rightarrow128$        \\
			Conv,ReLU & $128\times128$     & $256\rightarrow256$   & Concat        & $256\times256$     & $(128+128)\rightarrow256$  \\
			MaxPool   & $64\times64$       & -          & Conv,ReLU     & $256\times256$     & $256\rightarrow128$        \\
			&             &             & Conv,ReLU     & $256\times256$     & $128\rightarrow128$        \\ \hhline{---||---}
			Conv,ReLU & $64\times64$       & $256\rightarrow512$   & Upsample,Conv & $512\times512$     & $128\rightarrow64$         \\
			Conv,ReLU & $64\times64$       & $256\rightarrow512$   & Concat        & $512\times512$     & $(64+64)\rightarrow128$    \\
			Dropout   & $64\times64$       & -          & Conv,ReLU     & $512\times512$     & $128\rightarrow64$         \\
			MaxPool   & $32\times32$       & -          & Conv,ReLU     & $512\times512$     & $64\rightarrow64$          \\ \hhline{---||---}
			Conv,ReLU & $32\times32$       & $512\rightarrow1024$  & Conv,Sigmoid  & $512\times512$     & $64\rightarrow3$           \\
			Conv,ReLU & $32\times32$       & $1024\rightarrow1024$ &               &             &                  \\
			Dropout   & $32\times32$       & -          &               &             &                  \\ \hhline{---||---}\hline
		\end{tabular}
		\vspace{2mm}
		\label{tab:structure}
	\end{table*}
	
	\begin{table*}[]
		\centering
		\caption{The architecture of $\mathcal{R}_{kpn}$ and $\mathcal{R}_{wpn}$.     }
		\begin{tabular}{cc|cc||cc|cc}
			\hline
			\hhline{----||----}
			\multicolumn{4}{c||}{\textbf{$\mathcal{R}_{kpn}$}} & \multicolumn{4}{c}{\textbf{$\mathcal{R}_{wpn}$}}           \\ \hhline{----||----}
			\multicolumn{2}{c}{Layer}     & Output size & Filter      & \multicolumn{2}{c}{Layer}         & Output sizs & Filter           \\ \hhline{----||----}
			&Conv & $512\times512$     & $6\rightarrow64$      & &Conv & $512\times512$       & $6\rightarrow64$       \\
			\hline
			\multirow{2}{*}{Resblock} &\multicolumn{1}{|c|}{Conv, Relu} & $512\times512$    &  $64\rightarrow64$   & \multirow{2}{*}{Resblock} &\multicolumn{1}{|c|}{Conv, Relu} & $512\times512$      & $64\rightarrow64$       \\
			&\multicolumn{1}{|c|}{Conv} & $512\times512$    &  $64\rightarrow64$   &  &\multicolumn{1}{|c|}{Conv}& $512\times512$      & $64\rightarrow64$       \\
			\hline
			\multirow{2}{*}{Resblock} &\multicolumn{1}{|c|}{Conv, Relu} & $512\times512$    &  $64\rightarrow64$   & \multirow{2}{*}{Resblock} &\multicolumn{1}{|c|}{Conv, Relu} & $512\times512$      & $64\rightarrow64$       \\
			&\multicolumn{1}{|c|}{Conv} & $512\times512$    &  $64\rightarrow64$   &  &\multicolumn{1}{|c|}{Conv}& $512\times512$      & $64\rightarrow64$       \\
			\hline
			\multirow{2}{*}{Resblock} &\multicolumn{1}{|c|}{Conv, Relu} & $512\times512$    &  $64\rightarrow64$   & \multirow{2}{*}{Resblock} &\multicolumn{1}{|c|}{Conv, Relu} & $512\times512$      & $64\rightarrow64$       \\
			&\multicolumn{1}{|c|}{Conv} & $512\times512$    &  $64\rightarrow64$   &  &\multicolumn{1}{|c|}{Conv}& $512\times512$      & $64\rightarrow64$       \\
			\hline
			&Conv & $512\times512$     & $64\rightarrow35$      & &Conv & $512\times512$       & $64\rightarrow8$       \\ \hhline{----||----}\hline
		\end{tabular}
		\vspace{2mm}
		\label{tab:structure}
	\end{table*}
	
	\begin{figure*}[htbp]
		\centering
		\subfigure[Input]{
			\begin{minipage}[t]{0.24\linewidth}
				\centering
				\includegraphics[width=1\linewidth]{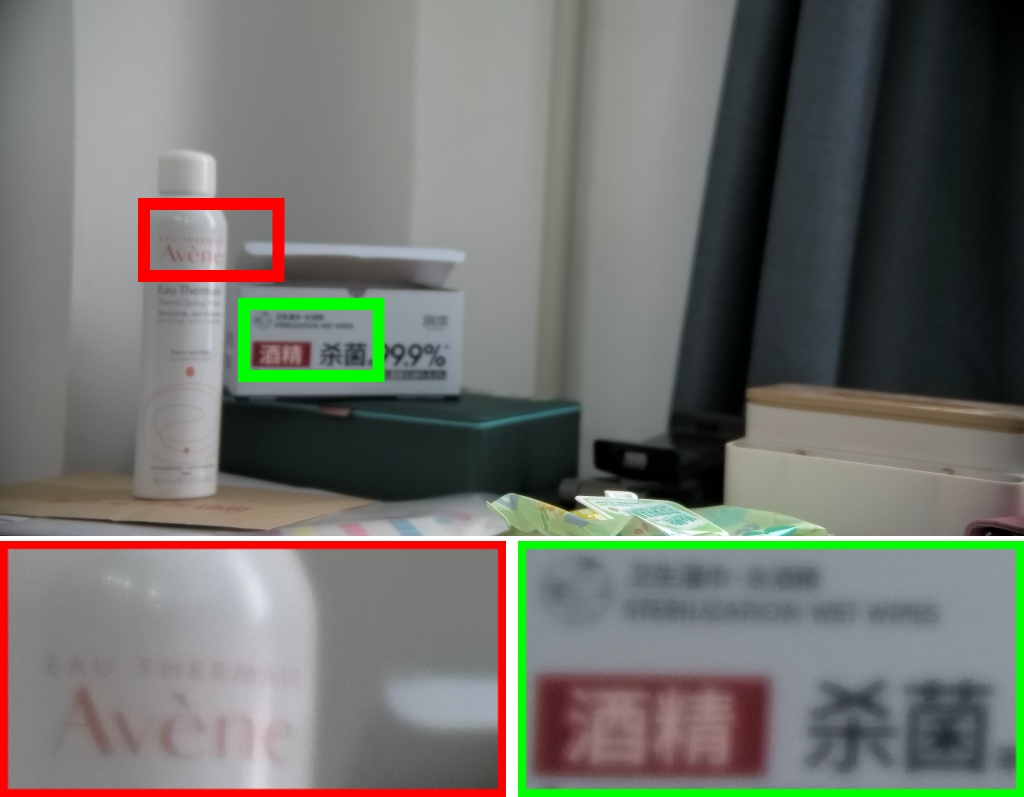}
			\end{minipage}%
			\hspace{0.04cm}
		}%
		\subfigure[UNet]{
			\begin{minipage}[t]{0.24\linewidth}
				\centering
				\includegraphics[width=1\linewidth]{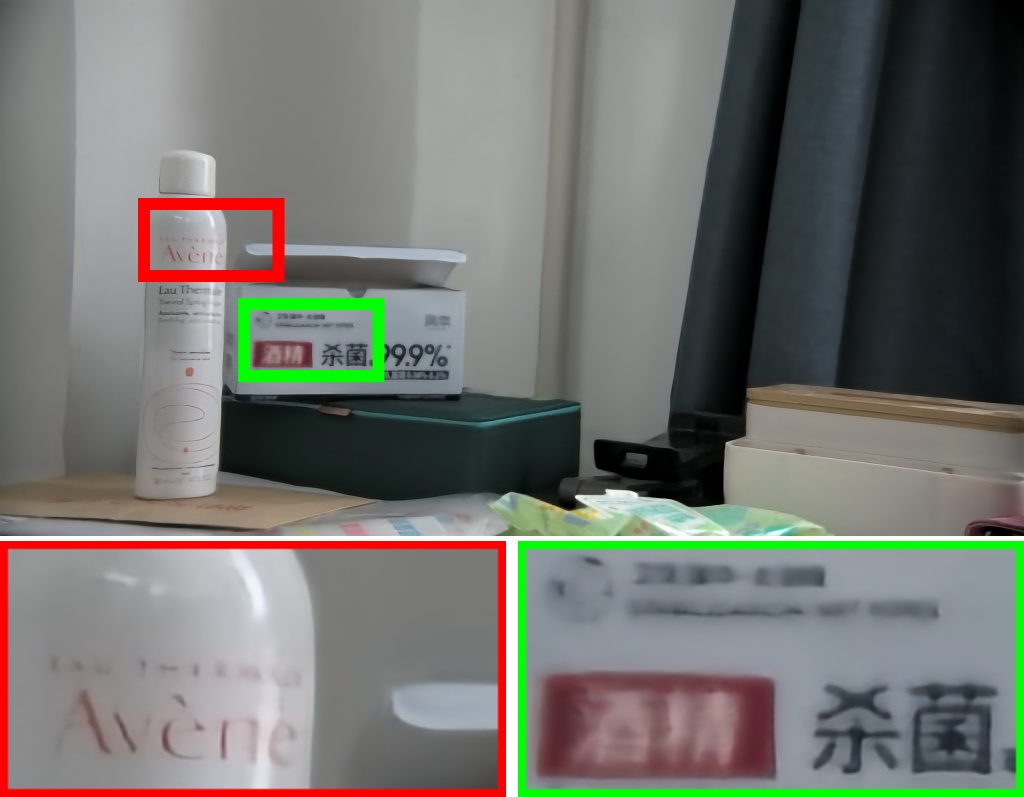}
			\end{minipage}%
			\hspace{0.04cm}
		}%
		\subfigure[MPRNet]{
			\begin{minipage}[t]{0.24\linewidth}
				\centering
				\includegraphics[width=1\linewidth]{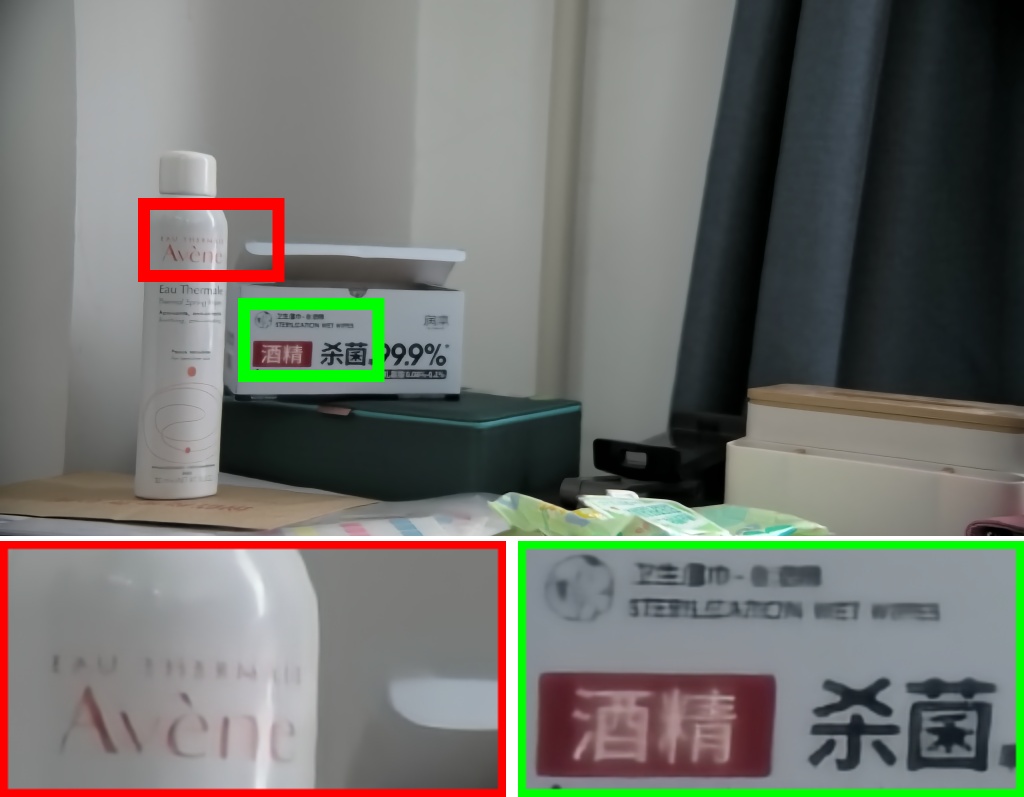}
			\end{minipage}
		}%
		\subfigure[DPDNet$_{S}$]{
			\begin{minipage}[t]{0.24\linewidth}
				\centering
				\includegraphics[width=1\linewidth]{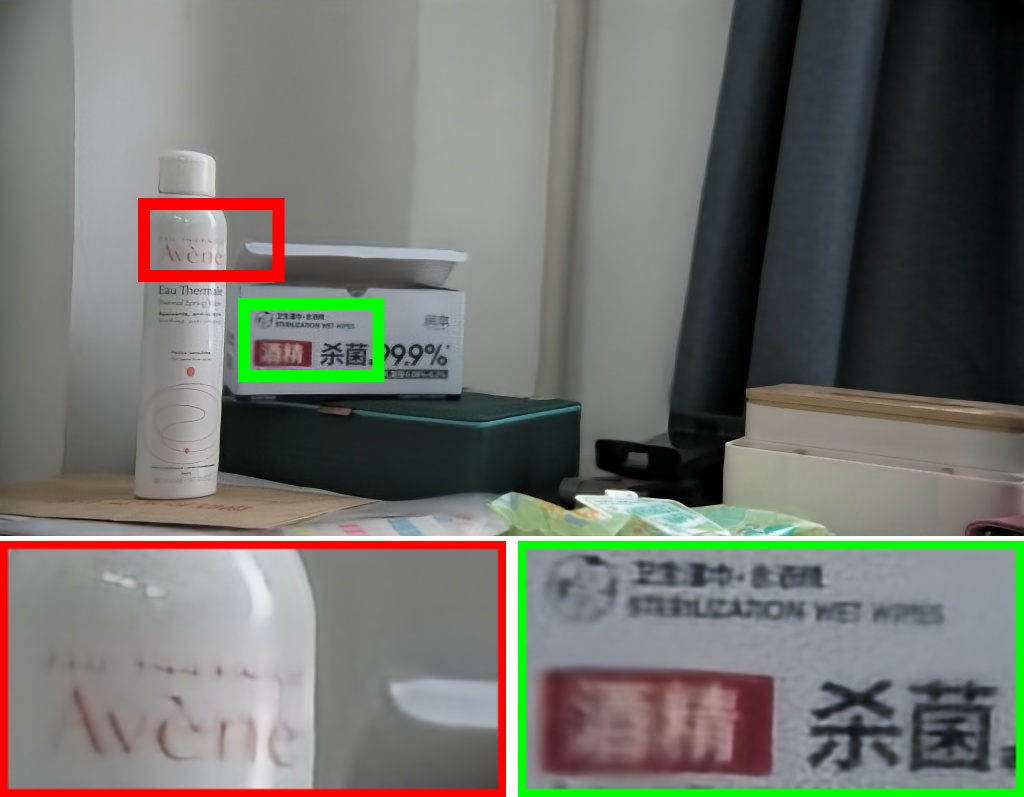}
			\end{minipage}
		}%
		\quad
		\subfigure[DMPHN]{
			\begin{minipage}[t]{0.24\linewidth}
				\centering
				\includegraphics[width=1\linewidth]{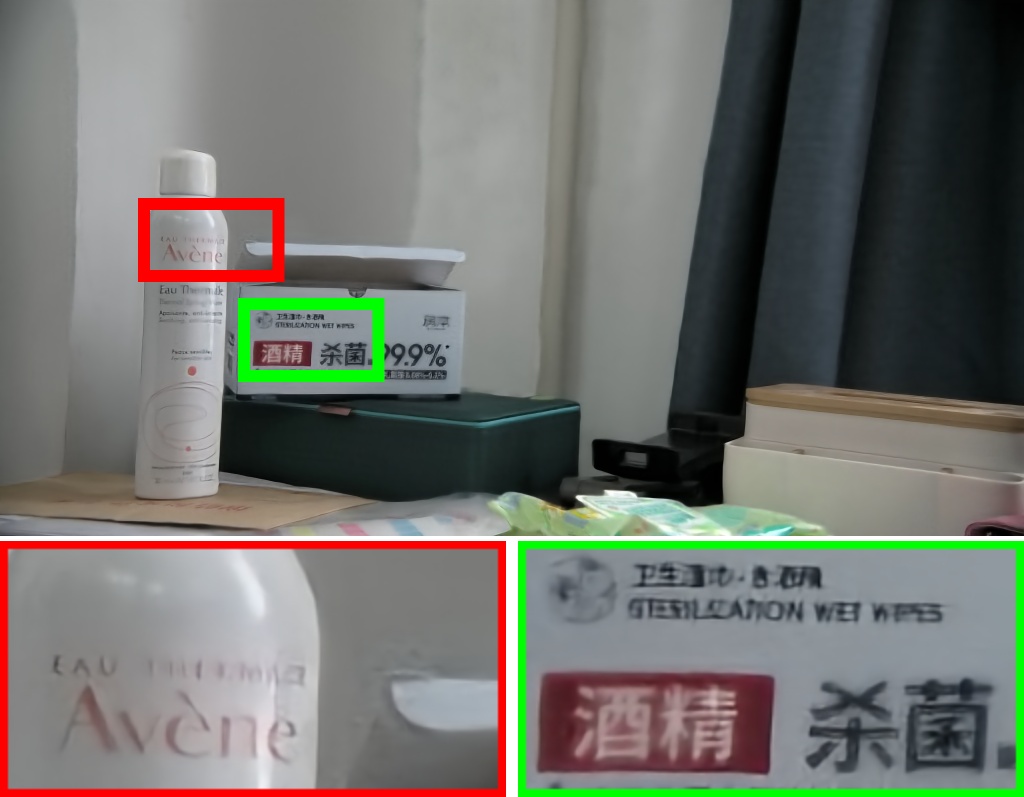}
			\end{minipage}%
			\hspace{0.04cm}
		}%
		\subfigure[UNet*]{
			\begin{minipage}[t]{0.24\linewidth}
				\centering
				\includegraphics[width=1\linewidth]{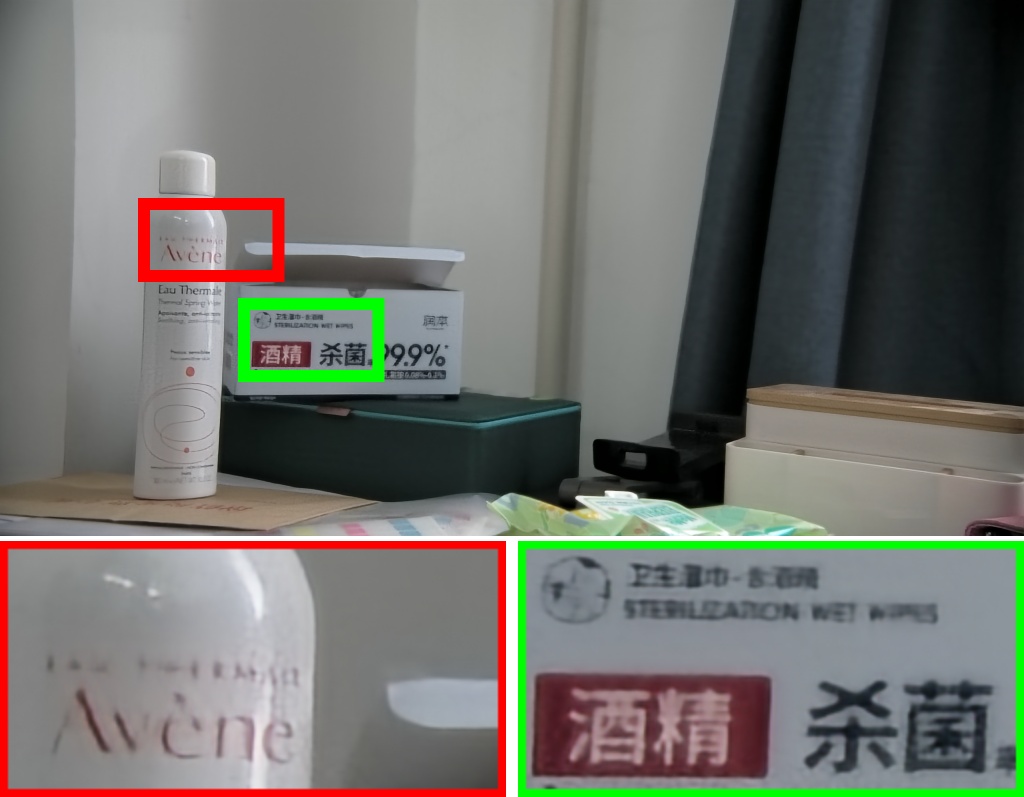}
			\end{minipage}%
			\hspace{0.04cm}
		}%
		\subfigure[MPRNet*]{
			\begin{minipage}[t]{0.24\linewidth}
				\centering
				\includegraphics[width=1\linewidth]{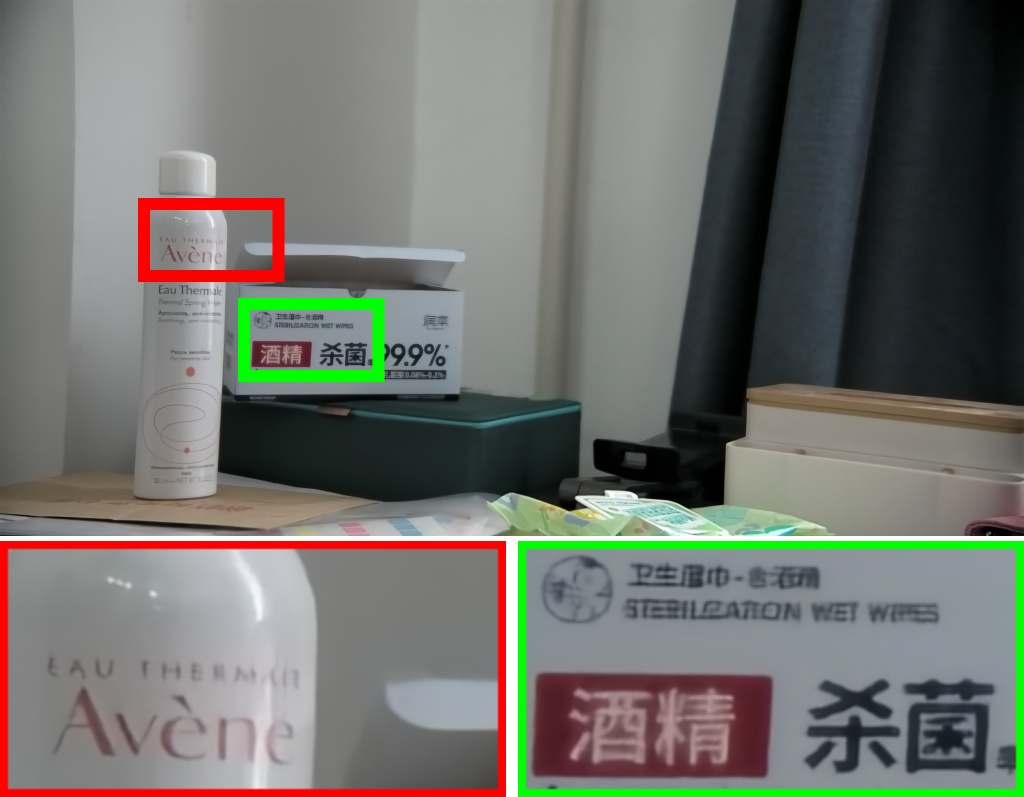}
			\end{minipage}
		}%
		\subfigure[GT]{
			\begin{minipage}[t]{0.24\linewidth}
				\centering
				\includegraphics[width=1\linewidth]{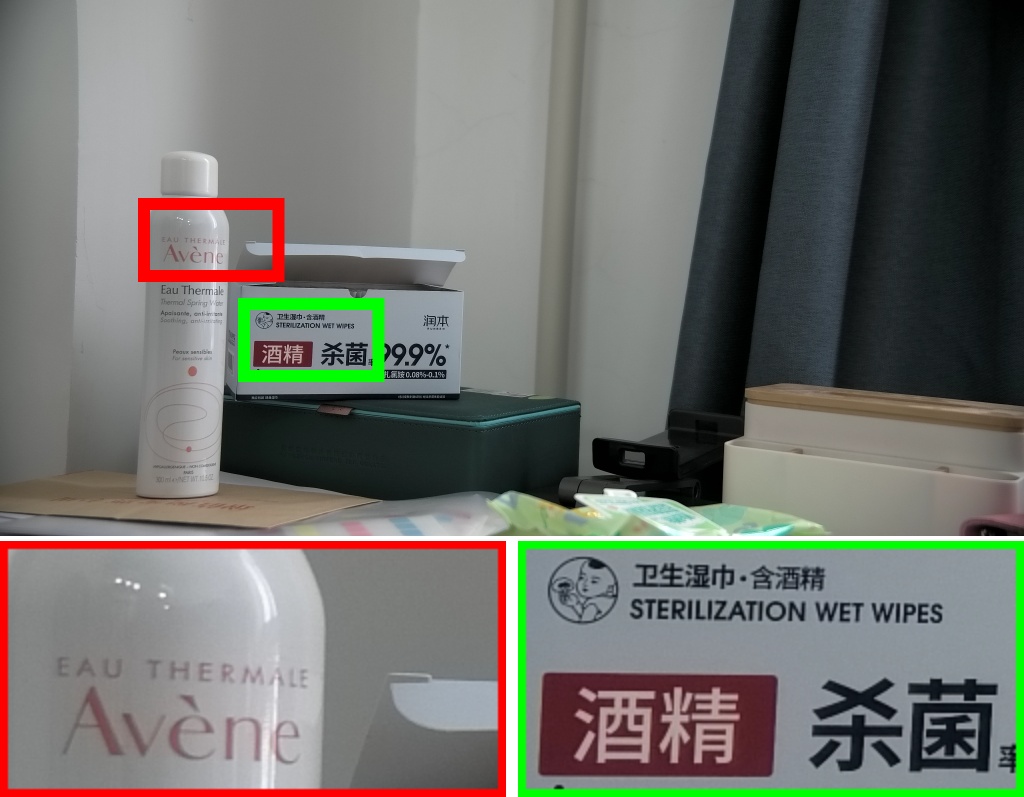}
			\end{minipage}
		}%
		\centering
		\label{sidd1}
		\caption{Visual comparison between different methods on SDD dataset.}
	\end{figure*}
	\begin{figure*}[htbp]
		\centering
		\subfigure[Input]{
			\begin{minipage}[t]{0.24\linewidth}
				\centering
				\includegraphics[width=1\linewidth]{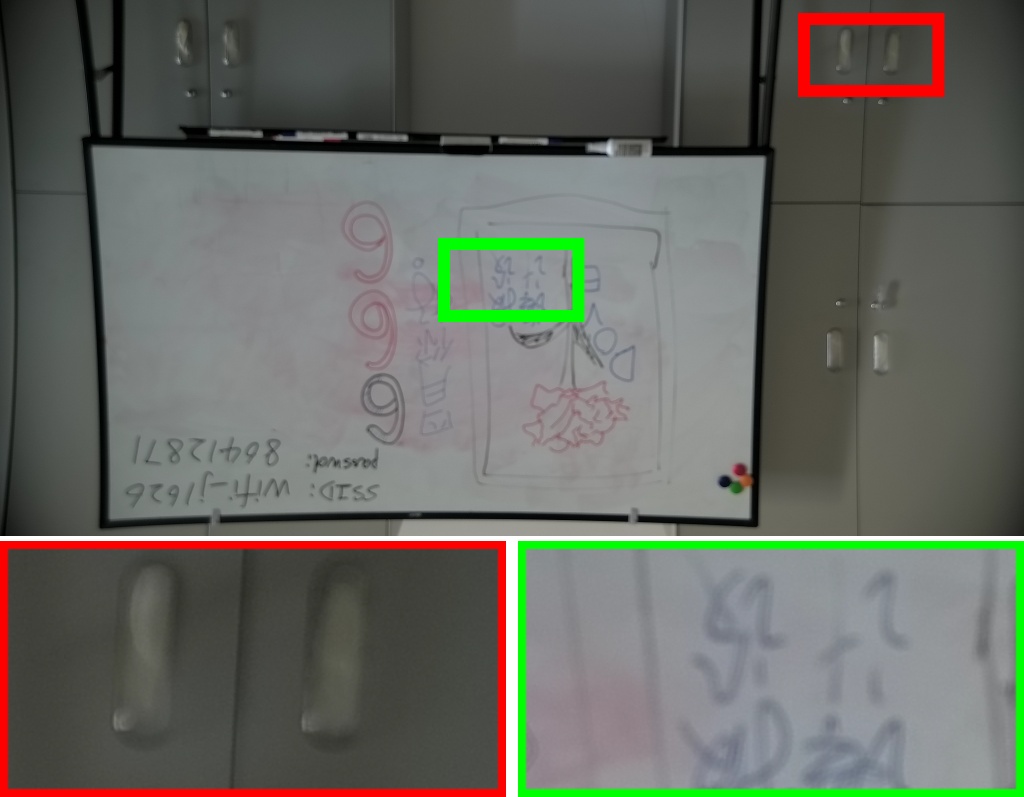}
			\end{minipage}%
			\hspace{0.04cm}
		}%
		\subfigure[UNet]{
			\begin{minipage}[t]{0.24\linewidth}
				\centering
				\includegraphics[width=1\linewidth]{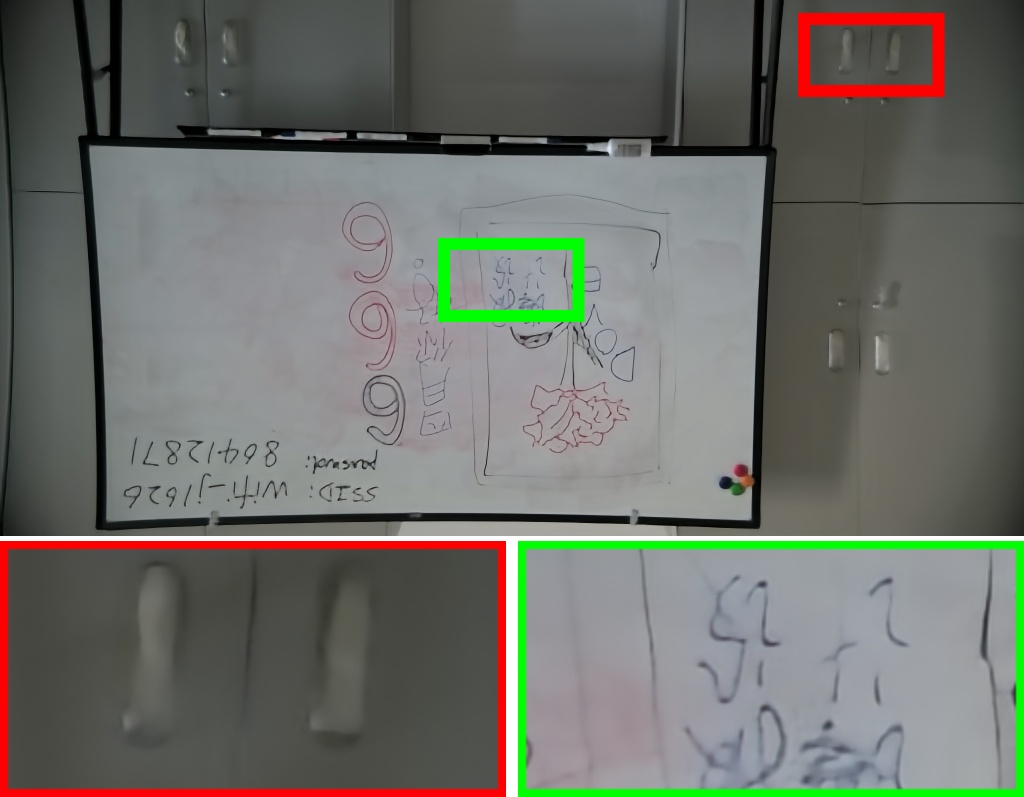}
			\end{minipage}%
			\hspace{0.04cm}
		}%
		\subfigure[MPRNet]{
			\begin{minipage}[t]{0.24\linewidth}
				\centering
				\includegraphics[width=1\linewidth]{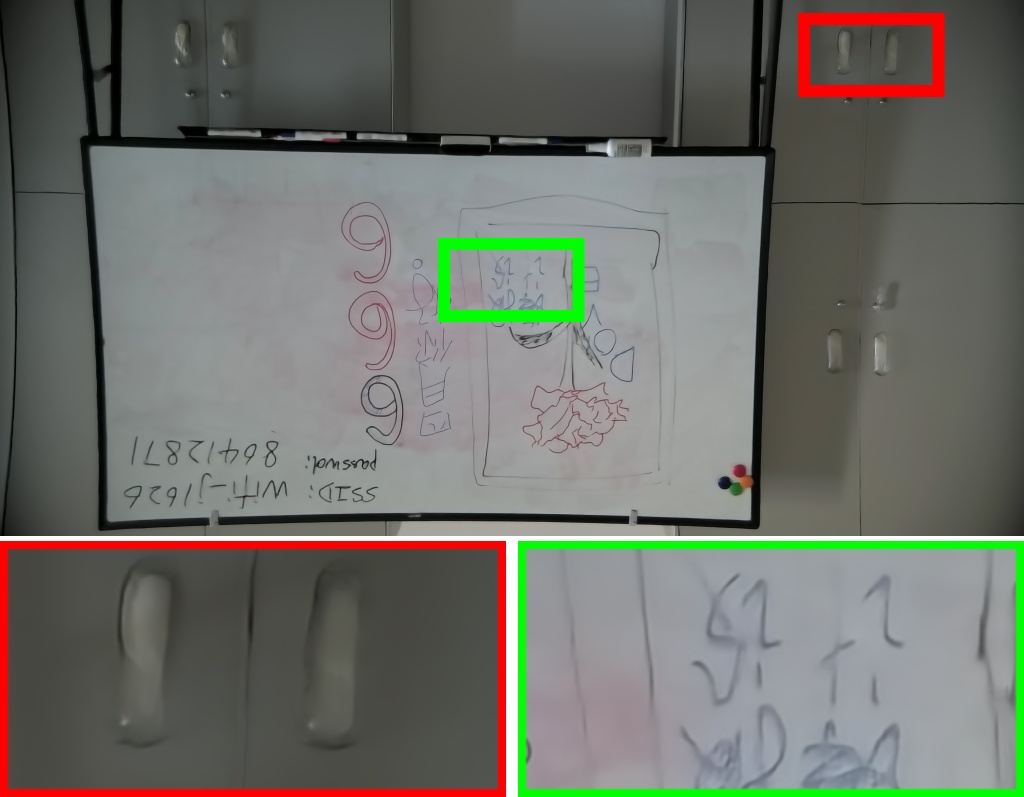}
			\end{minipage}
		}%
		\subfigure[DPDNet$_{S}$]{
			\begin{minipage}[t]{0.24\linewidth}
				\centering
				\includegraphics[width=1\linewidth]{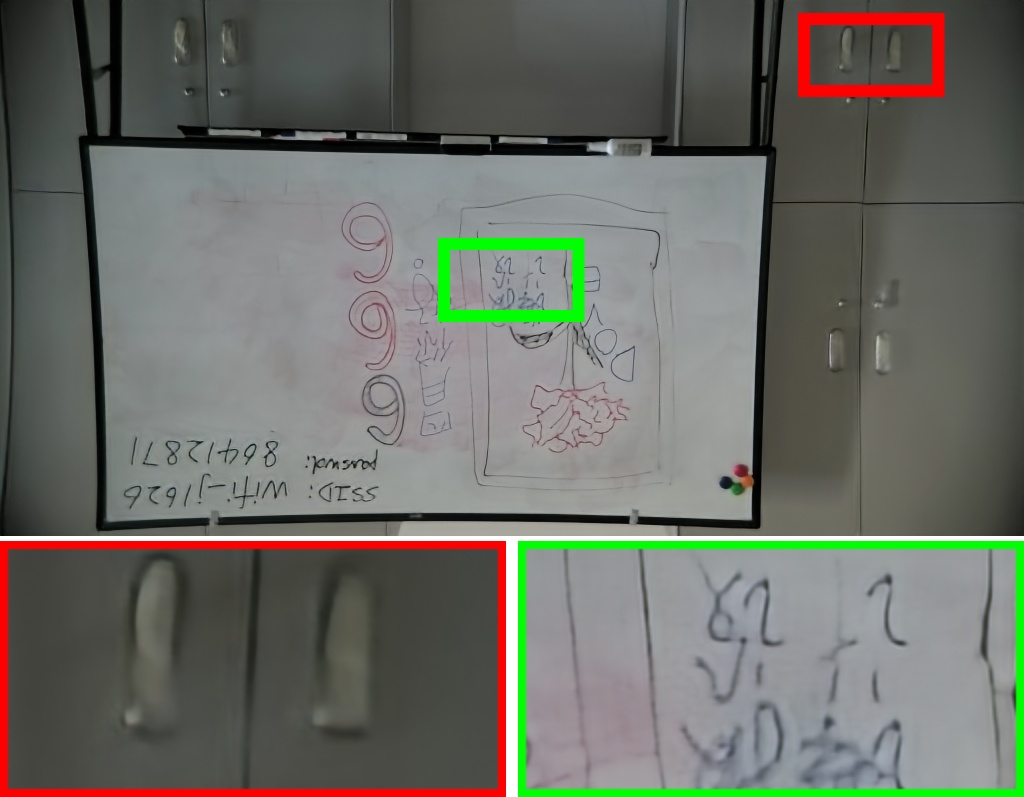}
			\end{minipage}
		}%
		\quad
		\subfigure[DMPHN]{
			\begin{minipage}[t]{0.24\linewidth}
				\centering
				\includegraphics[width=1\linewidth]{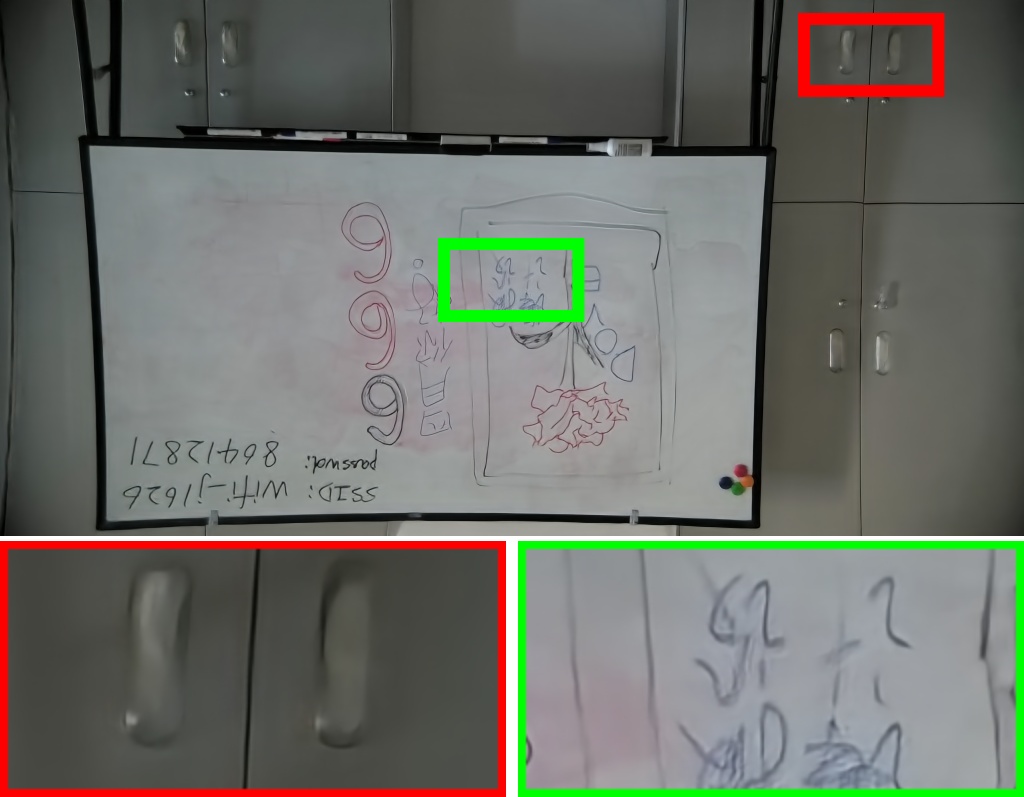}
			\end{minipage}%
			\hspace{0.04cm}
		}%
		\subfigure[UNet*]{
			\begin{minipage}[t]{0.24\linewidth}
				\centering
				\includegraphics[width=1\linewidth]{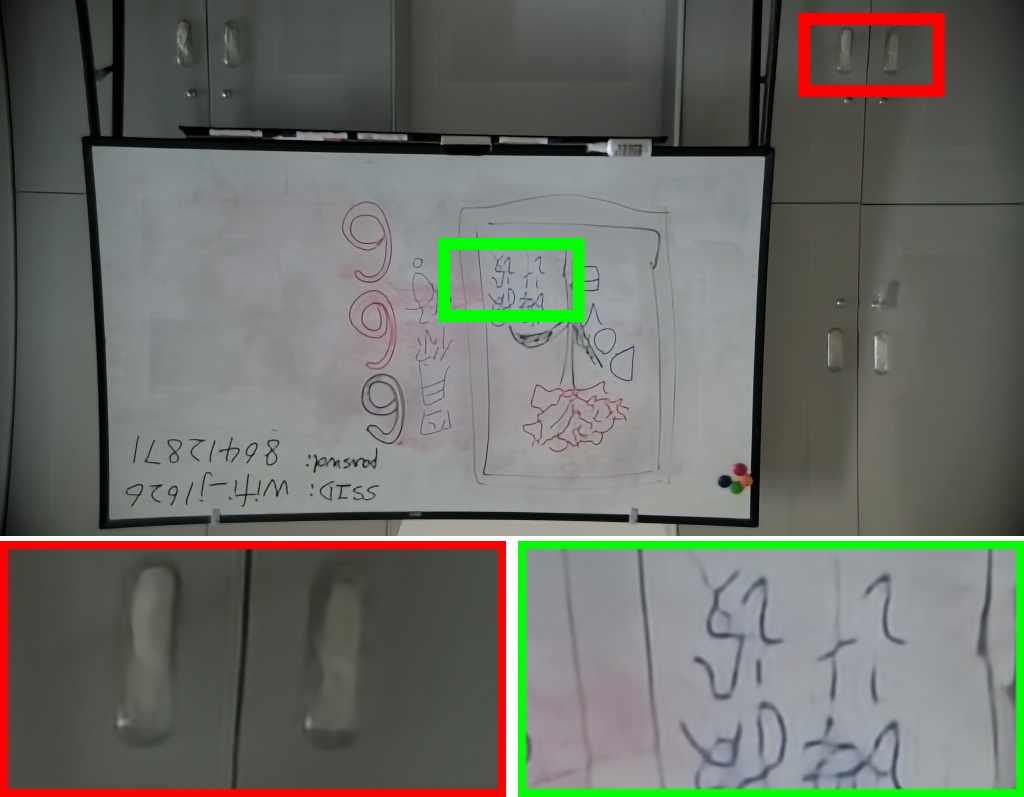}
			\end{minipage}%
			\hspace{0.04cm}
		}%
		\subfigure[MPRNet*]{
			\begin{minipage}[t]{0.24\linewidth}
				\centering
				\includegraphics[width=1\linewidth]{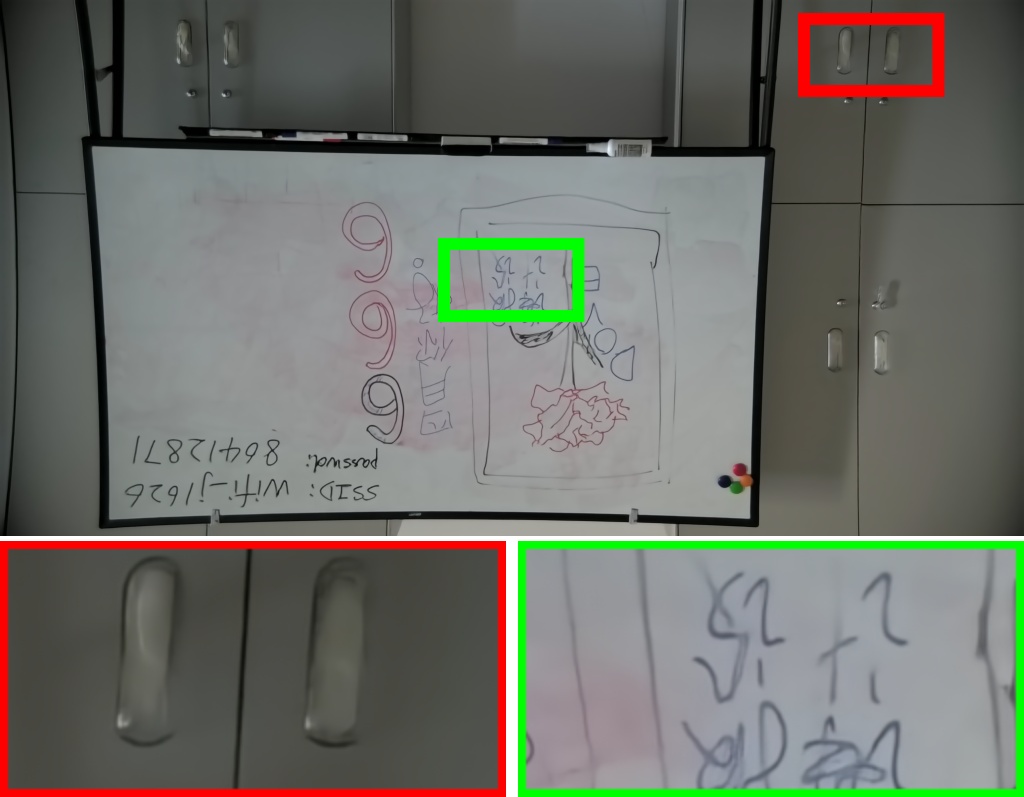}
			\end{minipage}
		}%
		\subfigure[GT]{
			\begin{minipage}[t]{0.24\linewidth}
				\centering
				\includegraphics[width=1\linewidth]{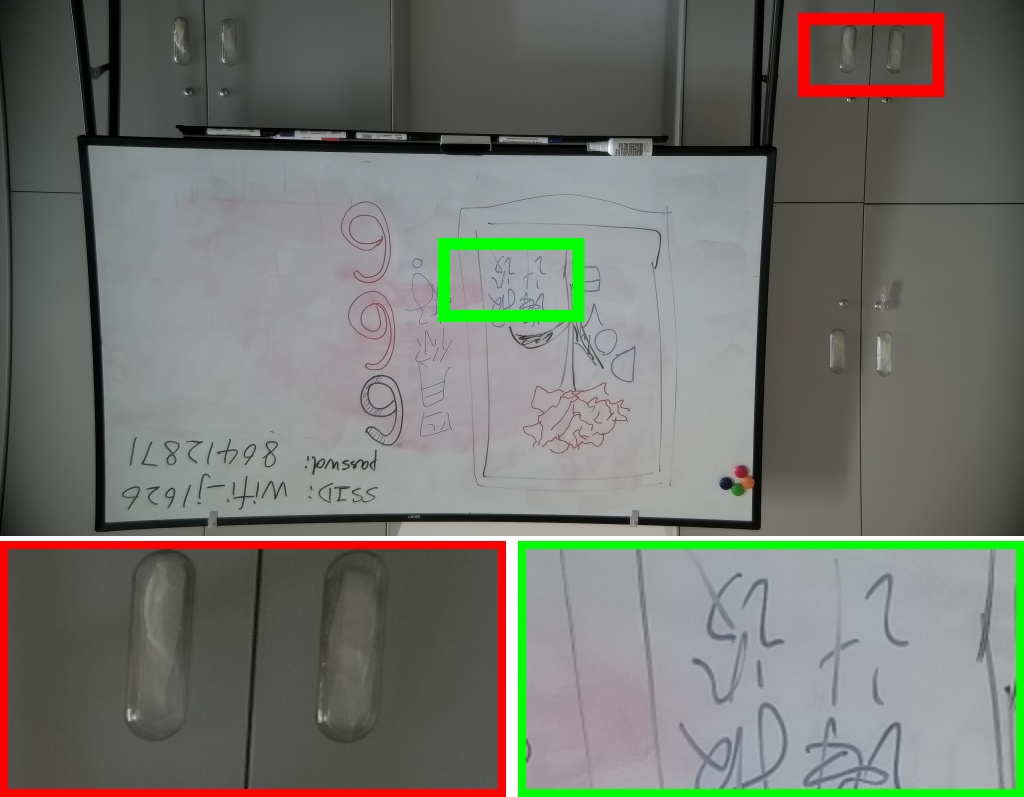}
			\end{minipage}
		}%
		\centering
		\caption{Visual comparison between different methods on SDD dataset.}
	\end{figure*}
	\vspace{-2mm}
	\begin{figure*}[htbp]
		\centering
		\subfigure[Input]{
			\begin{minipage}[t]{0.24\linewidth}
				\centering
				\includegraphics[width=1\linewidth]{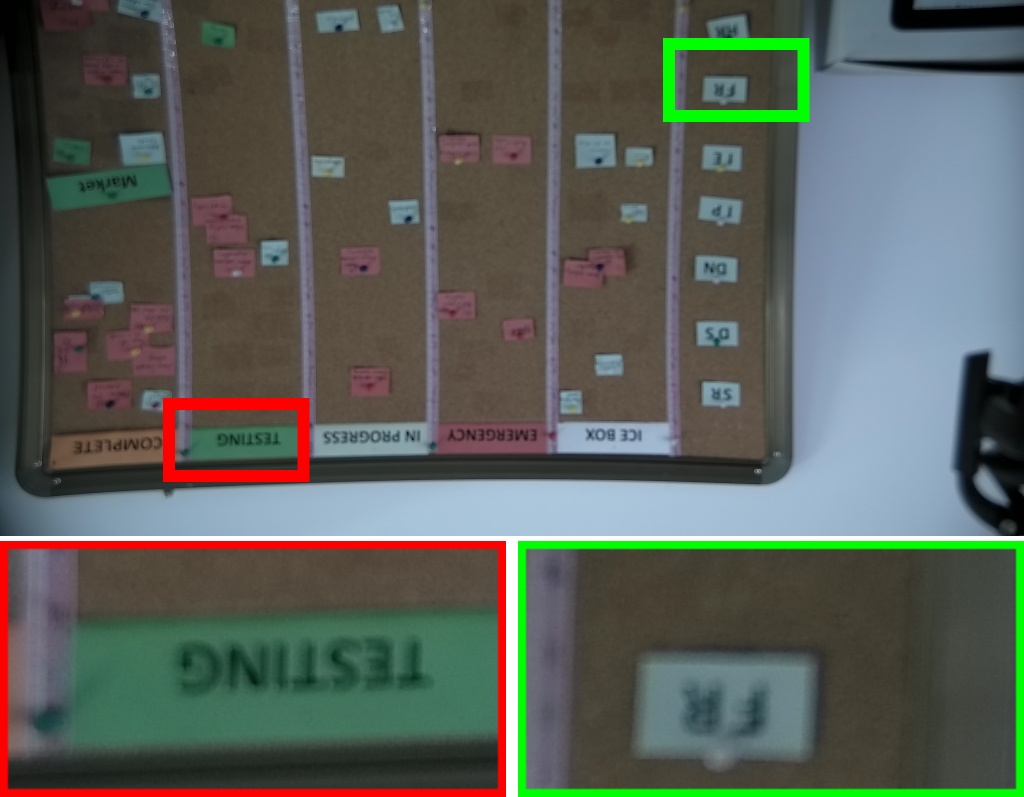}
			\end{minipage}%
			\hspace{0.04cm}
		}%
		\subfigure[UNet]{
			\begin{minipage}[t]{0.24\linewidth}
				\centering
				\includegraphics[width=1\linewidth]{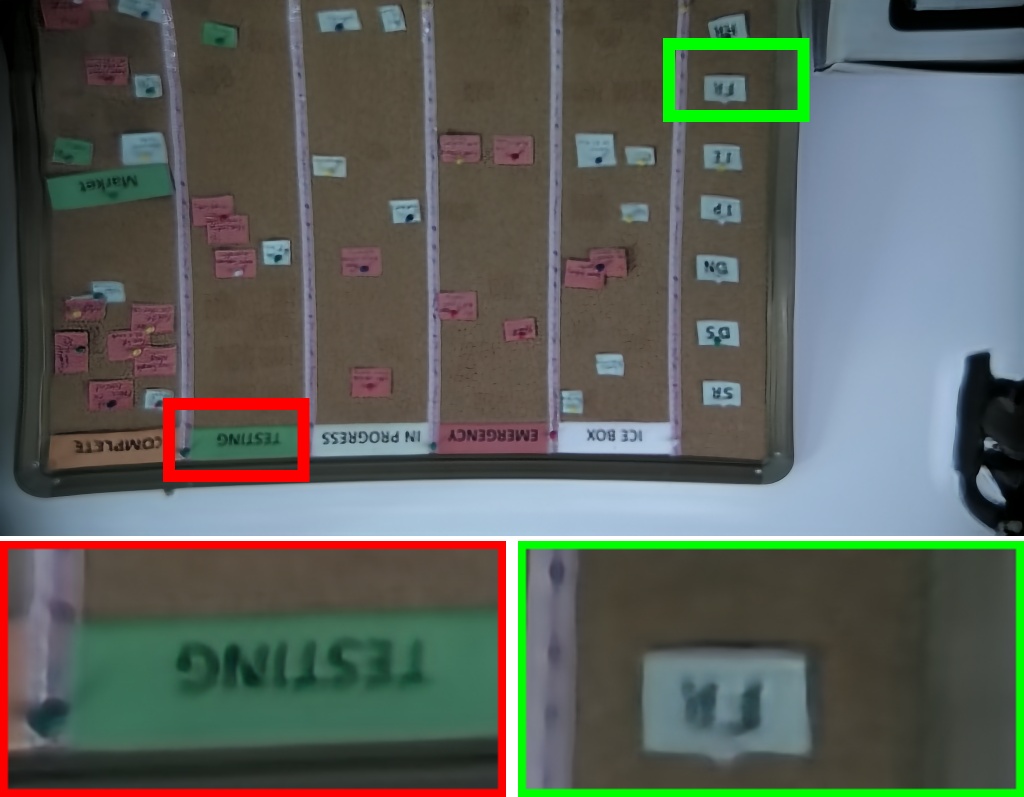}
			\end{minipage}%
			\hspace{0.04cm}
		}%
		\subfigure[MPRNet]{
			\begin{minipage}[t]{0.24\linewidth}
				\centering
				\includegraphics[width=1\linewidth]{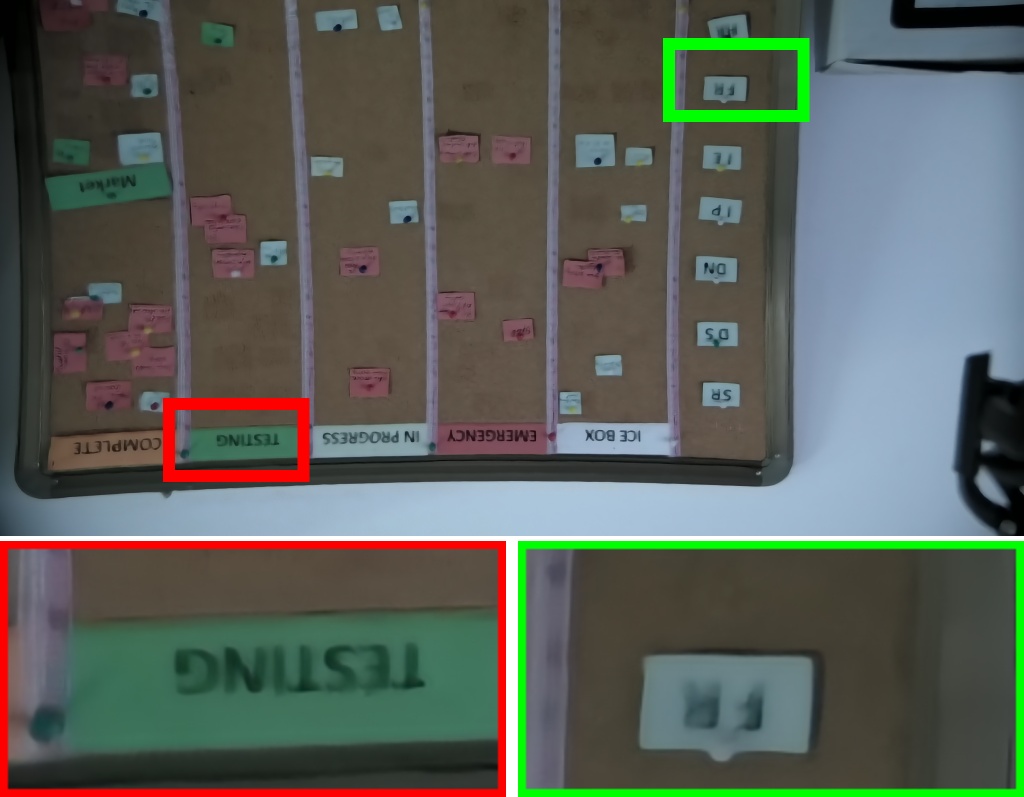}
			\end{minipage}
		}%
		\subfigure[DPDNet$_{S}$]{
			\begin{minipage}[t]{0.24\linewidth}
				\centering
				\includegraphics[width=1\linewidth]{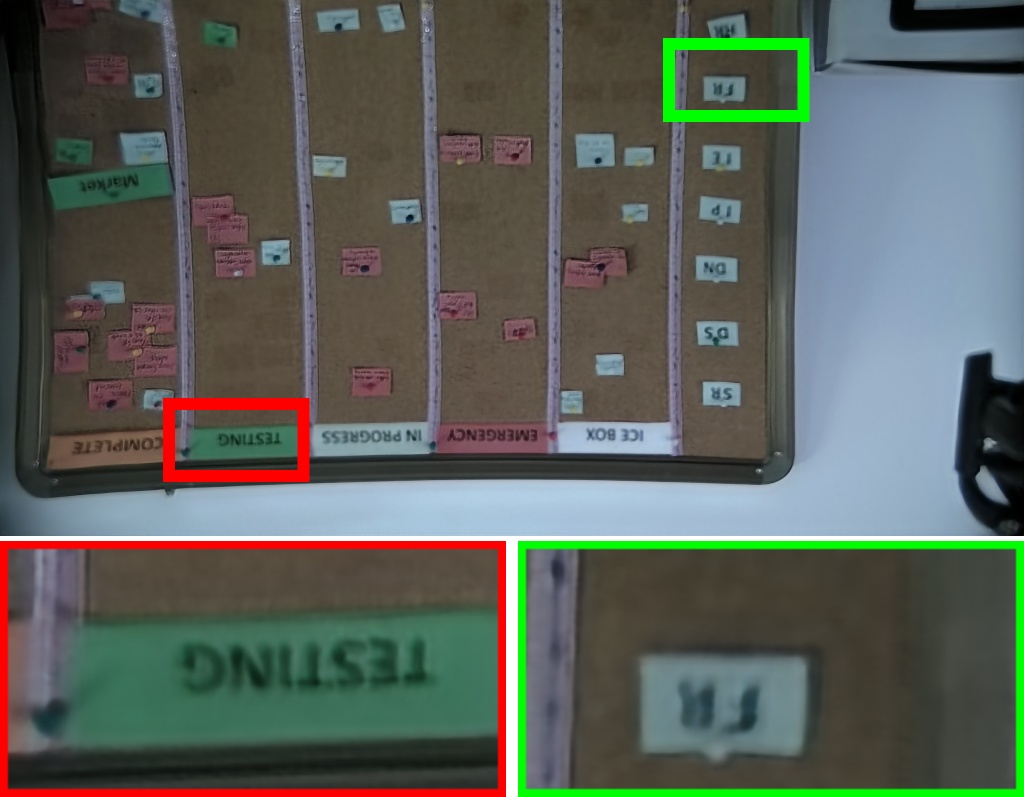}
			\end{minipage}
		}%
		\quad
		\subfigure[DMPHN]{
			\begin{minipage}[t]{0.24\linewidth}
				\centering
				\includegraphics[width=1\linewidth]{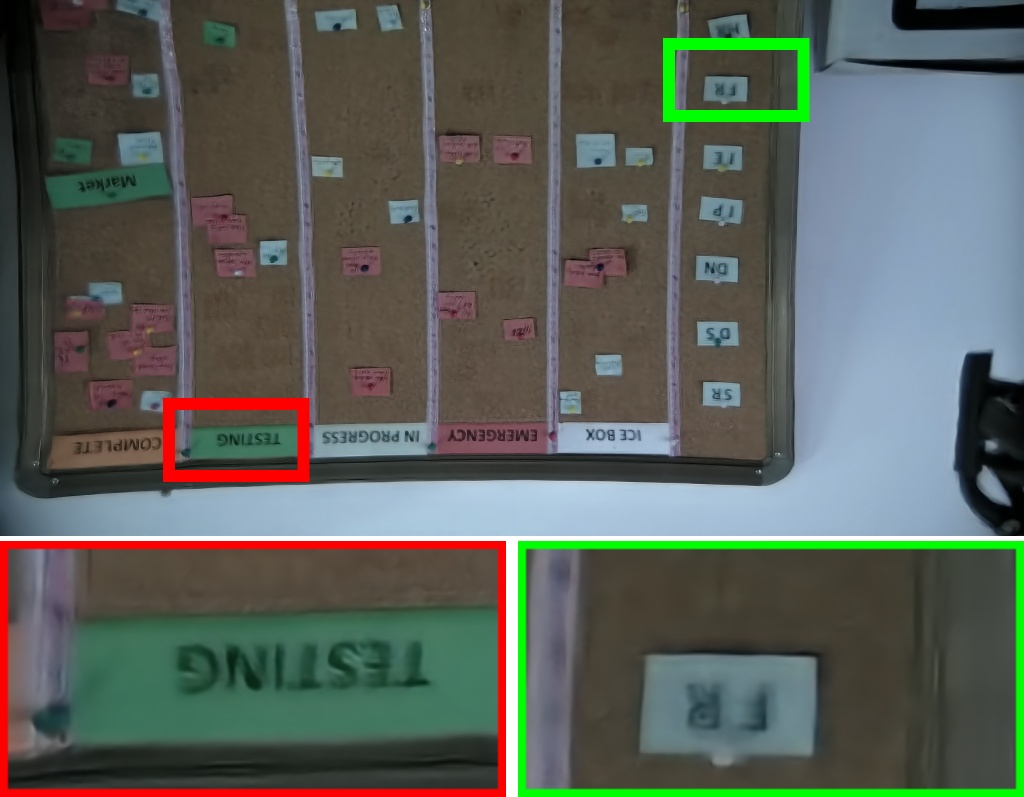}
			\end{minipage}%
			\hspace{0.04cm}
		}%
		\subfigure[UNet*]{
			\begin{minipage}[t]{0.24\linewidth}
				\centering
				\includegraphics[width=1\linewidth]{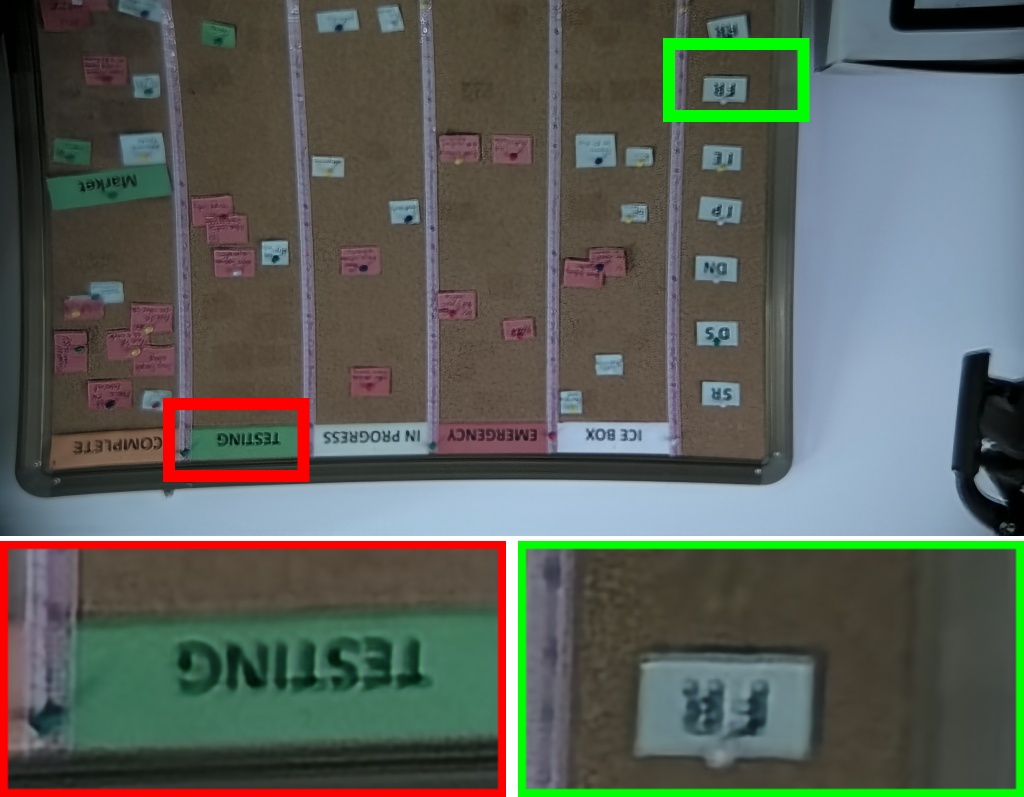}
			\end{minipage}%
			\hspace{0.04cm}
		}%
		\subfigure[MPRNet*]{
			\begin{minipage}[t]{0.24\linewidth}
				\centering
				\includegraphics[width=1\linewidth]{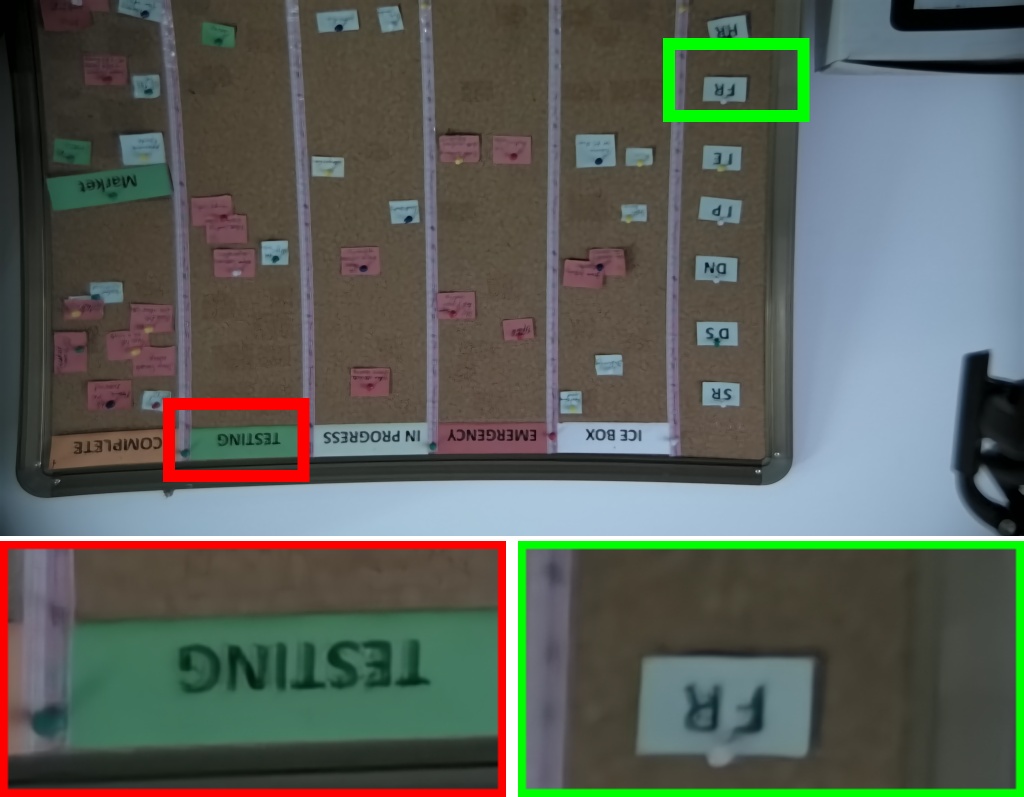}
			\end{minipage}
		}%
		\subfigure[GT]{
			\begin{minipage}[t]{0.24\linewidth}
				\centering
				\includegraphics[width=1\linewidth]{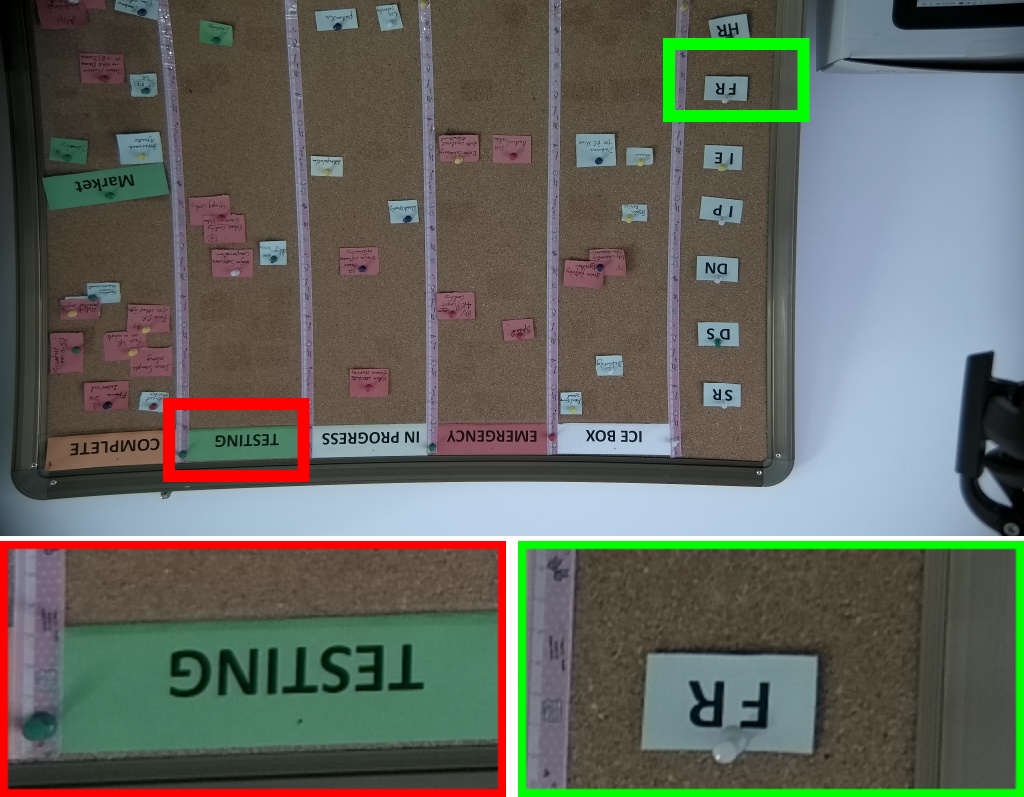}
			\end{minipage}
		}%
		\centering
		\caption{Visual comparison between different methods on SDD dataset.}
	\end{figure*}
	
	\begin{figure*}[htbp]
		\centering
		\subfigure[Input]{
			\begin{minipage}[t]{0.24\linewidth}
				\centering
				\includegraphics[width=1\linewidth]{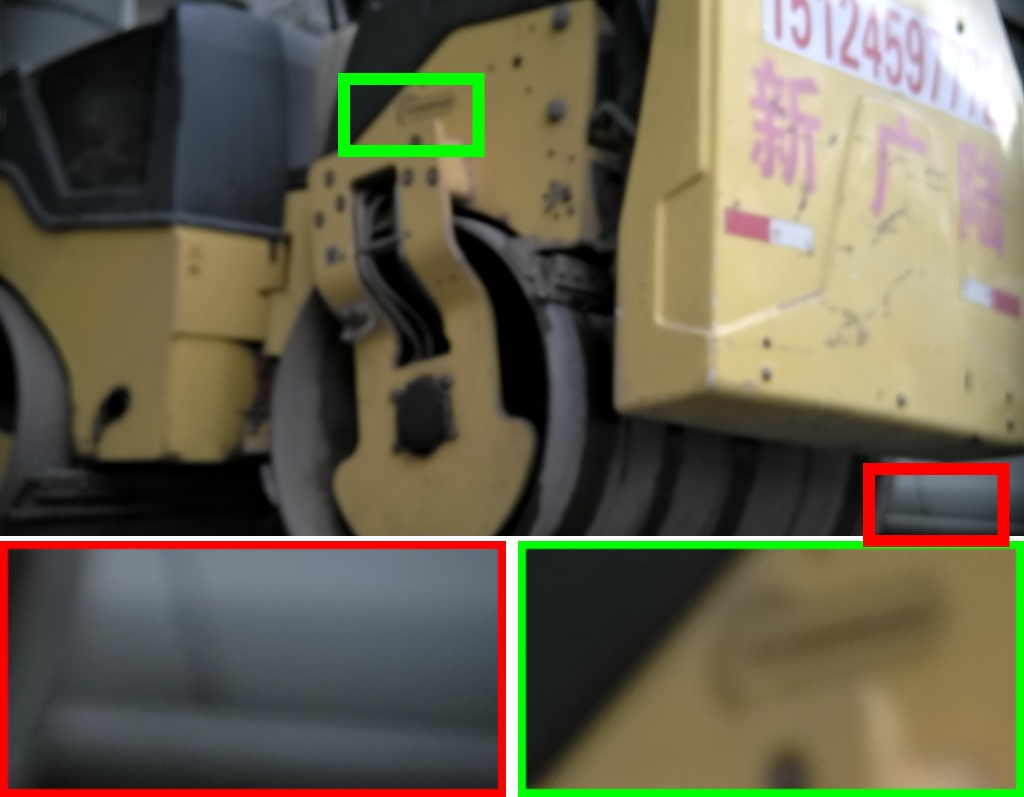}
			\end{minipage}%
			\hspace{0.04cm}
		}%
		\subfigure[UNet]{
			\begin{minipage}[t]{0.24\linewidth}
				\centering
				\includegraphics[width=1\linewidth]{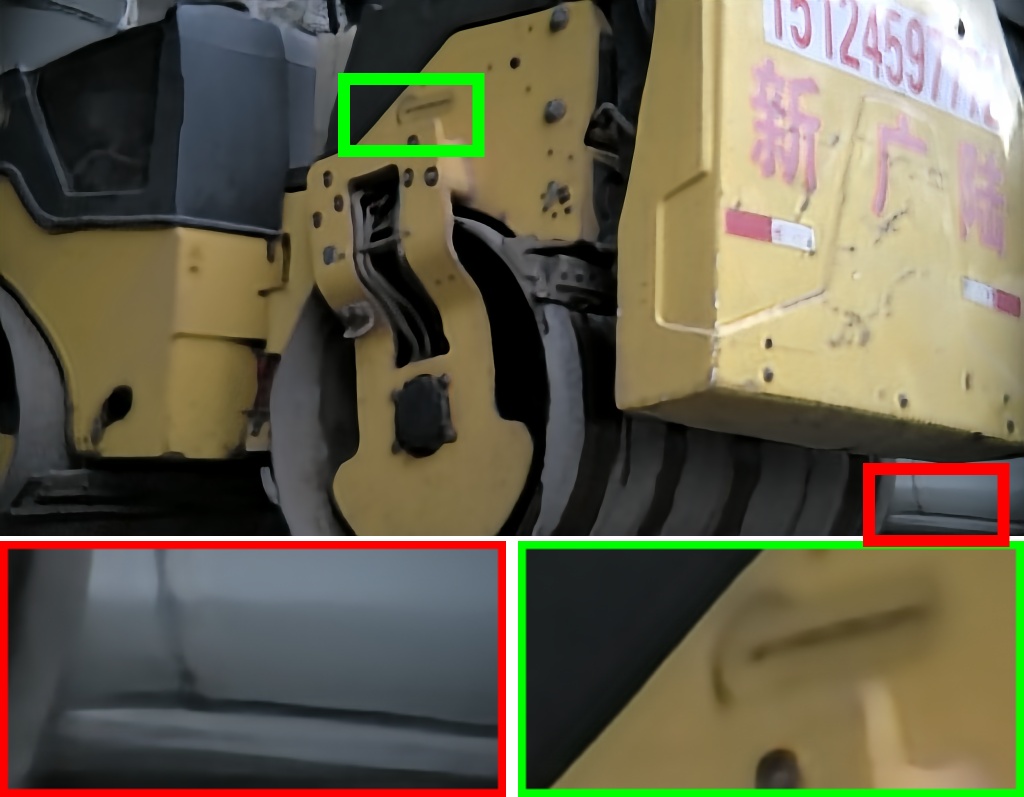}
			\end{minipage}%
			\hspace{0.04cm}
		}%
		\subfigure[MPRNet]{
			\begin{minipage}[t]{0.24\linewidth}
				\centering
				\includegraphics[width=1\linewidth]{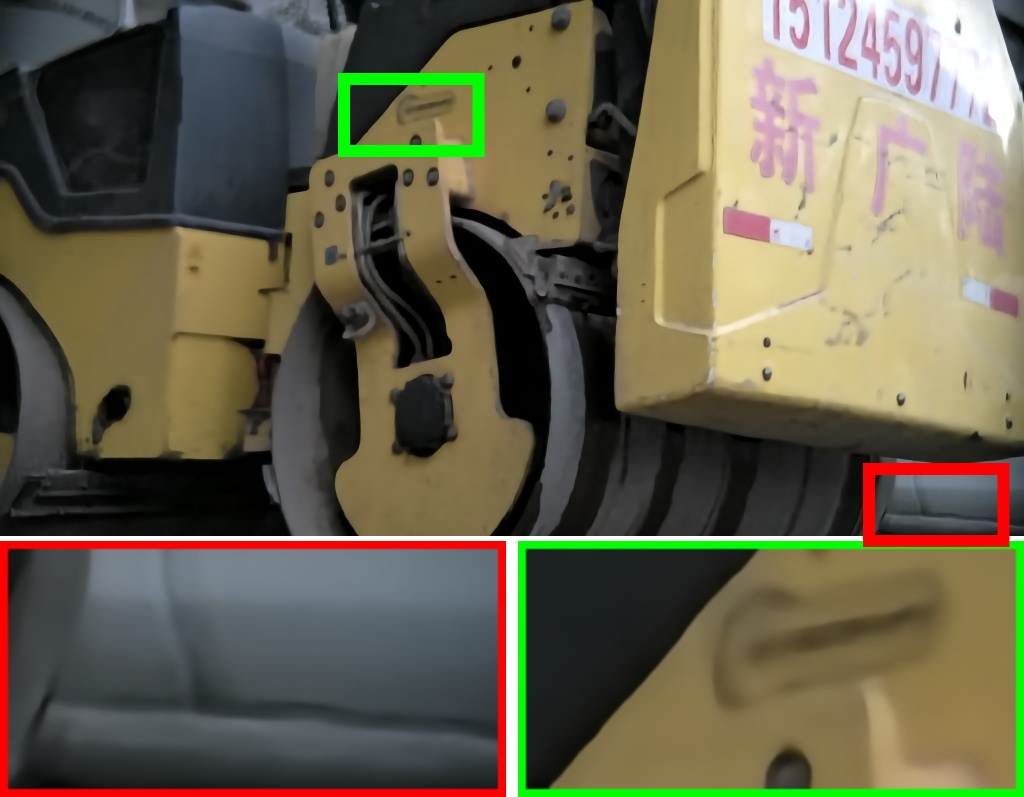}
			\end{minipage}
		}%
		\subfigure[DPDNet$_{S}$]{
			\begin{minipage}[t]{0.24\linewidth}
				\centering
				\includegraphics[width=1\linewidth]{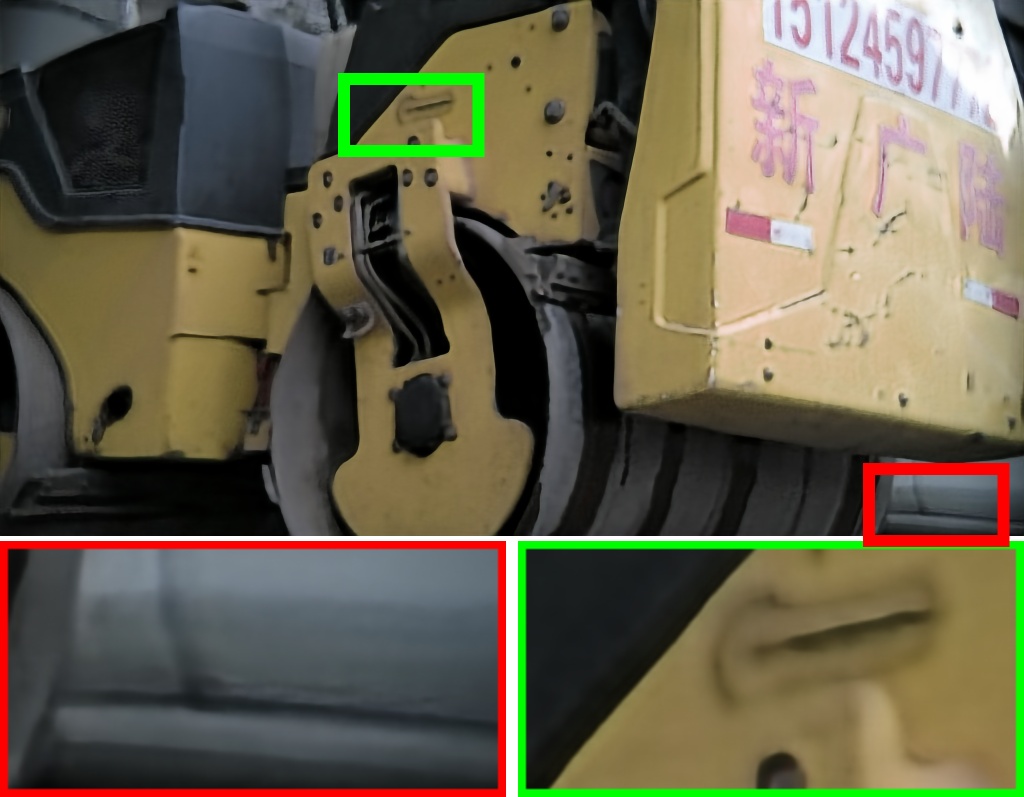}
			\end{minipage}
		}%
		\quad
		\subfigure[DMPHN]{
			\begin{minipage}[t]{0.24\linewidth}
				\centering
				\includegraphics[width=1\linewidth]{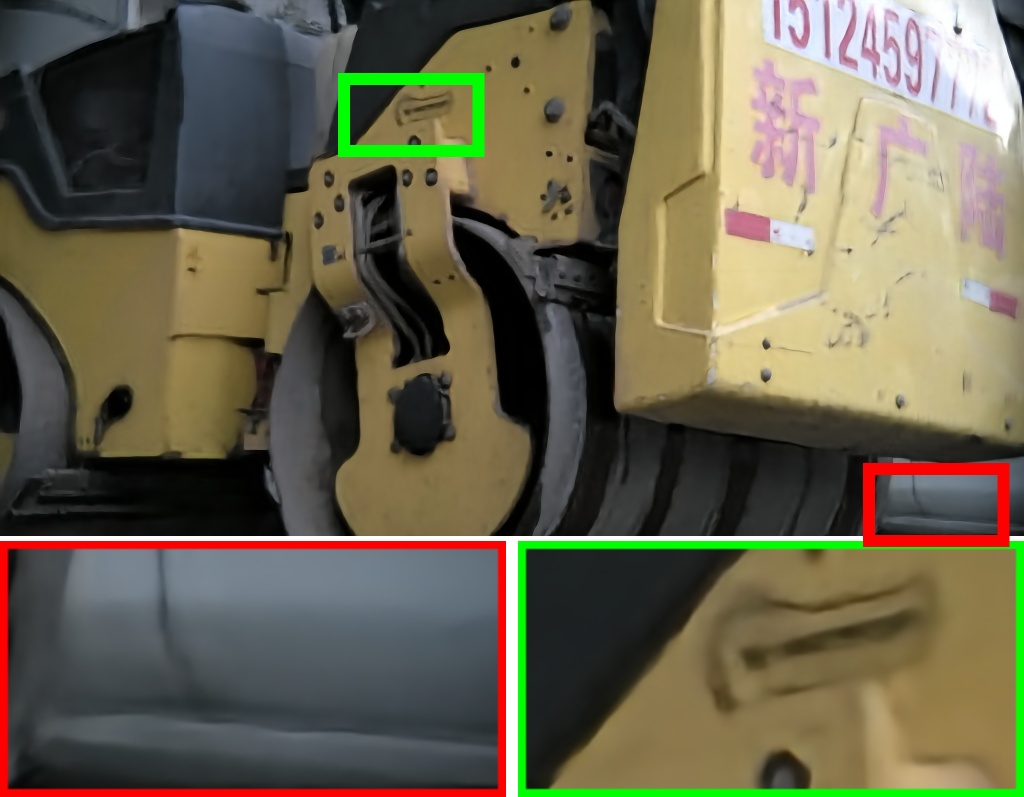}
			\end{minipage}%
			\hspace{0.04cm}
		}%
		\subfigure[UNet*]{
			\begin{minipage}[t]{0.24\linewidth}
				\centering
				\includegraphics[width=1\linewidth]{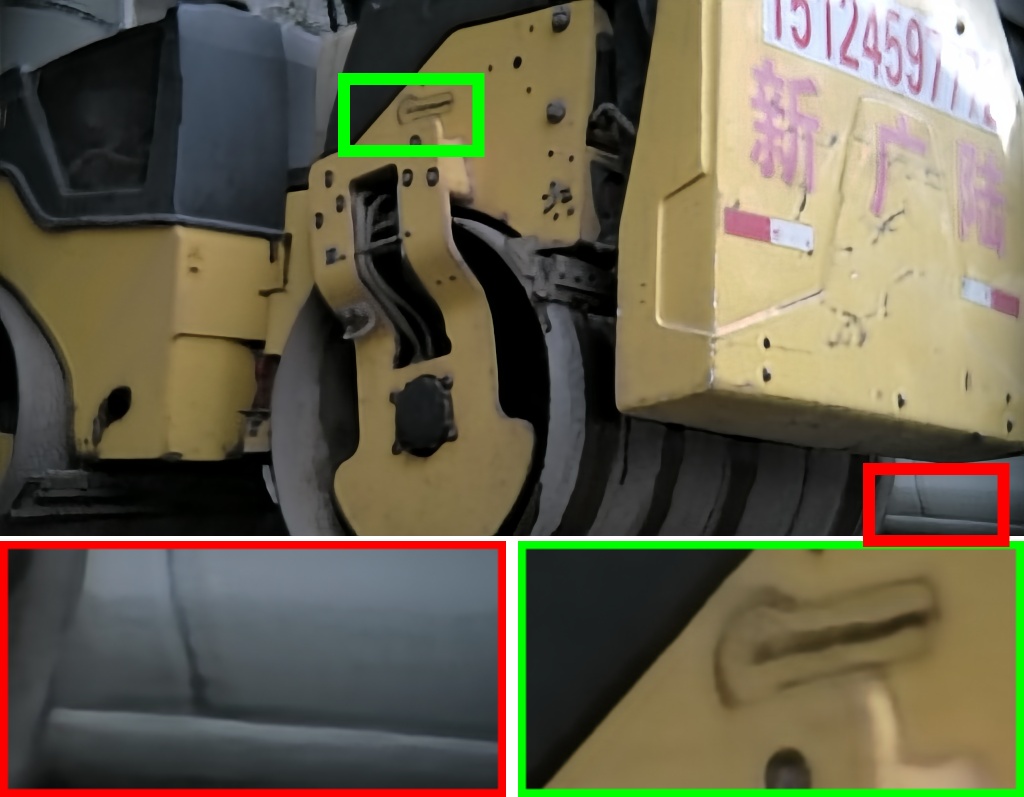}
			\end{minipage}%
			\hspace{0.04cm}
		}%
		\subfigure[MPRNet*]{
			\begin{minipage}[t]{0.24\linewidth}
				\centering
				\includegraphics[width=1\linewidth]{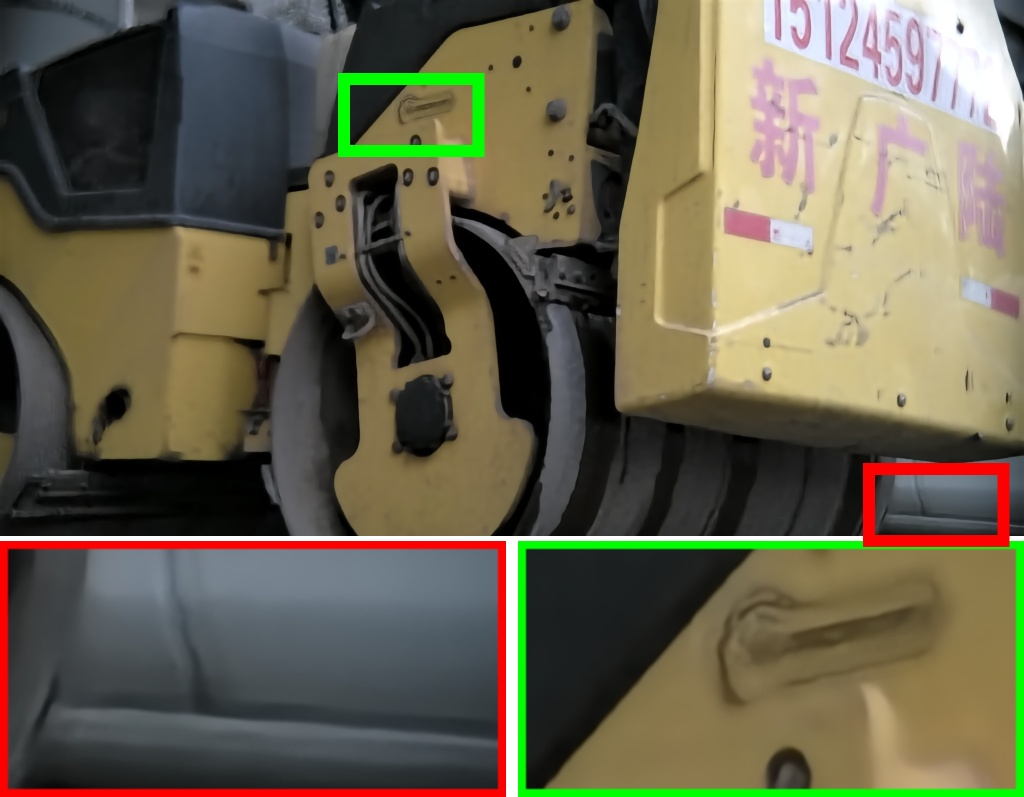}
			\end{minipage}
		}%
		\subfigure[GT]{
			\begin{minipage}[t]{0.24\linewidth}
				\centering
				\includegraphics[width=1\linewidth]{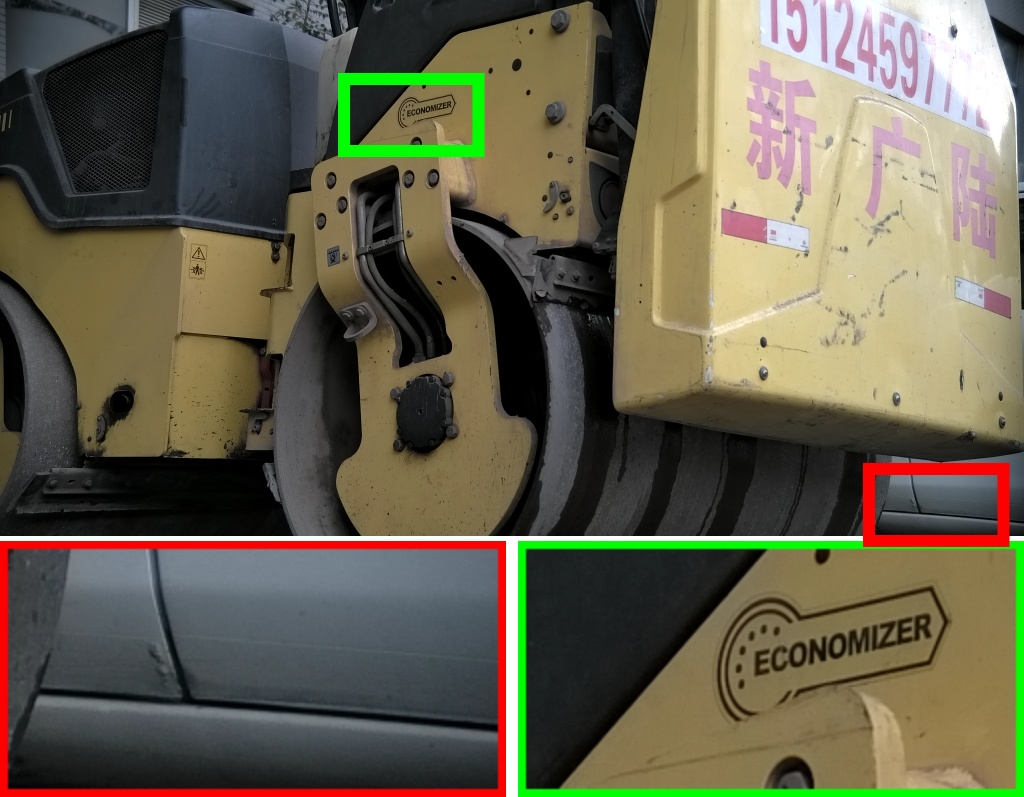}
			\end{minipage}
		}%
		\centering
		\caption{Visual comparison between different methods on SDD dataset.}
	\end{figure*}
	\begin{figure*}[htbp]
		\centering
		\label{cp2}
		\subfigure[Input]{
			\begin{minipage}[t]{0.24\linewidth}
				\centering
				\includegraphics[width=1\linewidth]{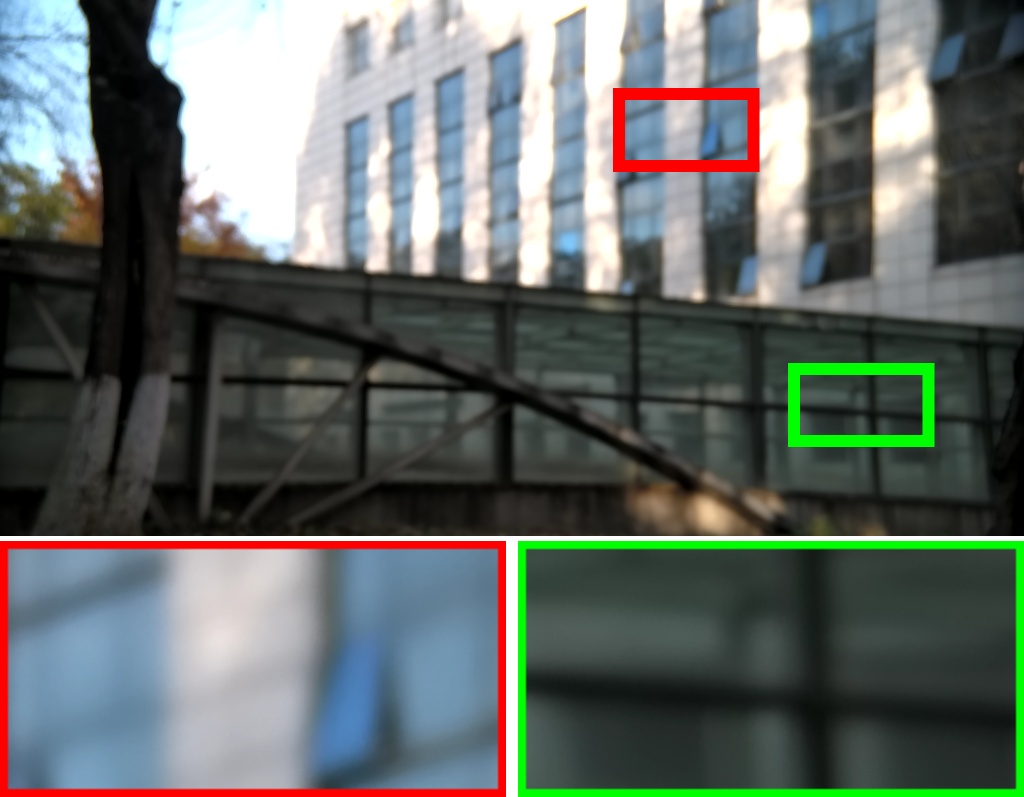}
			\end{minipage}%
			\hspace{0.04cm}
		}%
		\subfigure[UNet]{
			\begin{minipage}[t]{0.24\linewidth}
				\centering
				\includegraphics[width=1\linewidth]{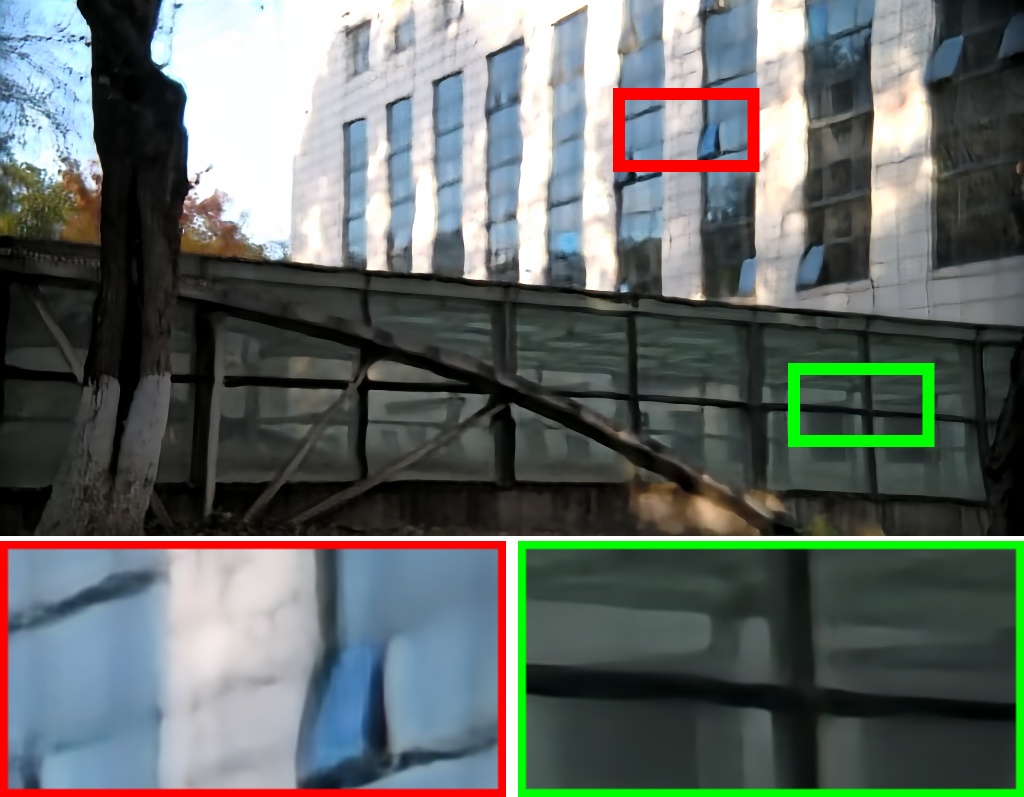}
			\end{minipage}%
			\hspace{0.04cm}
		}%
		\subfigure[MPRNet]{
			\begin{minipage}[t]{0.24\linewidth}
				\centering
				\includegraphics[width=1\linewidth]{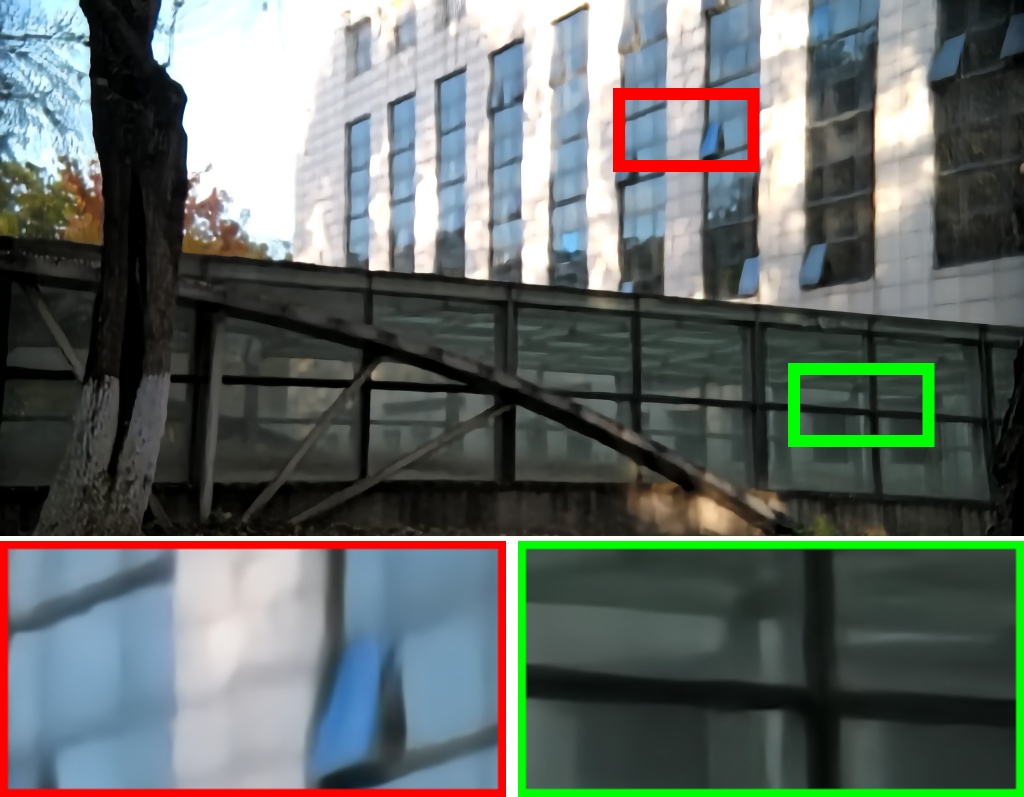}
			\end{minipage}
		}%
		\subfigure[DPDNet$_{S}$]{
			\begin{minipage}[t]{0.24\linewidth}
				\centering
				\includegraphics[width=1\linewidth]{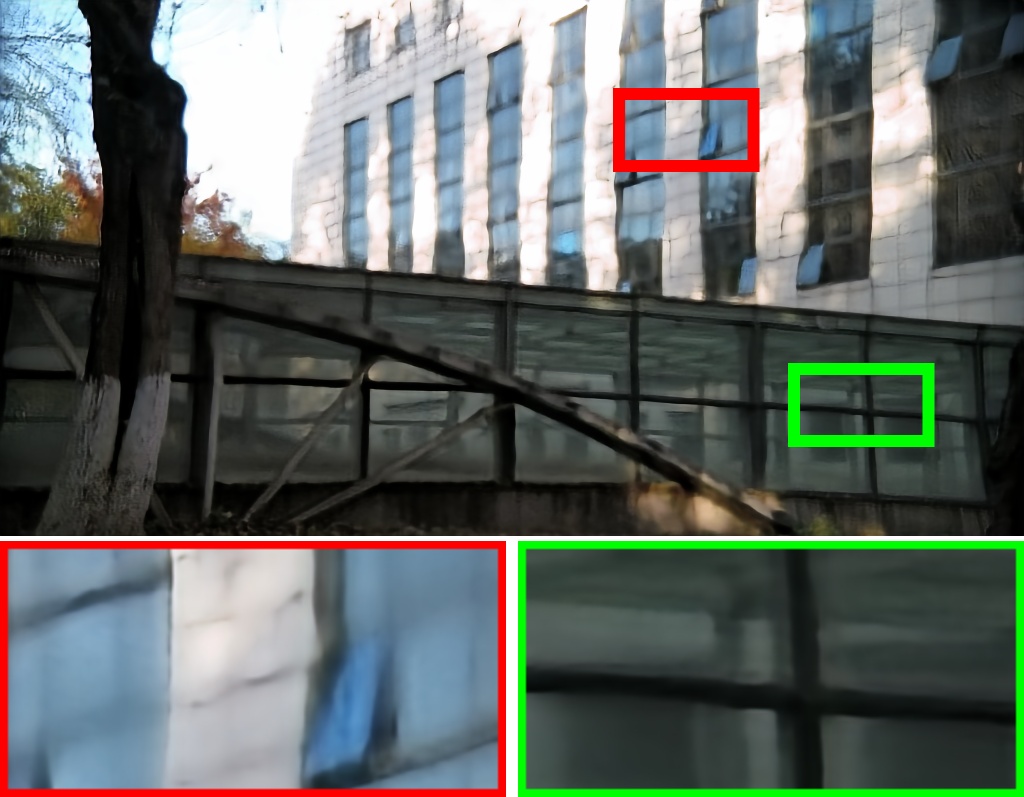}
			\end{minipage}
		}%
		\quad
		\subfigure[DMPHN]{
			\begin{minipage}[t]{0.25\linewidth}
				\centering
				\includegraphics[width=1\linewidth]{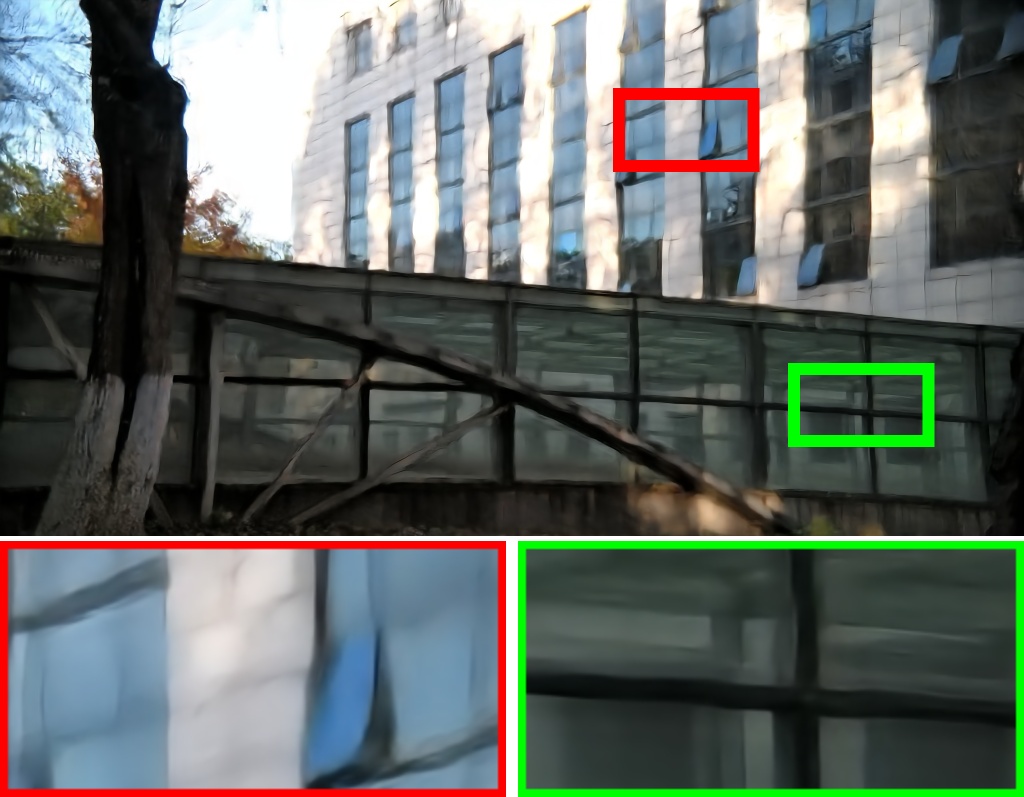}
			\end{minipage}%
			\hspace{0.04cm}
		}%
		\subfigure[UNet*]{
			\begin{minipage}[t]{0.24\linewidth}
				\centering
				\includegraphics[width=1\linewidth]{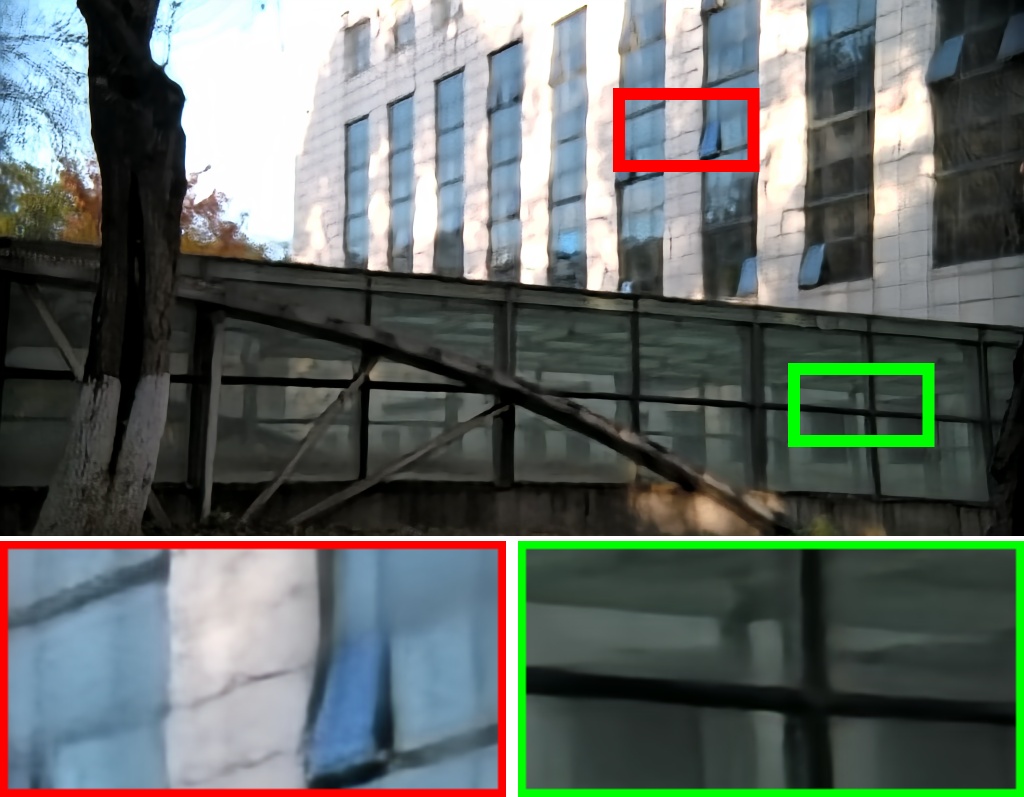}
			\end{minipage}%
			\hspace{0.04cm}
		}%
		\subfigure[MPRNet*]{
			\begin{minipage}[t]{0.24\linewidth}
				\centering
				\includegraphics[width=1\linewidth]{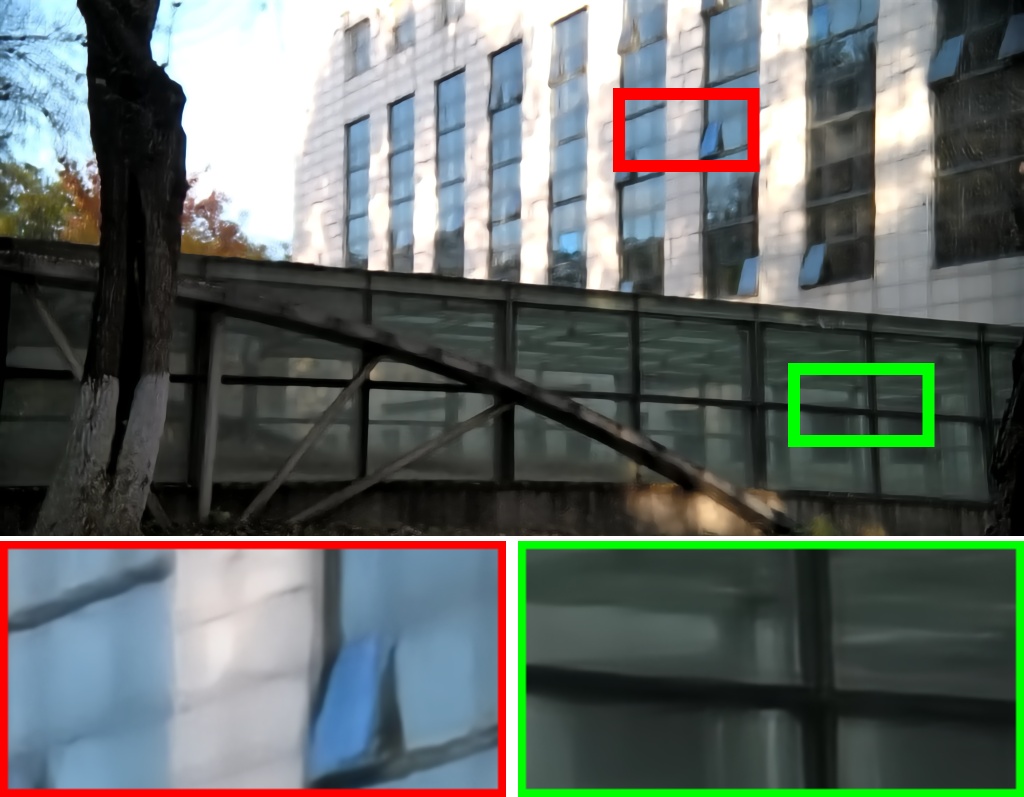}
			\end{minipage}
		}%
		\subfigure[GT]{
			\begin{minipage}[t]{0.24\linewidth}
				\centering
				\includegraphics[width=1\linewidth]{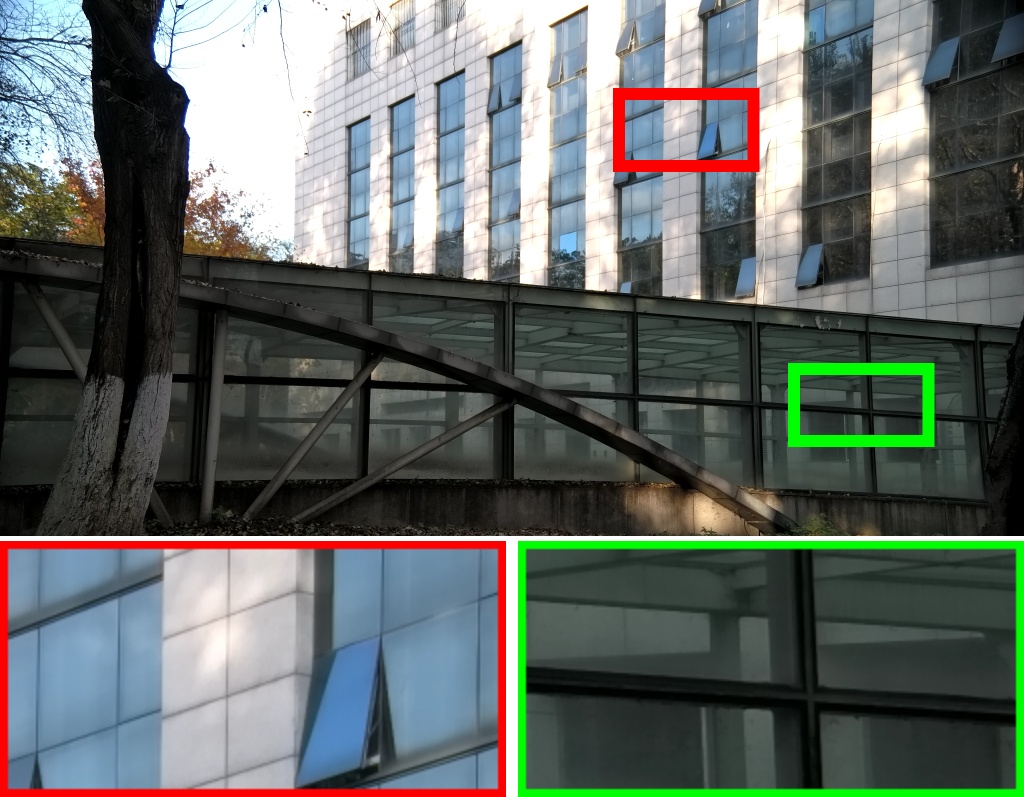}
			\end{minipage}
		}%
		\centering
		\caption{Visual comparison between different methods on SDD dataset.}
	\end{figure*}

\begin{figure*}[h]
	\subfigure[Visual comparison between different methods on DPDD dataset.]{
		\begin{tabular}{cccccc}			
			\includegraphics[width=0.155\textwidth]{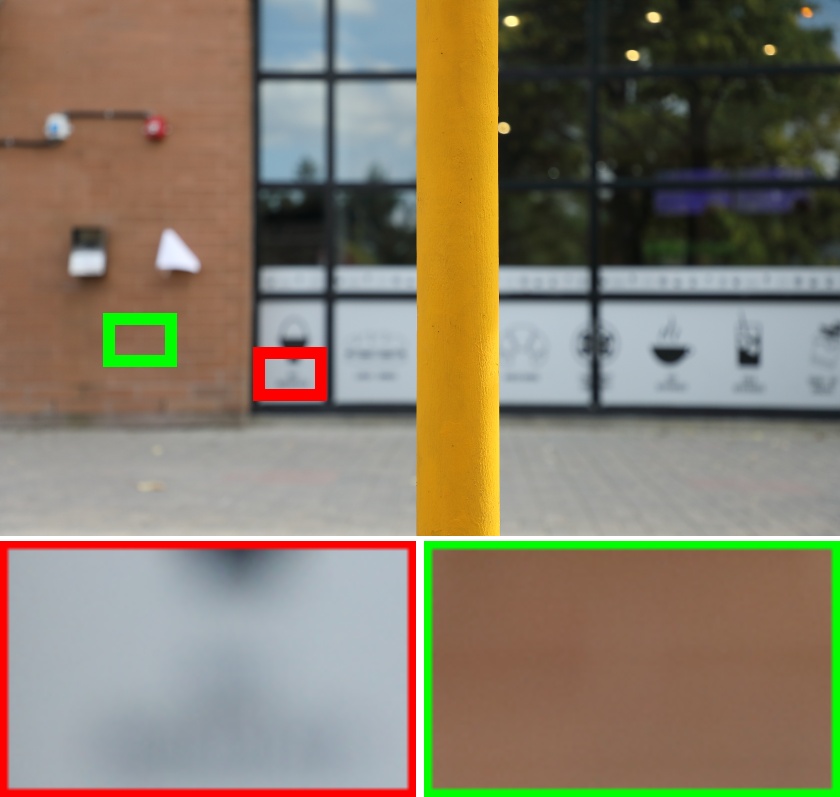}&\hspace{-4mm}
			\includegraphics[width=0.155\textwidth]{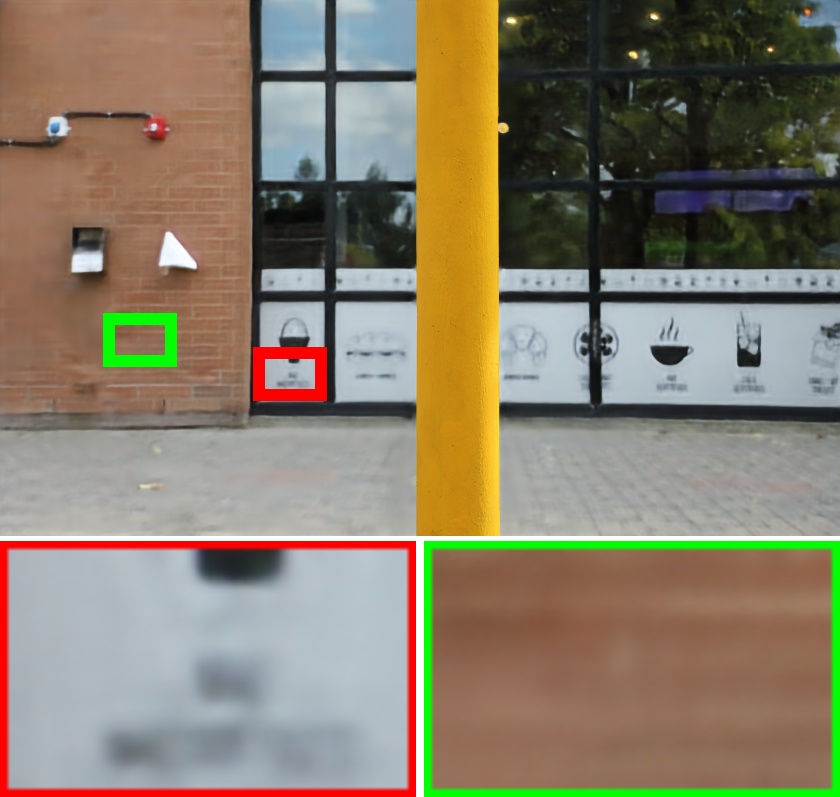}&\hspace{-4mm}
			\includegraphics[width=0.155\textwidth]{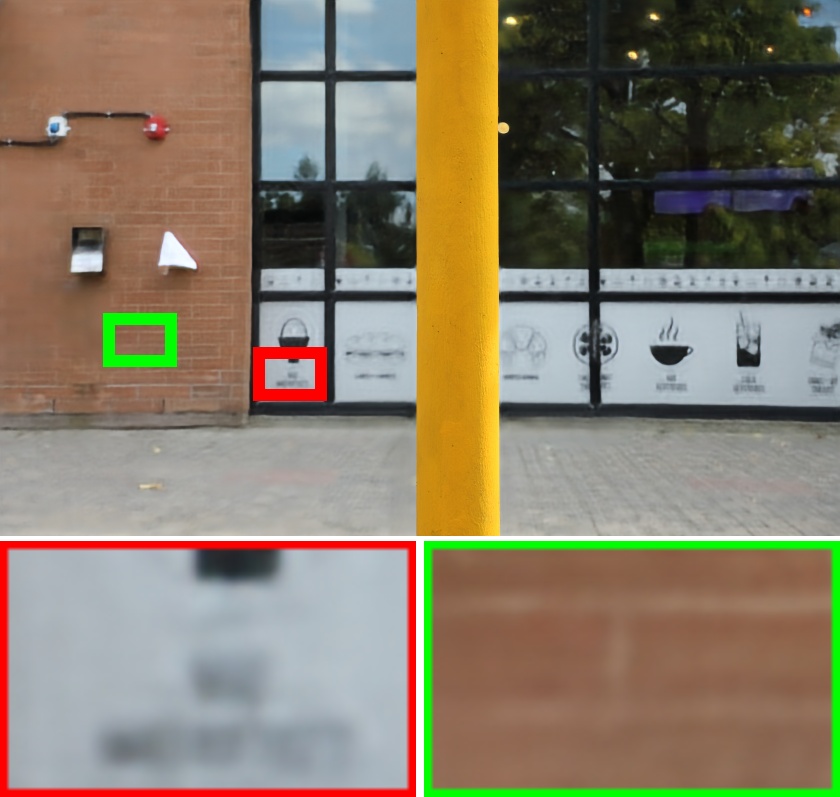}&\hspace{-4mm}
			\includegraphics[width=0.155\textwidth]{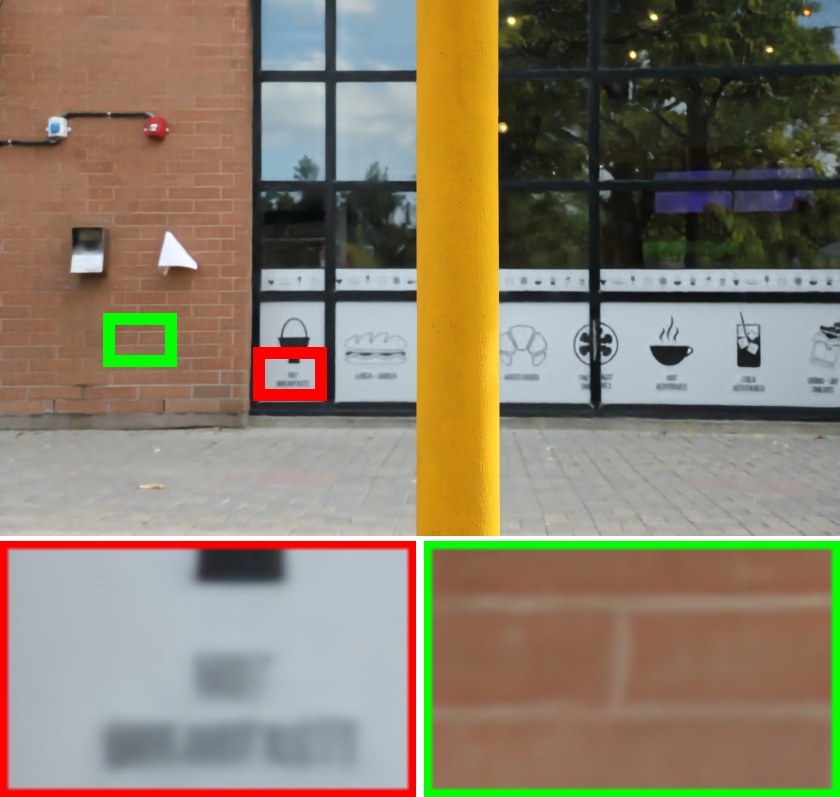}&\hspace{-4mm}
			\includegraphics[width=0.155\textwidth]{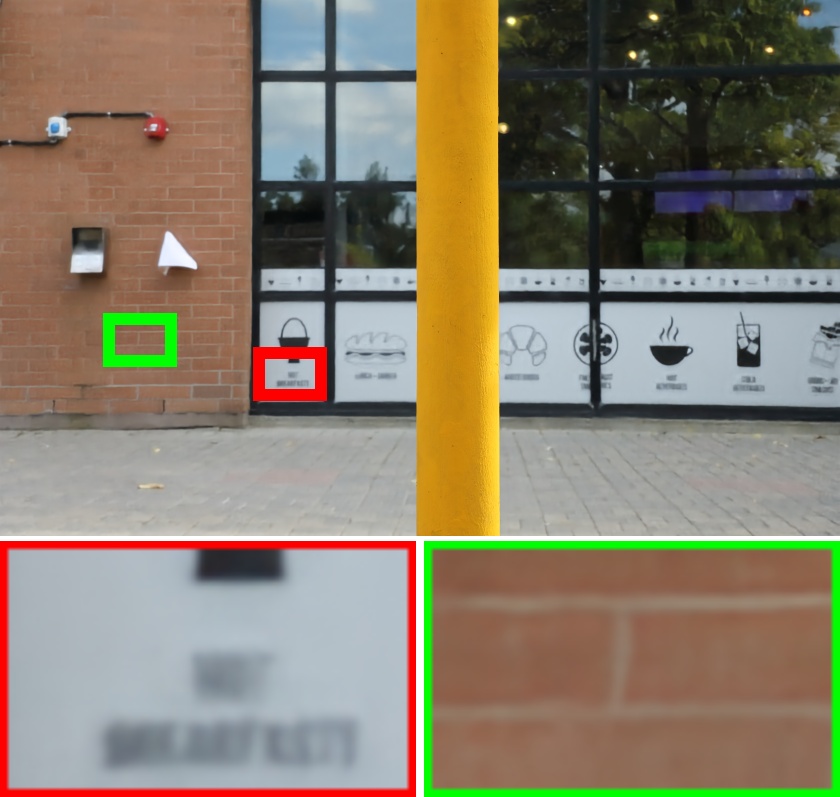}&\hspace{-4mm}	\includegraphics[width=0.155\textwidth]{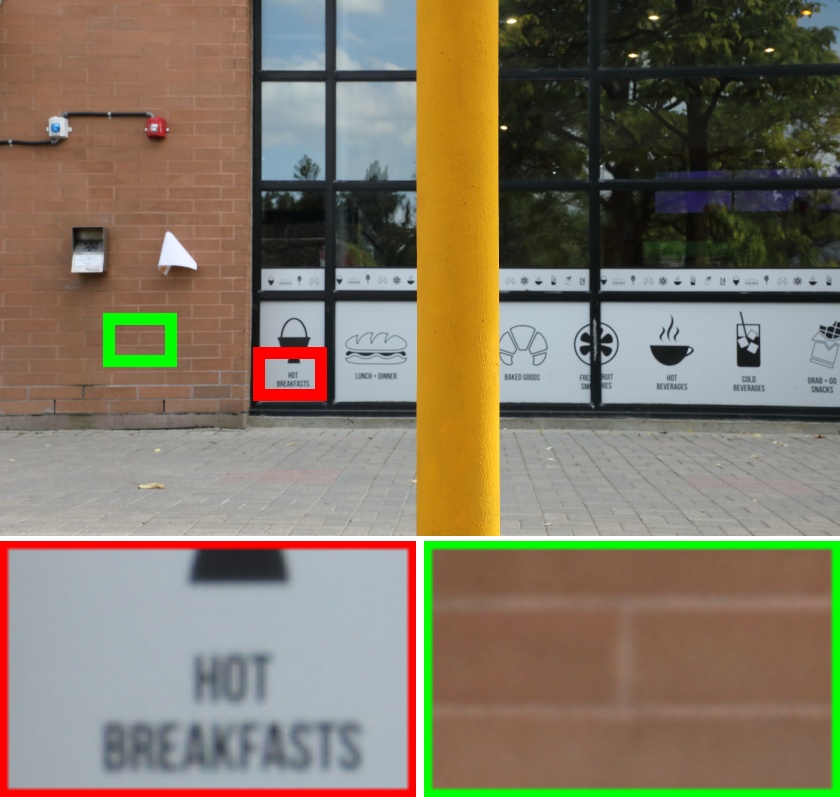}
			\\
			\includegraphics[width=0.155\textwidth]{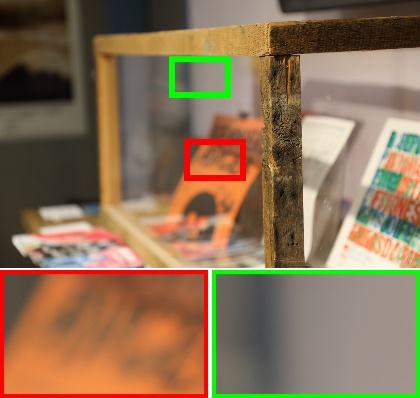}&\hspace{-4mm}
			\includegraphics[width=0.155\textwidth]{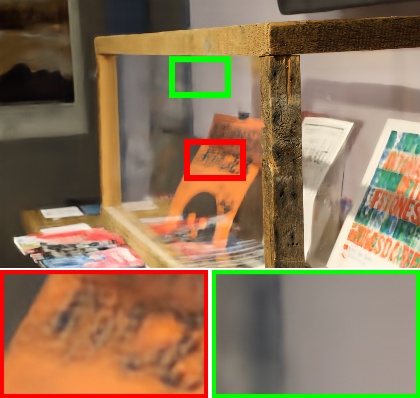}&\hspace{-4mm}
			\includegraphics[width=0.155\textwidth]{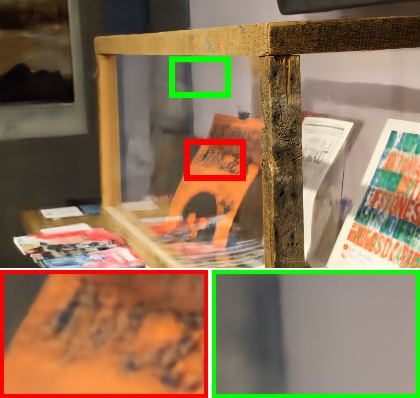}&\hspace{-4mm}
			\includegraphics[width=0.155\textwidth]{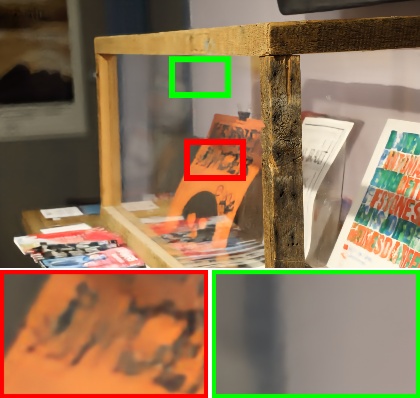}&\hspace{-4mm}
			\includegraphics[width=0.155\textwidth]{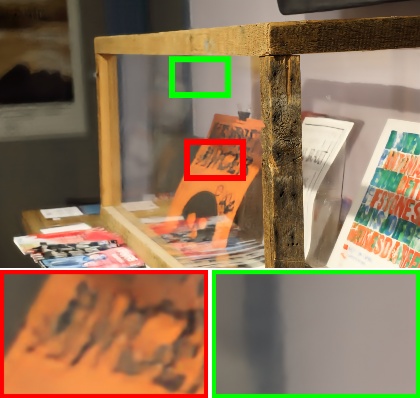}&	\hspace{-4mm}		\includegraphics[width=0.155\textwidth]{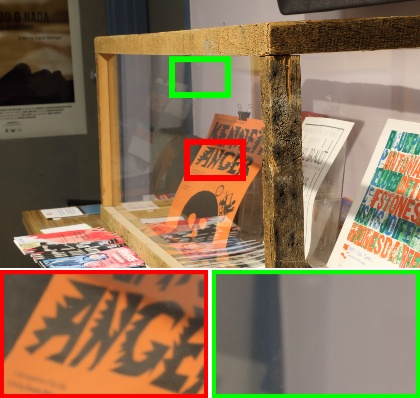}
			\\
			Input&\hspace{-4mm}
			UNet&\hspace{-4mm}
			UNet*&\hspace{-4mm}
			MPRNet& \hspace{-4mm}
			MPRNet*&\hspace{-4mm}
			GT
			\\
			\vspace{0.05cm}
		\end{tabular}
	}
\vspace{0.15cm}
	\subfigure[Visual comparison between different methods on DPDD and RealDOF datasets. The first two rows are from DPDD and the last two rows are from RealDOF.]{
		\begin{tabular}{cccccc}			
			\includegraphics[width=0.155\textwidth]{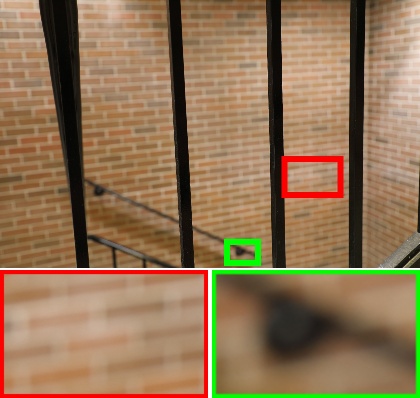}&\hspace{-4mm}
			\includegraphics[width=0.155\textwidth]{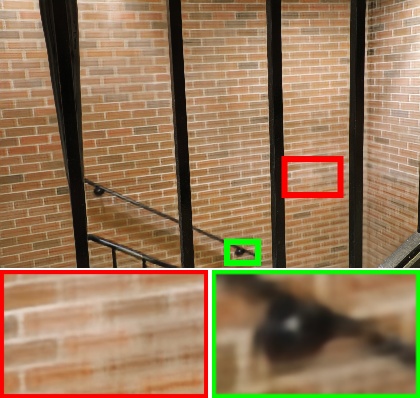}&\hspace{-4mm}
			\includegraphics[width=0.155\textwidth]{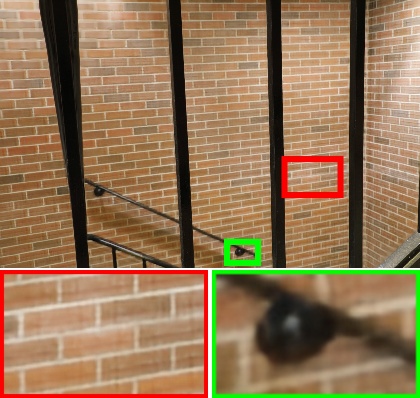}&\hspace{-4mm}
			\includegraphics[width=0.155\textwidth]{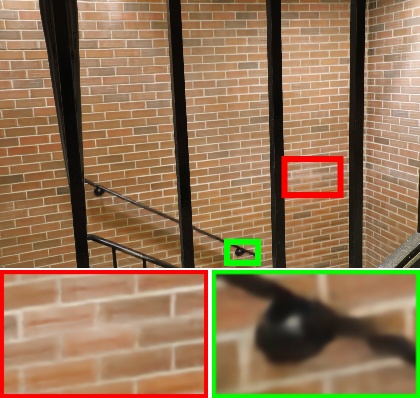}&\hspace{-4mm}
			\includegraphics[width=0.155\textwidth]{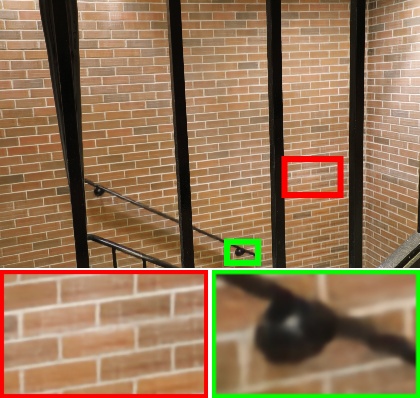}&\hspace{-4mm}	\includegraphics[width=0.155\textwidth]{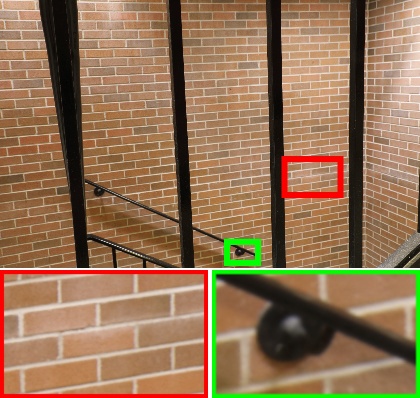}
			\\
			\includegraphics[width=0.155\textwidth]{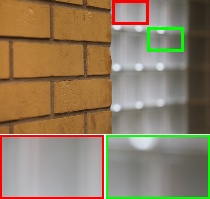}&\hspace{-4mm}
			\includegraphics[width=0.155\textwidth]{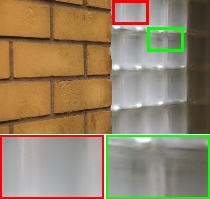}&\hspace{-4mm}
			\includegraphics[width=0.155\textwidth]{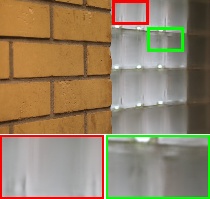}&\hspace{-4mm}
			\includegraphics[width=0.155\textwidth]{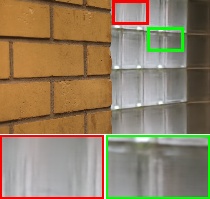}&\hspace{-4mm}
			\includegraphics[width=0.155\textwidth]{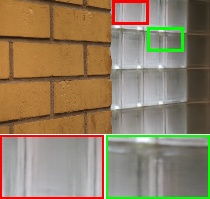}&	\hspace{-4mm}		\includegraphics[width=0.155\textwidth]{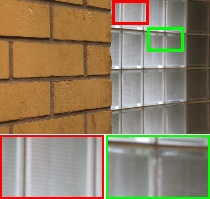}
			\\
			\includegraphics[width=0.155\textwidth]{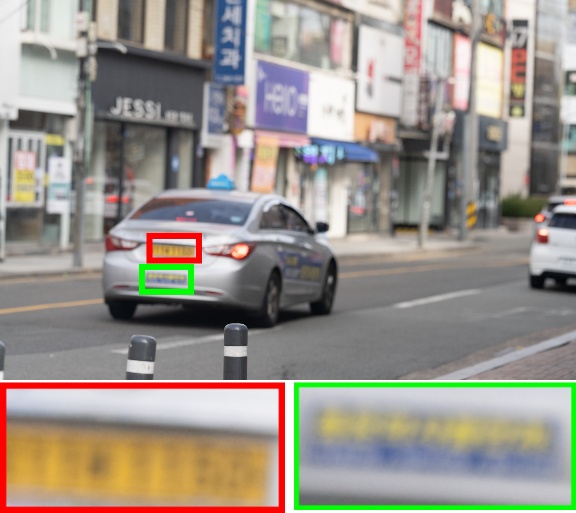}&\hspace{-4mm}
			\includegraphics[width=0.155\textwidth]{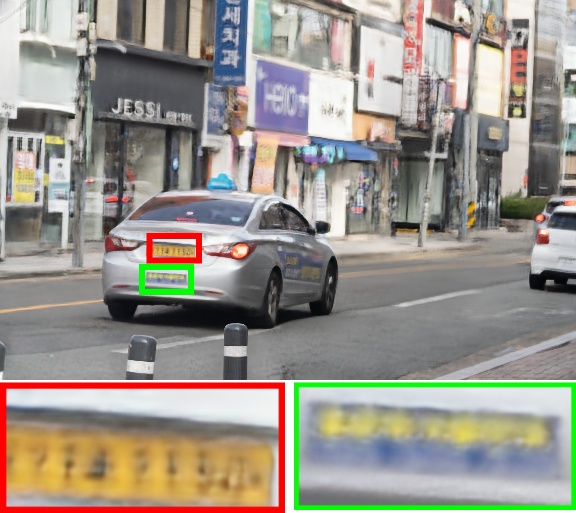}&\hspace{-4mm}
			\includegraphics[width=0.155\textwidth]{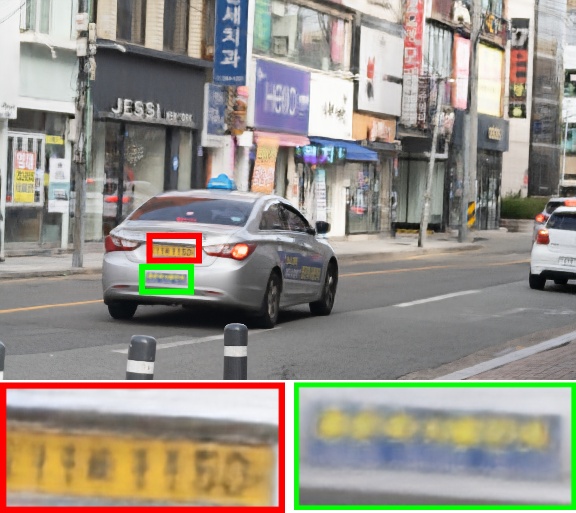}&\hspace{-4mm}
			\includegraphics[width=0.155\textwidth]{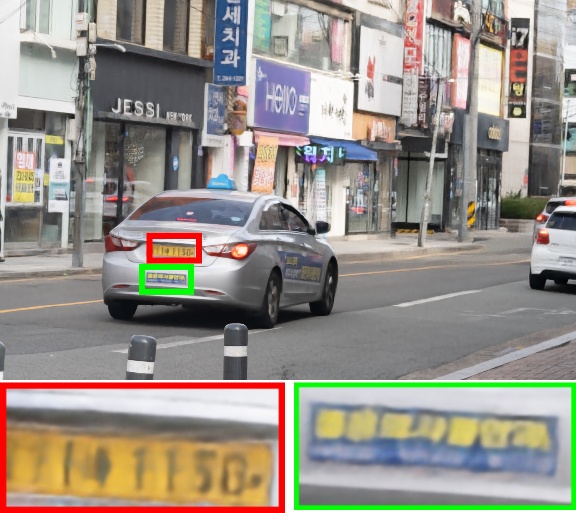}&\hspace{-4mm}
			\includegraphics[width=0.155\textwidth]{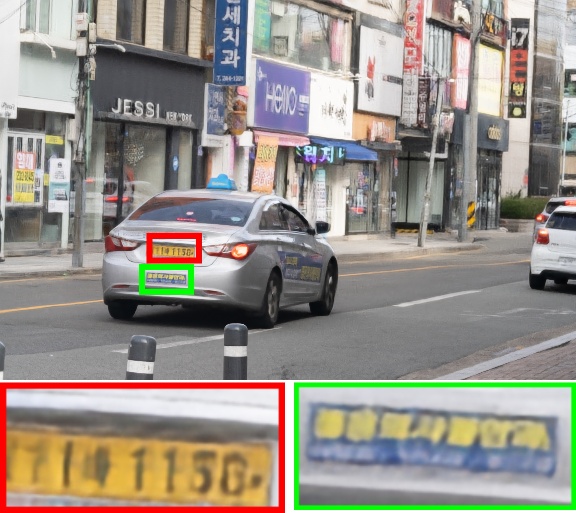}&\hspace{-4mm}			\includegraphics[width=0.155\textwidth]{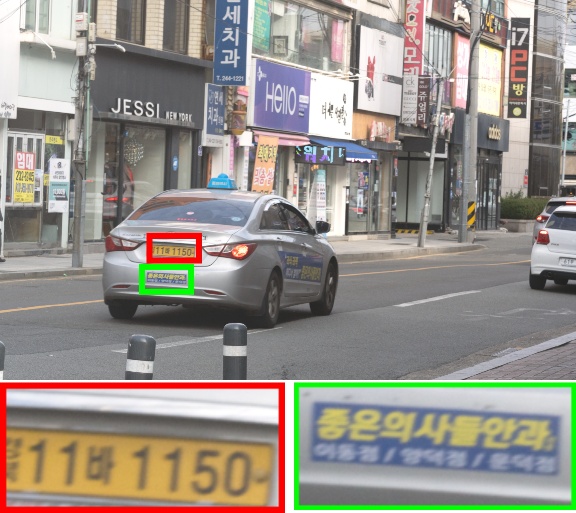}
			\\
			\includegraphics[width=0.155\textwidth]{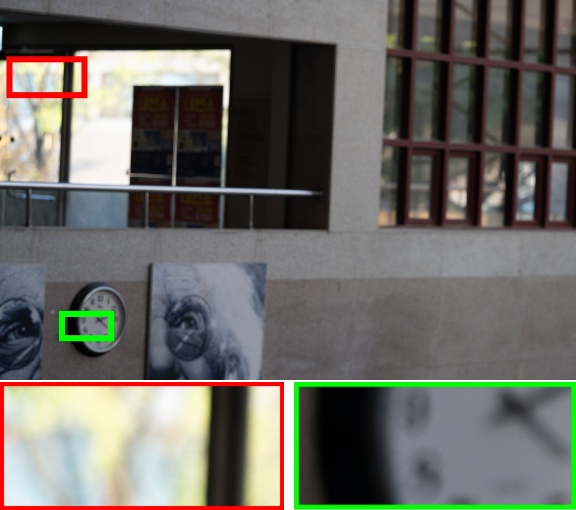}&\hspace{-4mm}
			\includegraphics[width=0.155\textwidth]{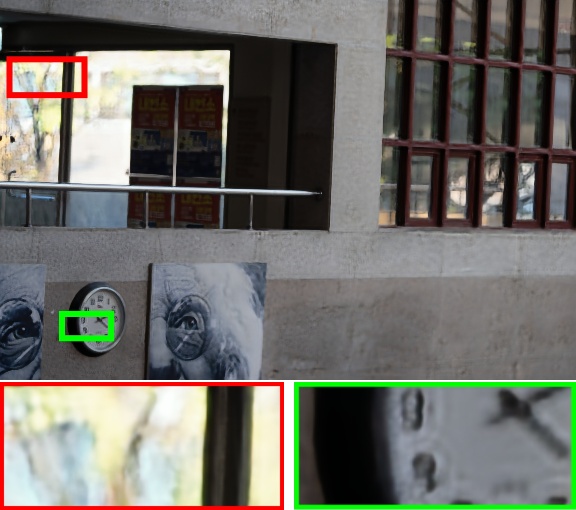}&\hspace{-4mm}
			\includegraphics[width=0.155\textwidth]{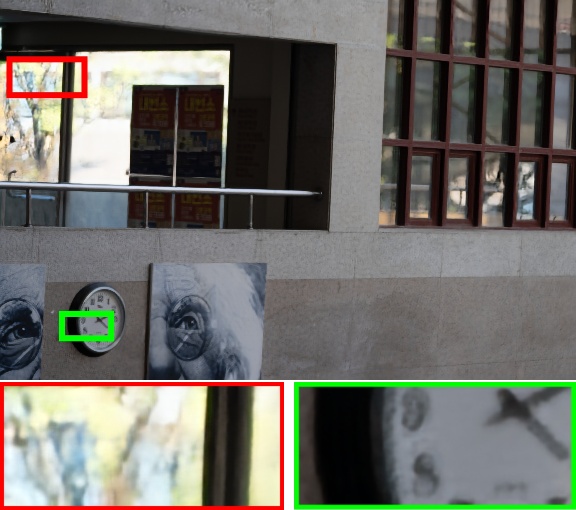}&\hspace{-4mm}
			\includegraphics[width=0.155\textwidth]{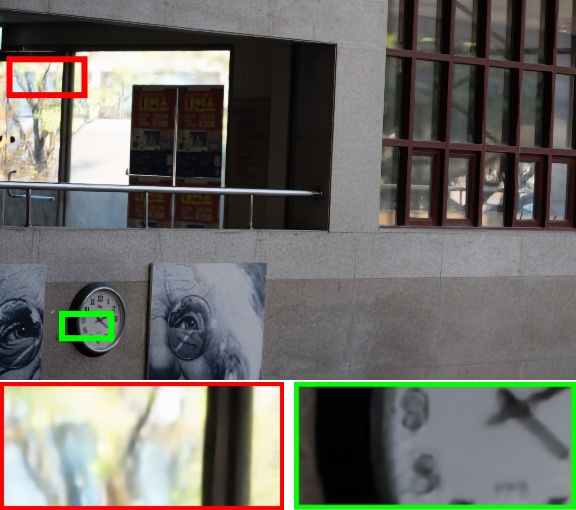}&\hspace{-4mm}
			\includegraphics[width=0.155\textwidth]{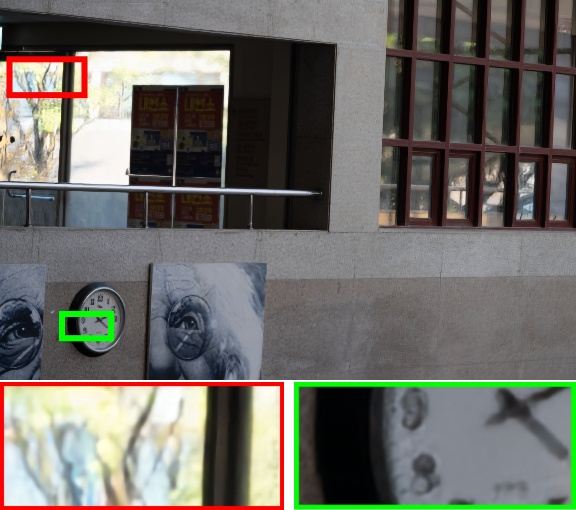}&\hspace{-4mm}
			\includegraphics[width=0.155\textwidth]{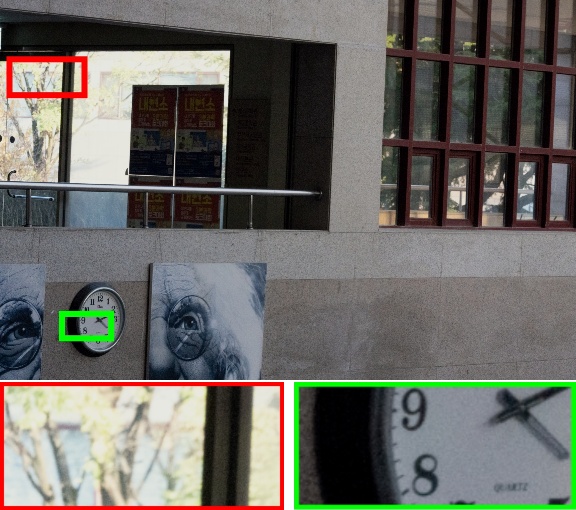}
			\\
			Input&\hspace{-4mm}
			DPDNet$_{S}$&\hspace{-4mm}
			Son et al. &\hspace{-4mm}
			IFAN& \hspace{-4mm}
			IFAN*&\hspace{-4mm}
			GT
			\\
			\vspace{0.05cm}
		\end{tabular}
	}
	\caption{Visual results on DPDD and RealDOF dataset.}
	\label{fig:compare}
\end{figure*}

\end{document}